\documentclass{article}

\PassOptionsToPackage{numbers,compress}{natbib}

\usepackage[dblblindworkshop,final]{neurips_2025}

\providecommand{\justification}[1]{#1}

\usepackage[utf8]{inputenc}
\usepackage[T1]{fontenc}
\usepackage{microtype}
\usepackage{nicefrac}

\usepackage{amsmath}
\usepackage{amssymb}
\usepackage{amsfonts}

\usepackage{algorithm}
\usepackage{algpseudocode}

\usepackage{graphicx}
\usepackage{subcaption}
\usepackage{booktabs}
\usepackage{tabularx}
\usepackage{adjustbox}
\usepackage{wrapfig}

\usepackage{xcolor}
\usepackage{url}
\usepackage{hyperref}
\usepackage{comment}
\usepackage{cancel}

\title{Tabular Deep Learning vs Classical Machine Learning for Urban Land Cover Classification}

\workshoptitle{The 5\textsuperscript{th} Muslims In ML (MusIML)}

\author{%
  \textsuperscript{1}Muntasir Tabasum\\
  Department of Geology and Geography\\
  West Virginia University\\
  Morgantown, WV 26506, USA \\
  \texttt{mt00079@mix.wvu.edu}
  \And
  \textsuperscript{2}Tanpia Tasnim\\
  Department of CSE\\
  Green University of Bangladesh\\
  Narayanganj-1461, Bangladesh\\
  \texttt{tanpia@cse.green.edu.bd}
  \AND
  \textsuperscript{3}Md. Ekramul Islam\\
  Department of CSE\\
  Stamford University Bangladesh\\
  Dhaka-1217, Bangladesh\\
  \texttt{eislam706@gmail.com}
  \And
  \textsuperscript{4}Al Zadid Sultan Bin Habib\thanks{Corresponding author. Website: \href{https://www.zadidhabib.com/}{https://www.zadidhabib.com}.} \\
  Lane Department of CSEE\\
  West Virginia University\\
  Morgantown, WV 26506, USA \\
  \texttt{ah00069@mix.wvu.edu}
}

\begin{document}

\maketitle

\begin{abstract}
Urban Land Cover (ULC) classification plays a crucial role in urban planning, environmental monitoring, and sustainable development. We study this task using the ULC dataset from the UCI Machine Learning Repository, which includes tabular features derived from high-resolution aerial imagery across nine classes (e.g., roads, trees, grass, water). The dataset presents typical remote sensing challenges, including high dimensionality, heterogeneous features, and class imbalance. In a unified, reproducible pipeline, we benchmark classical machine learning models (e.g., Logistic Regression, SVM, Random Forest, XGBoost, CatBoost) against Tabular Deep Learning (TDL) models (TabNet, FT-Transformer, TabTransformer, TabSeq, and 1D CNNs). To address class imbalance, we employ weighted cross-entropy loss for TDL models and evaluate performance using accuracy, macro-precision, macro-recall, macro-F1, AUC-ROC, and confusion matrices. Our results show that while tree ensembles remain strong general baselines, TDL models can match or exceed their performance when non-linear interactions are significant and imbalance handling is effective, providing complementary advantages for urban land cover mapping. See code: \url{https://github.com/mtesha/tdl-vs-ml-urbanlandcover}
\end{abstract}
%%%%%
\section{Introduction}
Urban Land Cover (ULC) classification supports sustainable development, urban planning, and environmental management by charting the geographical distribution and dynamics of essential surface types in urban areas (e.g., impervious surfaces, vegetation, and water). Accurate ULC maps support downstream analyses such as heat-island mitigation, stormwater management, and land-use monitoring. In this work, we examine ULC classification using the ULC dataset from the UCI Machine Learning Repository \citep{johnson2013_uci_ulc}, which provides tabular descriptors derived from a high-resolution aerial image. The features span multiple descriptor families (e.g., spectral, textural, shape/size) and nine classes (e.g., roads, trees, grass, water), and present typical remote-sensing challenges: high dimensionality, heterogeneous feature types, and class imbalance.

We build robust traditional Machine Learning (ML) baselines, including Logistic Regression (LR) \citep{rao1973linear}, Decision Trees \citep{quinlan1986id3}, Random Forests (RF) \citep{breiman2001rf}, Support Vector Machines (SVM) \citep{cortes1995svm}, $k$-Nearest Neighbors (KNN) \citep{cover1967knn}, Naive Bayes \citep{duda1973pattern}, and boosting methods such as AdaBoost \citep{freund1997adaboost}, Gradient Boosting \citep{friedman2001gbm}, XGBoost \citep{chen2016xgboost}, and CatBoost \citep{prokhorenkova2018catboost}, and compare them to some of the Tabular Deep Learning (TDL) models: TabNet \citep{arik2021tabnet}, TabTransformer \citep{huang2020tabtransformer,lucidrains_tabtransformer}, FT-Transformer and related modern TDL baselines \citep{gorishniy2021revisiting}, as well as TabSeq \citep{habib2024tabseq,tabseq_repo} and a 1D CNN variant \citep{lecun1998document}. To mitigate class imbalance in TDL, we use weighted cross-entropy. 

Our study is motivated by mixed evidence in the literature on when TDL provides gains over tree ensembles for tabular problems. In remote sensing and urban mapping, the integration of diverse features and the management of imbalance are paramount \citep{johnson2016osm_lulc,johnson2015swsf,wickham2014mrlc,jozdani2019urban_dl_vs_ml,jin2019ncld,ducey2018demography}. We therefore ask: (i) How do strong classical ML methods compare to TDL on ULC? (ii) Under what conditions (e.g., non-linear interactions, imbalance handling) do TDL models match or surpass tree ensembles? (iii) What practical recommendations emerge for reproducible ULC pipelines? 

\textbf{Contributions.} (1) A unified, reproducible pipeline benchmarking strong ML baselines against some of the TDL models on UCI ULC \citep{johnson2013_uci_ulc}. (2) A systematic evaluation with accuracy, macro-precision/recall/F1, AUC-ROC, and confusion matrices. (3) An analysis of imbalance mitigation (weighted cross-entropy) and when representation learning in TDL closes or exceeds the performance of tree ensembles, yielding actionable guidance for urban mapping practitioners. 
%%%%%%
\section{Related Work}
ULC mapping has been a long-standing focus in remote sensing for monitoring urban expansion and environmental change. Methodological advances span data fusion, object-based analysis, and large-scale national products. For rapid land use/land cover mapping, \citep{johnson2016osm_lulc} integrate OpenStreetMap (OSM) with Landsat time-series and report strong performance with classical classifiers (e.g., Random Forest, Naive Bayes), while noting challenges from noise and class imbalance. At the descriptor level, \citep{johnson2015swsf} propose spatially-weighted segment-level fusion (SWSF) to better aggregate low-spatial-resolution signals, improving small/narrow class discrimination when the coarse resolution is $\geq$3$\times$ the high-resolution reference. At national scale, the MRLC consortium underpins NLCD and related products (C-CAP, CDL, LANDFIRE), leveraging Landsat to deliver consistent land cover layers for the United States that enable diverse environmental applications \citep{wickham2014mrlc,jin2019ncld}. Beyond mapping itself, demographic land cover interactions have been quantified statistically; for the U.S. Great Lakes States, \citep{ducey2018demography} show population growth, housing density, and recreational/retirement status as key drivers of land conversion. 

Comparisons of modeling paradigms for urban LULC remain mixed. In an object-based setting, \citep{jozdani2019urban_dl_vs_ml} find that a multilayer perceptron can outperform gradient boosting, XGBoost, and SVMs, while CNN integrations offer limited gains under complex urban morphology and segmentation errors. In parallel, tabular-learning research has introduced modern deep models for mixed-type features (e.g., TabNet, TabTransformer, FT-Transformer) alongside strong tree ensembles \citep{arik2021tabnet,huang2020tabtransformer,gorishniy2021revisiting,chen2016xgboost,prokhorenkova2018catboost,breiman2001rf}. However, few studies systematically benchmark some of the recent TDL architectures against well-tuned classical baselines on tabular ULC descriptors where heterogeneity, high dimensionality, and class imbalance are prominent. Earlier UHI-based land cover studies relied on Landsat imagery and GIS-based statistical analyses but did not use modern AI/ML/TDL methods for ULC change classification \citep{ritu1, ritu2}. A Khulna traffic-delay study similarly used survey/GIS and speed-flow analysis but did not employ modern ML/TDL prediction pipelines \citep{bashar}.

\textbf{Our focus.} We address this gap by evaluating strong classical ML methods and recent TDL models within a unified, reproducible pipeline on the UCI ULC dataset, with attention to imbalance mitigation and multi-metric reporting. This complements prior work on fused imagery and object-based analysis by centering the tabular descriptor regime common in operational ULC workflows.
%%%%%
\vspace{-0.4cm}
\section{Methodology}
\vspace{-0.2cm}
\paragraph{Dataset.}
We utilize the ULC dataset from the UCI Machine Learning Repository \citep{johnson2013_uci_ulc}. It provides engineered tabular descriptors from a single high-resolution aerial image with samples of $n_{\text{train}}{=}168$, $n_{\text{test}}{=}507$, $d{=}147$ features, and $K{=}9$ classes (asphalt, building, car, concrete, grass, pool, shadow, soil, tree), exhibiting heterogeneous descriptors and class imbalance.
\paragraph{Preprocessing.}
Labels are integer-encoded $\{1,\dots,K\}$. Features are z-scored using training statistics, i.e., $\tilde{x}_{ij}=(x_{ij}-\mu_j)/\sigma_j$ with $(\mu_j,\sigma_j)$ from $X_{\text{train}}$ and reused for validation/test. A stratified $10\%$ of $X_{\text{train}}$ is held out for validation.
\paragraph{Models.}
\emph{Classical ML and GBDT:} Logistic Regression (LR) \citep{rao1973linear}, Decision Tree (DT) \citep{quinlan1986id3}, Random Forest (RF) \citep{breiman2001rf}, SVM \citep{cortes1995svm}, $k$NN \citep{cover1967knn}, Naive Bayes (NB) \citep{duda1973pattern}, Gradient Boosting (GBM) \citep{friedman2001gbm}, AdaBoost \citep{freund1997adaboost}, XGBoost \citep{chen2016xgboost}, CatBoost \citep{prokhorenkova2018catboost}, and a shallow MLP \citep{rumelhart1986backprop}.
\emph{TDL:} TabNet \citep{arik2021tabnet}, TabTransformer \citep{huang2020tabtransformer,lucidrains_tabtransformer}, FT-Transformer \citep{gorishniy2021revisiting}, a 1D CNN \citep{lecun1998document}, and TabSeq \citep{habib2024tabseq,tabseq_repo}.
%%%%%
\paragraph{Discussion.} Tree ensembles (RF, GBM, XGBoost, CatBoost) are strong defaults on heterogeneous tabular data. They capture non-linearities and higher-order interactions with modest tuning, tolerate mixed feature scales, and are comparatively robust on small-$n$ regimes like ULC. CatBoost additionally handles categorical variables and reduces target leakage via ordered boosting, which often improves calibration and minority-class recall. Linear/margin methods (LR, SVM) and $k$NN/NB provide interpretable, low-variance references that help contextualize gains from more complex models. The shallow MLP probes whether simple neural capacity suffices without tabular-specific inductive bias. TDL models introduce representation learning tailored to feature interactions (transformers) or sparse attentive selection (TabNet); however, they typically require stronger regularization and benefit from imbalance-aware training to avoid overfitting at this scale. The 1D CNN imposes a sequential inductive bias over features (useful for local groups of correlated descriptors but sensitive to feature order), whereas TabSeq explicitly learns a stable ordering and denoised representation before classification. Together, the suite spans linear $\rightarrow$ tree $\rightarrow$ neural paradigms, enabling a controlled comparison under identical preprocessing, validation, and metrics.
\paragraph{Training, loss, and metrics.}
All models train on $(X_{\text{train}}^{\text{scaled}},y_{\text{train}})$, select by validation loss/accuracy, and report on $(X_{\text{test}}^{\text{scaled}},y_{\text{test}})$. For TDL we use class-weighted cross-entropy
$\mathcal{L}_{\mathrm{wCE}}=-\frac{1}{N}\sum_{i=1}^{N} w_{y_i}\log p_{\theta}(y_i\mid x_i)$ with $w_k=\frac{N}{K N_k}$, where $N_k$ counts class-$k$ training samples; tree ensembles are left unweighted. We report test accuracy; macro-precision/recall/F1; one-vs-rest macro AUC-ROC; and confusion matrices for error analysis.
\paragraph{Key hyperparameters.}
Optimizer: Adam, batch size $32$, learning rate $10^{-3}$, early stopping on validation loss.
\emph{TabNet} \citep{arik2021tabnet}: $n_d{=}64$, $n_a{=}64$, $n_{\text{steps}}{=}3$, $\gamma{=}1.3$, $\lambda_{\text{sparse}}{=}10^{-3}$, lr $2{\times}10^{-2}$, patience $20$.
\emph{TabTransformer} \citep{huang2020tabtransformer,lucidrains_tabtransformer}: dim $32$, depth $6$, heads $8$, attn/ffn dropout $0.1$, MLP multipliers $(4,2)$, lr $10^{-3}$, weighted cross-entropy.
\emph{FT-Transformer} \citep{gorishniy2021revisiting}: dim $32$, depth $6$, heads $8$, dropout $0.1$.
\emph{1D CNN} \citep{lecun1998document}: conv ($64$ ch, $k{=}3$) $\rightarrow$ max-pool ($k{=}2$, stride $2$) $\rightarrow$ FC($128$)$\times2$ with dropout $0.3$ $\rightarrow$ softmax.
\emph{TabSeq} \citep{habib2024tabseq,tabseq_repo}: feature ordering via $k$-means ($k{=}5$) + dispersion minimization; DAE encoder/decoder $\{128,64\}$ with dropout $0.2$ and noise $0.1$; classifier MLP $\{128,64\}$ with dropout $0.5$. Optuna \cite{optuna} was used to tune the hyperparameters of the TDL models, while the classical ML baselines followed scikit-learn defaults, and the GBDT models used their respective library defaults (e.g., CatBoost \citep{prokhorenkova2018catboost}, XGBoost \citep{chen2016xgboost}).
\paragraph{Implementation.}
Python/Scikit-learn for classical models; PyTorch/TensorFlow for TDL; fixed seeds and persisted preprocessing stats for reproducibility. 
%%%%%%
\section{Results and Analysis}
\begin{figure}[htbp]
  \centering
  \includegraphics[width=0.85\linewidth]{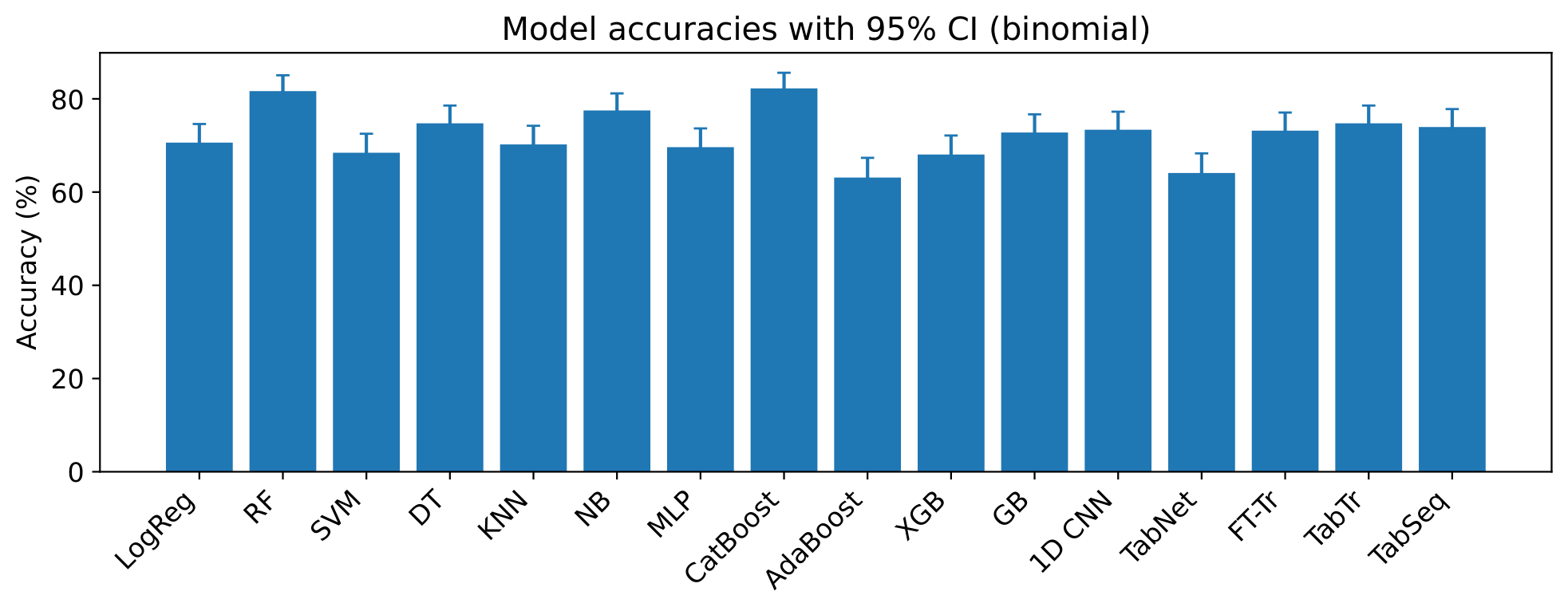}%
  \caption{Model accuracies on the ULC test set with 95\% binomial confidence intervals.}
  \label{fig:ulc-accuracy}
\end{figure}

\paragraph{Setup.} We evaluate 16 models (classical ML and TDL) on the UCI ULC test set and report Test Accuracy (\%), macro Precision/Recall/F1, and one-vs-rest macro AUC-ROC. Summary tables (Tabs.~\ref{tab:weighted_tdl}--\ref{tab:all_models}) are complemented by six visualizations (Fig.\ref{fig:dl_acc} and Figs.~\ref{fig:acc_all_app}--\ref{fig:weighted_metrics_app} in Appendix~\ref{app:morefigs}).
\paragraph{Baseline performance (all models).}
CatBoost attains the highest test accuracy (82.25\%), followed by Random Forest (81.66\%) and Naive Bayes (77.51\%), confirming the strength of tree ensembles and simple probabilistic baselines on tabular descriptors. Among TDL models, TabTransformer (74.75\%) and TabSeq (73.96\%) are competitive with the best non-ensemble classical baselines. In Appendix~\ref{app:morefigs}, Fig.~\ref{fig:acc_all_app} contrasts accuracies across all models; Fig.~\ref{fig:radar_top_app} compares Precision/Recall/F1/AUC for top contenders; Fig.~\ref{fig:heatmap_all_app} shows a full metric heatmap, highlighting trade-offs (e.g., models with similar accuracy but different Recall/AUC). Fig.~\ref{fig:dl_acc} focuses on the five deep models. Table~\ref{tab:all_models} presents the full results. See Appendix~\ref{ad} for ROC curves and confusion matrices for all models.
\paragraph{TDL with weighted cross-entropy.}
To address class imbalance, we train 1D-CNN, TabNet, FT-Transformer, TabTransformer, and TabSeq with class-weighted cross-entropy. Accuracy results (Fig.~\ref{fig:weighted_acc_app} in Appendix~\ref{app:morefigs}) show TabSeq (73.57\%) and TabTransformer (73.37\%) leading this group. Fig.~\ref{fig:weighted_metrics_app} in Appendix~\ref{app:morefigs} also indicates consistent gains in Recall and F1 for most models under weighting, echoing improvements noted in Table~\ref{tab:weighted_tdl}. See Appendix~\ref{wad} for ROC curves and confusion matrices for TDL models with weighted cross entropy.
%%%%
\begin{wraptable}{r}{0.4\columnwidth}
%\vspace{-0.8em} % tweak vertical placement
\centering
\caption{TDL models with class-weighted cross-entropy.}
\label{tab:weighted_tdl}
\tiny
\setlength{\tabcolsep}{4pt}
\begin{tabular}{lccccc}
\toprule
\textbf{Model} & \textbf{Acc.\%} & \textbf{Prec.} & \textbf{Rec.} & \textbf{F1} & \textbf{AUC} \\
\midrule
1D CNN       & 73.18 & 0.70 & 0.74 & 0.71 & 0.94 \\
TabNet      & 66.67 & 0.65 & 0.70 & 0.66 & 0.93 \\
FT-Transformer & 72.39 & 0.66 & 0.72 & 0.68 & 0.93 \\
TabTransformer & 73.37 & 0.69 & 0.74 & 0.71 & 0.96 \\
TabSeq        & 73.57 & 0.70 & 0.74 & 0.71 & 0.95 \\
\bottomrule
\end{tabular}
%\vspace{-0.6em}
\end{wraptable}
%%%%%
\paragraph{Findings.}
(1) \emph{Ensemble dominance.} CatBoost/RF remain strong on heterogeneous tabular features.  
(2) \emph{TDL potential.} Attention-based TDL (TabTransformer/TabSeq) are competitive and close the gap with better imbalance handling.  
(3) \emph{Weighted loss helps.} Class-weighted cross-entropy generally boosts Recall/F1, improving minority-class sensitivity without large AUC degradation. 
(4) \emph{Metric trade-offs.} Accuracy alone obscures differences; macro Recall/F1 and AUC are crucial for imbalanced ULC.
%%%%%
\begin{table}[htbp]
\centering
\caption{Performance of different models on ULC (test set).}
\label{tab:all_models}
\begin{adjustbox}{max width=\linewidth}
\tiny
\begin{tabular}{lcccccc}
\toprule
\textbf{Model} & \textbf{Test Acc. (\%)} & \textbf{Precision} & \textbf{Recall} & \textbf{F1} & \textbf{AUC} \\
\midrule
Logistic Regression \citep{rao1973linear} & 70.61 & 0.70 & 0.73 & 0.71 & 0.94 \\
Random Forest \citep{breiman2001rf}      & 81.66 & 0.81 & 0.83 & 0.81 & 0.97 \\
SVM  \citep{cortes1995svm}               & 68.44 & 0.67 & 0.71 & 0.68 & 0.96 \\
Decision Tree \citep{quinlan1986id3}      & 74.75 & 0.72 & 0.75 & 0.73 & 0.96 \\
KNN  \citep{cover1967knn}               & 70.22 & 0.69 & 0.69 & 0.69 & 0.93 \\
Naive Bayes \citep{duda1973pattern}     & 77.51 & 0.76 & 0.78 & 0.76 & 0.96 \\
MLP \citep{rumelhart1986backprop}                & 69.63 & 0.65 & 0.70 & 0.67 & 0.97 \\
\textbf{CatBoost} \citep{prokhorenkova2018catboost}   & \textbf{82.25} & 0.81 & 0.82 & 0.82 & 0.98 \\
AdaBoost \citep{freund1997adaboost}           & 63.12 & 0.67 & 0.62 & 0.64 & 0.93 \\
XGBoost  \citep{chen2016xgboost}           & 68.05 & 0.69 & 0.64 & 0.64 & 0.95 \\
GB   \citep{friedman2001gbm}               & 72.78 & 0.70 & 0.72 & 0.71 & 0.93 \\
1D CNN  \citep{lecun1998document}            & 73.37 & 0.73 & 0.75 & 0.75 & 0.97 \\
TabNet  \citep{arik2021tabnet}            & 64.10 & 0.57 & 0.60 & 0.57 & 0.91 \\
FT-Transformer \citep{gorishniy2021revisiting}     & 73.18 & 0.70 & 0.74 & 0.70 & 0.93 \\
TabTransformer \citep{huang2020tabtransformer}     & 74.75 & 0.71 & 0.72 & 0.72 & 0.96 \\
TabSeq \citep{habib2024tabseq}              & 73.96 & 0.70 & 0.74 & 0.74 & 0.95 \\
\bottomrule
\end{tabular}
\end{adjustbox}
\end{table}
%%%%
% Main text figure (single panel)
\begin{figure}[htbp]
\centering
\includegraphics[width=0.75\linewidth]{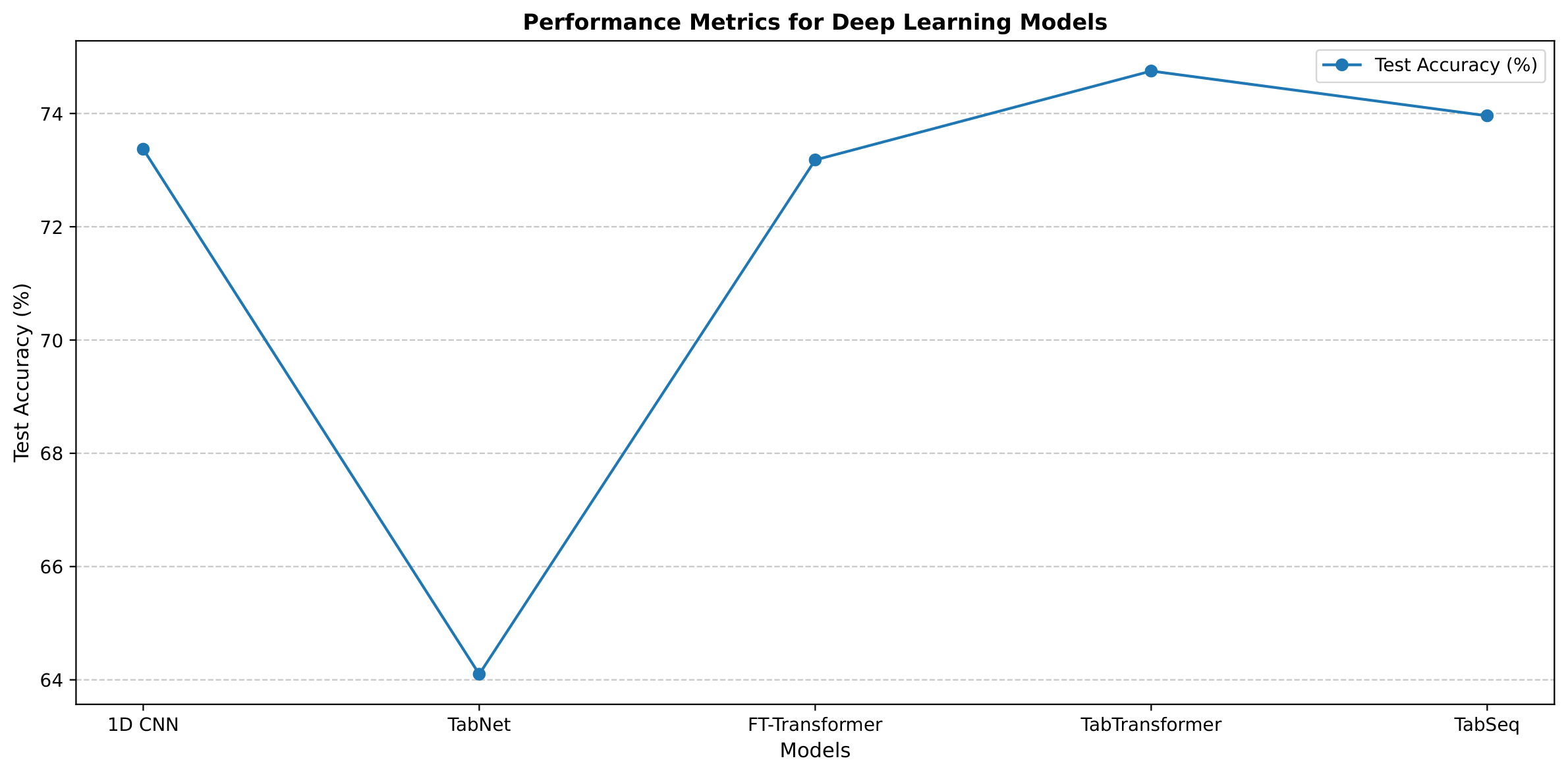}
\caption{Deep-model accuracies (TDL only). Remaining visualizations appear in Appendix~\ref{app:morefigs}.}
\label{fig:dl_acc}
\end{figure}
%%%%
\paragraph{Statistical comparison of model accuracies.}
Table \ref{tab:all_models} reports single-run test accuracies for all baselines, and Figure \ref{fig:ulc-accuracy} visualizes the same results with 95\% binomial confidence intervals computed from the test-set size. Because the intervals for many methods overlap, the evidence for large performance gaps on this single split is limited. Even so, the plot indicates that CatBoost, Random Forest, and Naive Bayes form the strongest group, with CatBoost achieving the highest observed accuracy. Since per-instance predictions or multiple randomized splits were not available, we used an approximate two-proportion (unpaired) comparison rather than a paired test such as McNemar or Friedman.
%%%% statistics figure
%%%%
\paragraph{Future direction.}
For future work, we will extend this study to a broader set of tabular benchmarks and develop a tabular Transformer-style model for remote-sensing and flood-related tabular datasets to better handle heterogeneous features and limited labels. This direction directly addresses the current single-dataset and novelty concerns and further clarifies the role of tabular deep learning for ULC. The same pipeline can also be applied in Muslim-majority or rapidly urbanizing cities (e.g., Dhaka, Jakarta, Lahore, Cairo), where recent urban land-cover products are scarce.
%%%%%
\section{Conclusion}
We evaluated robust classical ML models alongside newer TDL techniques for ULC classification using the UCI ULC dataset. Tree ensembles especially CatBoost and Random Forest achieved the highest accuracy and competitive macro metrics, reaffirming their suitability for heterogeneous, imbalanced tabular features. Nevertheless, TDL models (TabTransformer, TabSeq) were competitive and improved further with class-weighted losses, indicating clear potential when representation learning and imbalance handling are aligned with data characteristics. Practically, we recommend ensembles as robust baselines and TDL as complementary models when non-linear feature interactions and minority-class sensitivity are priorities. Limitations include a single-dataset scope and modest sample size; future work will expand to additional ULC benchmarks, explore stronger calibration/uncertainty modeling, and refine TDL architectures and sampling/weighting strategies to close the remaining gap.

\section*{Acknowledgments}
We thank Dr.~Aaron Maxwell, Associate Professor in the Department of Geology and Geography at West Virginia University, USA, for his insightful feedback on this project. See his webpage: \href{https://wvview.org/maxwell.html}{https://www.wvview.org/maxwell.html}.

\bibliographystyle{unsrtnat}   % at end
\bibliography{ref.bib}
\appendix
\section{Technical Appendices and Supplementary Material}
\label{sec:appendix_tech_details}

This supplementary material accompanies our main paper entitled 
\textit{Tabular Deep Learning vs Classical Machine Learning for Urban Land Cover Classification} 
(submitted to the 5\textsuperscript{th} Muslims In ML (MusIML) Workshop co-located with NeurIPS~2025). 
It provides additional experimental results that complement the main text. 

%%%%%%%%%%%%%%%%%%%%%%%%%%%%%%%%%%%%%%%%%%%%%%%%%%%%%%%%%%%%

\newpage
\section*{NeurIPS Paper Checklist}

\begin{enumerate}
\item \textbf{Claims}
    \item[] \textbf{Question:} Do the main claims made in the abstract and introduction accurately reflect the paper's contributions and scope?
    \item[] \textbf{Answer:} \answerYes{}
    \item[] \textbf{Justification:} \justification{The abstract and introduction state that the paper (i) benchmarks strong classical ML models against Tabular Deep Learning (TDL) models on the UCI Urban Land Cover (ULC) dataset, (ii) uses a unified, reproducible pipeline, and (iii) analyzes metric-level differences (accuracy, macro F1, AUC). These are exactly the contributions developed in the main sections, without overclaiming generalization to large multimodal or multisensor settings.}

\item \textbf{Limitations}
    \item[] \textbf{Question:} Does the paper discuss the limitations of the work performed by the authors?
    \item[] \textbf{Answer:} \answerYes{}
    \item[] \textbf{Justification:} \justification{We explicitly note that experiments are conducted on a single public UCI ULC dataset with a modest test split, and that we do not report paired significance tests because per-instance predictions or multiple randomized splits were not available. We also mention that future work will extend the study to additional remote-sensing and flood-related tabular benchmarks and stronger TDL variants.}

\item \textbf{Theory assumptions and proofs}
    \item[] \textbf{Question:} For each theoretical result, does the paper provide the full set of assumptions and a complete (and correct) proof?
    \item[] \textbf{Answer:} \answerNA{}
    \item[] \textbf{Justification:} \justification{This paper is empirical/benchmarking in nature (TDL vs classical ML for ULC) and does not introduce new theoretical results or formal proofs. We therefore state the modeling assumptions (dataset, metrics, splits, baselines) but no completeness of proofs is required.}

\item \textbf{Experimental result reproducibility}
    \item[] \textbf{Question:} Does the paper fully disclose all the information needed to reproduce the main experimental results of the paper to the extent that it affects the main claims and/or conclusions of the paper (regardless of whether the code and data are provided or not)?
    \item[] \textbf{Answer:} \answerYes{}
    \item[] \textbf{Justification:} \justification{The paper specifies the dataset (UCI ULC), the train/validation/test split, the set of classical ML and TDL baselines, the evaluation metrics (accuracy, macro precision/recall/F1, AUC), and the main training settings for TDL models; it also provides the public GitHub link for code and scripts, enabling independent reproduction of the reported tables and plots.}

\item \textbf{Open access to data and code}
    \item[] \textbf{Question:} Does the paper provide open access to the data and code, with sufficient instructions to faithfully reproduce the main experimental results, as described in supplemental material?
    \item[] \textbf{Answer:} \answerYes{}
    \item[] \textbf{Justification:} \justification{The dataset used in this study (UCI ULC) is publicly available, and we provide an open GitHub repository containing the training/evaluation scripts and configuration files needed to reproduce the reported results: \url{https://github.com/mtesha/tdl-vs-ml-urbanlandcover}.}

\item \textbf{Experimental setting/details}
    \item[] \textbf{Question:} Does the paper specify all the training and test details (e.g., data splits, hyperparameters, how they were chosen, type of optimizer, etc.) necessary to understand the results?
    \item[] \textbf{Answer:} \answerYes{}
    \item[] \textbf{Justification:} \justification{We describe the UCI ULC dataset, the train/validation/test split, preprocessing (scaling using train statistics), the classical ML vs TDL model list, optimizer and loss settings for TDL (Adam, weighted CE), and the use of Optuna for TDL hyperparameter tuning, so the evaluation protocol and reported tables can be fully understood.}

\item \textbf{Experiment statistical significance}
    \item[] \textbf{Question:} Does the paper report error bars suitably and correctly defined or other appropriate information about the statistical significance of the experiments?
    \item[] \textbf{Answer:} \answerYes{}
    \item[] \textbf{Justification:} \justification{We report model accuracies with 95\% binomial confidence intervals and include an approximate two-proportion comparison against the top model on the single test split. We also state the limitation that paired tests (e.g., McNemar, Friedman/Nemenyi) are not applicable without per-instance predictions or multiple randomized splits.}

\item \textbf{Experiments compute resources}
    \item[] \textbf{Question:} For each experiment, does the paper provide sufficient information on the computer resources (type of compute workers, memory, time of execution) needed to reproduce the experiments?
    \item[] \textbf{Answer:} \answerYes{}
    \item[] \textbf{Justification:} \justification{We specify the software stack (scikit-learn, PyTorch/TensorFlow), dataset size and splits, batch size and optimizer settings, and note that all runs are reproducible on a standard workstation (CPU or a single commodity GPU). Given the modest scale of UCI ULC, exact runtime is negligible and not required to reproduce the results.}
    
\item \textbf{Code of ethics}
    \item[] \textbf{Question:} Does the research conducted in the paper conform, in every respect, with the NeurIPS Code of Ethics (\url{https://neurips.cc/public/EthicsGuidelines})?
    \item[] \textbf{Answer:} \answerYes{}
    \item[] \textbf{Justification:} \justification{The work uses a public, non-PII remote-sensing/tabular dataset (UCI ULC), cites it properly, and targets urban/environmental analytics with no foreseeable harmful use. No human subjects, private data, or sensitive attributes are involved, and results are reported transparently.}

\item \textbf{Broader impacts}
    \item[] \textbf{Question:} Does the paper discuss both potential positive societal impacts and negative societal impacts of the work performed?
    \item[] \textbf{Answer:} \answerYes{}
    \item[] \textbf{Justification:} \justification{Our study targets ULC classification from tabular/remote sensing descriptors, which can support urban planning, environmental monitoring, and service provision in data-scarce cities, including Muslim-majority and rapidly urbanizing regions. At the same time, any automated land/asset mapping pipeline could be misused for inequitable resource allocation or area-level surveillance if applied without transparency or local validation, so we recommend responsible, policy-aware deployment.}
    
\item \textbf{Safeguards}
    \item[] \textbf{Question:} Does the paper describe safeguards that have been put in place for responsible release of data or models that have a high risk for misuse (e.g., pretrained language models, image generators, or scraped datasets)?
    \item[] \textbf{Answer:} \answerNA{}
    \item[] \textbf{Justification:} \justification{We use a small, public, non-sensitive UCI ULC dataset and release task-specific tabular classification code; no large pretrained or generative models, scraped data, or high-risk assets are introduced, so additional safeguards are not required.}

\item \textbf{Licenses for existing assets}
    \item[] \textbf{Question:} Are the creators or original owners of assets (e.g., code, data, models), used in the paper, properly credited and are the license and terms of use explicitly mentioned and properly respected?
    \item[] \textbf{Answer:} \answerYes{}
    \item[] \textbf{Justification:} \justification{We use the publicly available UCI ULC dataset and standard open-source libraries (scikit-learn, PyTorch), all of which are properly cited and used within their research/OSS licenses. We do not redistribute modified versions of proprietary assets.}

\item \textbf{New assets}
    \item[] \textbf{Question:} Are new assets introduced in the paper well documented and is the documentation provided alongside the assets?
    \item[] \textbf{Answer:} \answerYes{}
    \item[] \textbf{Justification:} \justification{We release a lightweight, documented codebase for reproducing the TDL vs classical ML experiments on the UCI ULC dataset (data loading, preprocessing, model configs, and evaluation). No new dataset is introduced; the only new asset is the reproducible pipeline, which is described in the paper and in the repository.}

\item \textbf{Crowdsourcing and research with human subjects}
    \item[] \textbf{Question:} For crowdsourcing experiments and research with human subjects, does the paper include the full text of instructions given to participants and screenshots, if applicable, as well as details about compensation (if any)?
    \item[] \textbf{Answer:} \answerNA{}
    \item[] \textbf{Justification:} \justification{No crowdsourcing or human-subject data collection was conducted; we only used a public remote-sensing/tabular dataset.}

\item \textbf{Institutional review board (IRB) approvals or equivalent for research with human subjects}
    \item[] \textbf{Question:} Does the paper describe potential risks incurred by study participants, whether such risks were disclosed to the subjects, and whether Institutional Review Board (IRB) approvals (or an equivalent approval/review based on the requirements of your country or institution) were obtained?
    \item[] \textbf{Answer:} \answerNA{}
    \item[] \textbf{Justification:} \justification{The work uses only publicly available, non-PII remote-sensing/tabular data and does not involve human subjects, so IRB review was not required.}

\item \textbf{Declaration of LLM usage}
    \item[] \textbf{Question:} Does the paper describe the usage of LLMs if it is an important, original, or non-standard component of the core methods in this research?
    \item[] \textbf{Answer:} \answerNA{}
    \item[] \textbf{Justification:} \justification{LLMs were used only for minor language polishing and LaTeX formatting; they did not contribute to the core methodology, experiments, or results, so a formal declaration is not required.}

\end{enumerate}

\newpage
\renewcommand{\thesection}{A}
\renewcommand{\thesubsection}{A.\arabic{subsection}}
\section{Technical Appendices and Supplementary Material}
\renewcommand{\thesection}{\arabic{section}}
This supplementary document supports our main paper titled \textit{Tabular Deep Learning vs Classical Machine Learning for Urban Land Cover Classification} (Submitted to the 5\textsuperscript{th}  Muslims In ML (MusIML) Workshop co-located with NeurIPS~2025). Specifically, it includes:
\begin{itemize}
    \item Additional Figures for Comparative Results
    \item ROC Curves and Confusion Matrices for TDL Models with Weighted Cross Entropy
    \item ROC Curves and Confusion Matrices for All Models with Standard Cross Entropy
\end{itemize}

\subsection{Additional Figures for Comparative Results}
\label{app:morefigs}
\noindent\textbf{Fig.~\ref{fig:acc_all_app} (All-model accuracy).}
Overall leaderboard across 16 methods; tree ensembles (CatBoost, RF) top the chart, while most TDL models sit mid-pack.

\medskip
\noindent\textbf{Fig.~\ref{fig:radar_top_app} (Top-model multi-metric).}
Across precision/recall/F1/AUC, CatBoost and RF remain consistently strong with balanced precision-recall; TabTransformer trails slightly on F1 but is competitive on AUC.

\medskip
\noindent\textbf{Fig.~\ref{fig:heatmap_all_app} (Full metric heatmap).}
Side-by-side macro metrics reveal where models trade precision for recall; CatBoost/RF are uniformly high, while TabNet and AdaBoost show weaker recall/F1.

\medskip
\noindent\textbf{Fig.~\ref{fig:weighted_acc_app} (TDL with weighted CE accuracy).}
Adding class weights narrows gaps among deep models; TabSeq/TabTransformer edge out FT-Transformer and 1D-CNN, with TabNet still behind.

\medskip
\noindent\textbf{Fig.~\ref{fig:weighted_metrics_app} (TDL with weighted CE multi-metric).}
Weighted loss mainly boosts recall and F1 while maintaining high AUC, indicating better minority-class sensitivity without sacrificing ranking quality.

% Appendix: separate figures
\renewcommand{\thefigure}{A\arabic{figure}}
\begin{figure*}[htbp]
\centering
\includegraphics[width=0.9\linewidth]{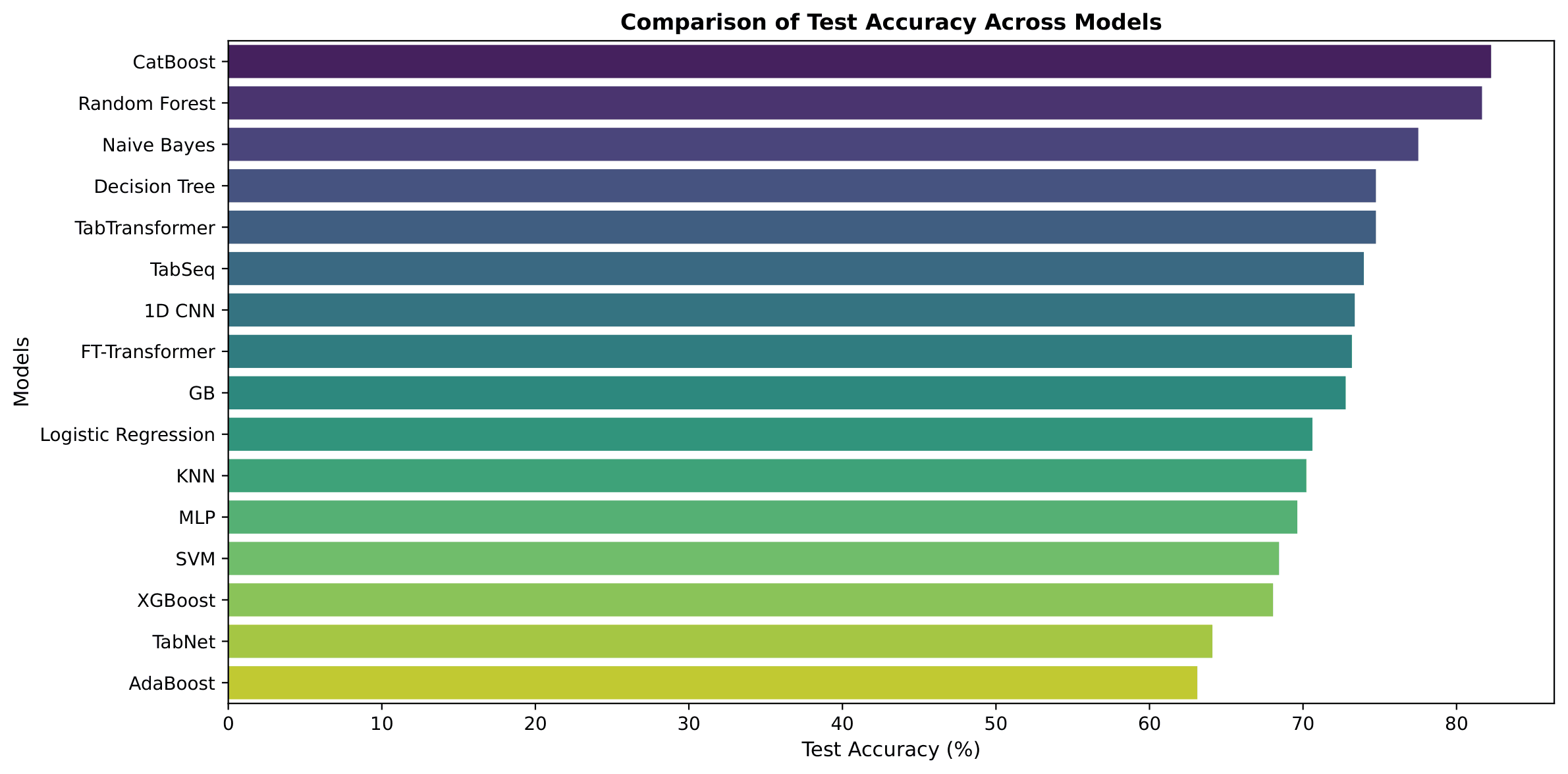}
\caption{All models test accuracy.}
\label{fig:acc_all_app}
\end{figure*}

\begin{figure*}[htbp]
\centering
\includegraphics[width=0.9\linewidth]{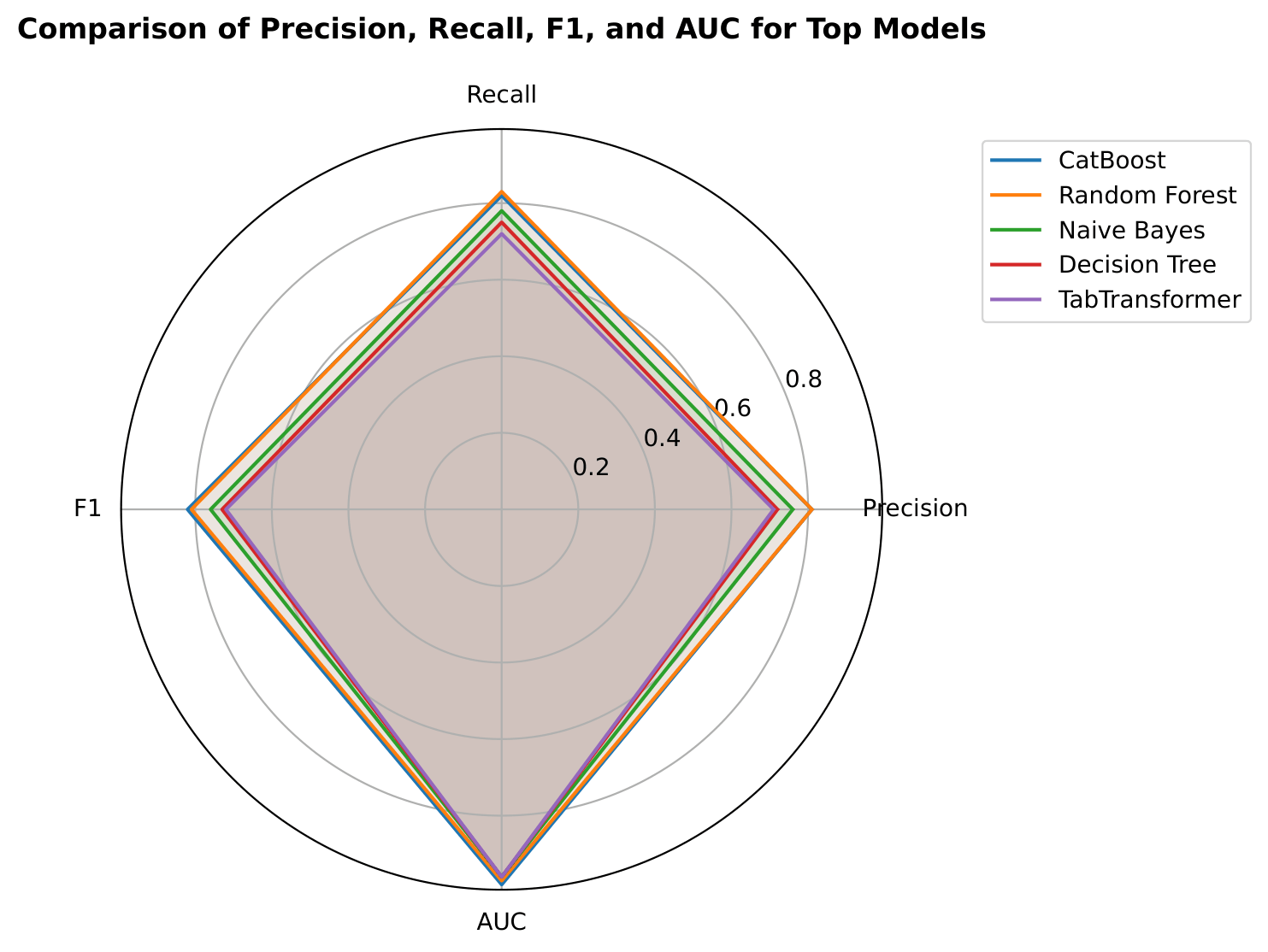}
\caption{Top models: Precision/Recall/F1/AUC.}
\label{fig:radar_top_app}
\end{figure*}

\begin{figure*}[htbp]
\centering
\includegraphics[width=0.9\linewidth]{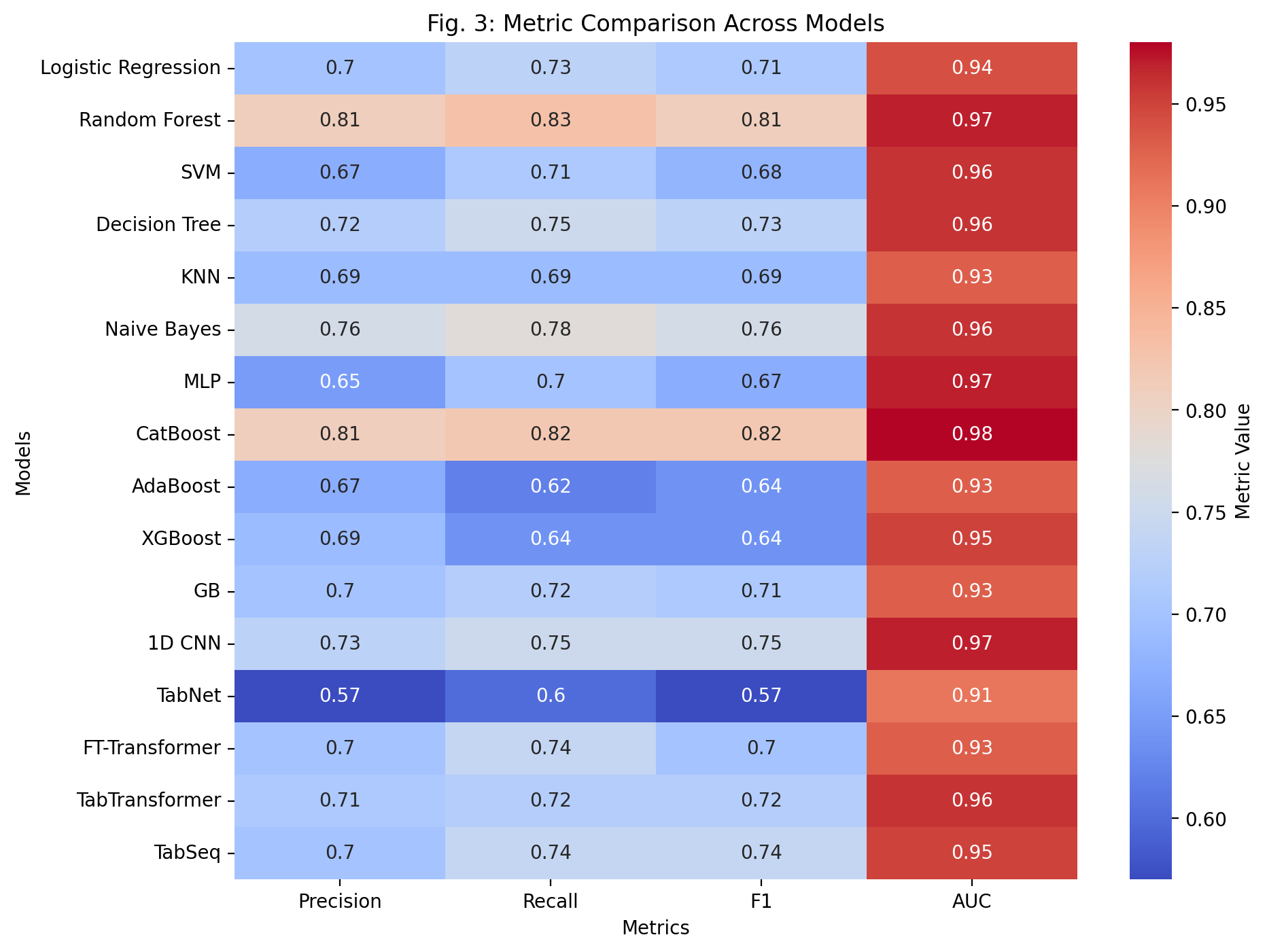}
\caption{Metric heatmap for all models.}
\label{fig:heatmap_all_app}
\end{figure*}

\begin{figure*}[htbp]
\centering
\includegraphics[width=0.9\linewidth]{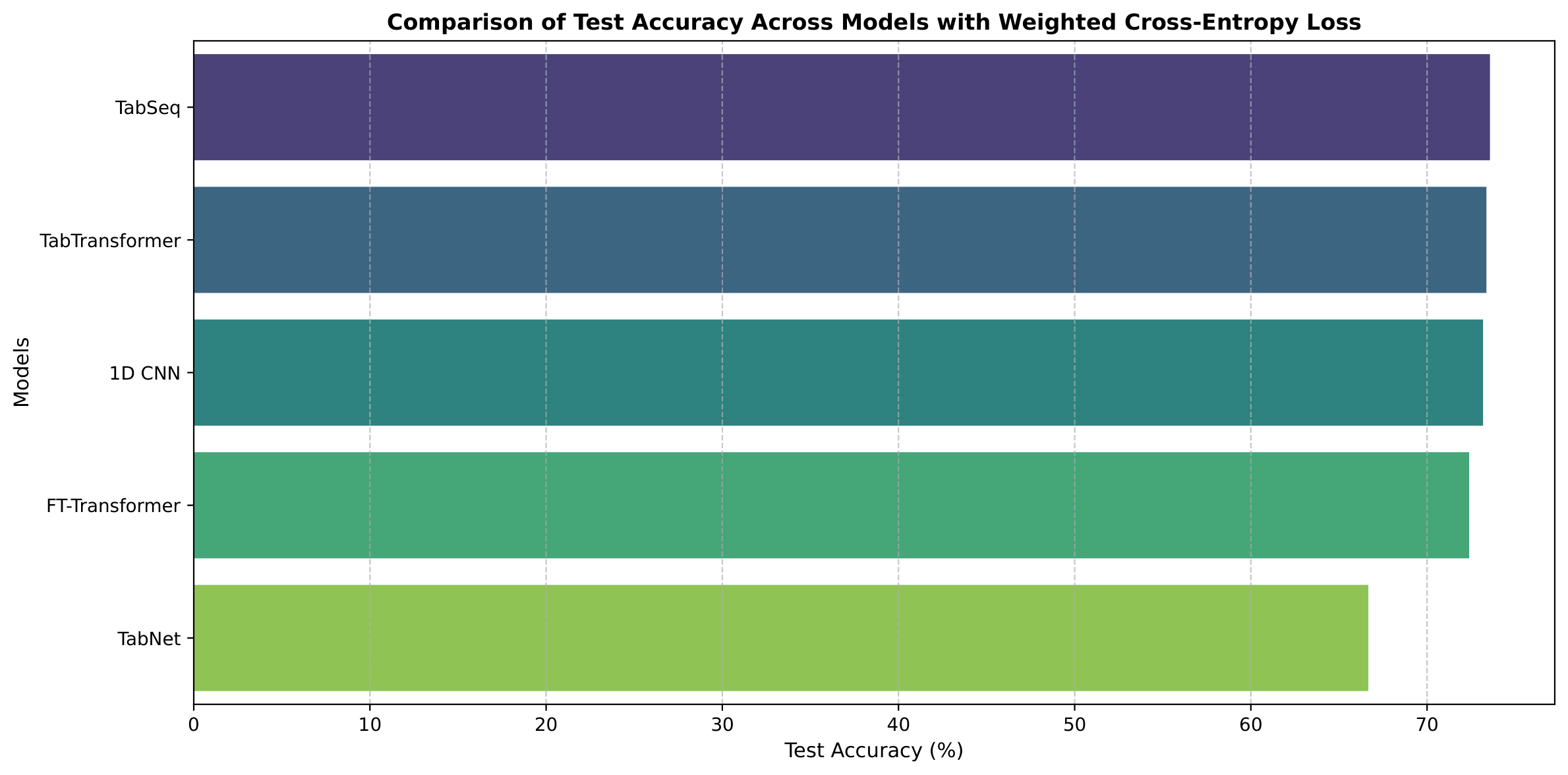}
\caption{Weighted CE (TDL): test accuracy.}
\label{fig:weighted_acc_app}
\end{figure*}

\begin{figure}[htbp]
\centering
\includegraphics[width=0.9\linewidth]{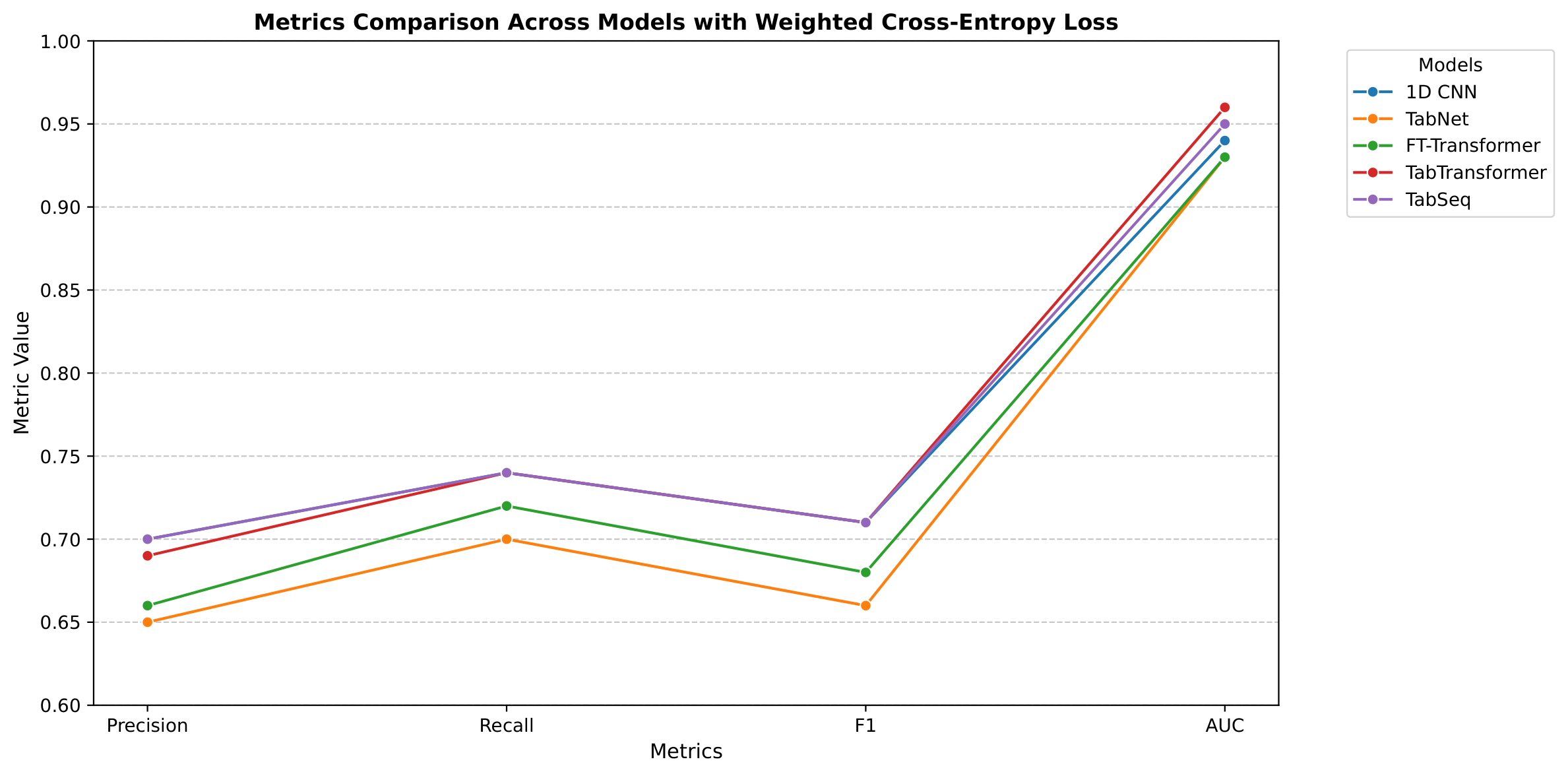}
\caption{Effect of class-weighted cross-entropy on multi-metric performance (TDL).}
\label{fig:weighted_metrics_app}
\end{figure}

\subsection{ROC Curves and Confusion Matrices for TDL Models with Weighted Cross Entropy}
\label{wad}
\begin{figure}[htbp]
\centering
\begin{subfigure}{0.48\linewidth}
  \centering
  \includegraphics[width=\linewidth]{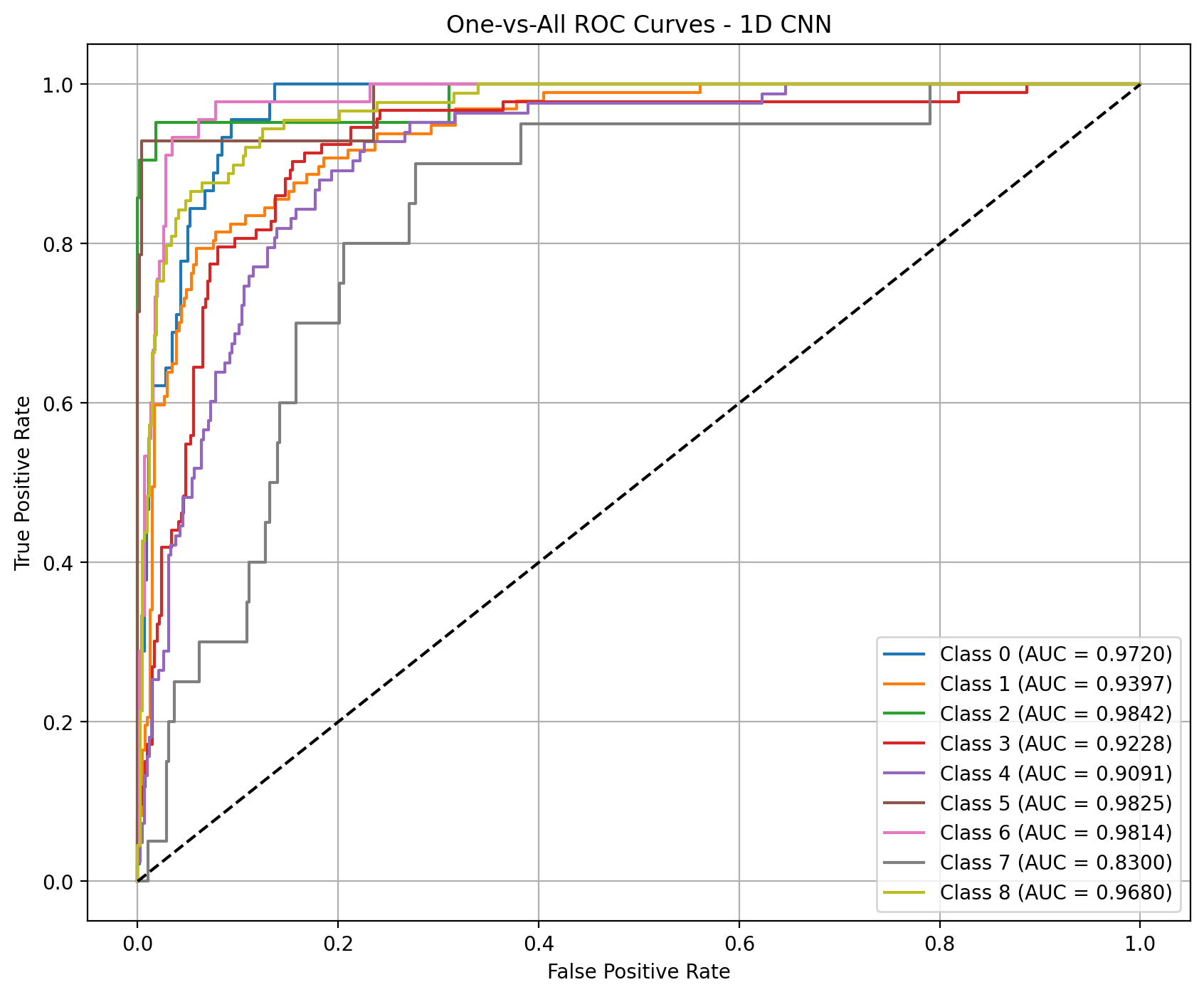}
  \caption{ROC (1D CNN, W)}
  \label{fig:roc_cnn_w}
\end{subfigure}\hfill
\begin{subfigure}{0.48\linewidth}
  \centering
  \includegraphics[width=\linewidth]{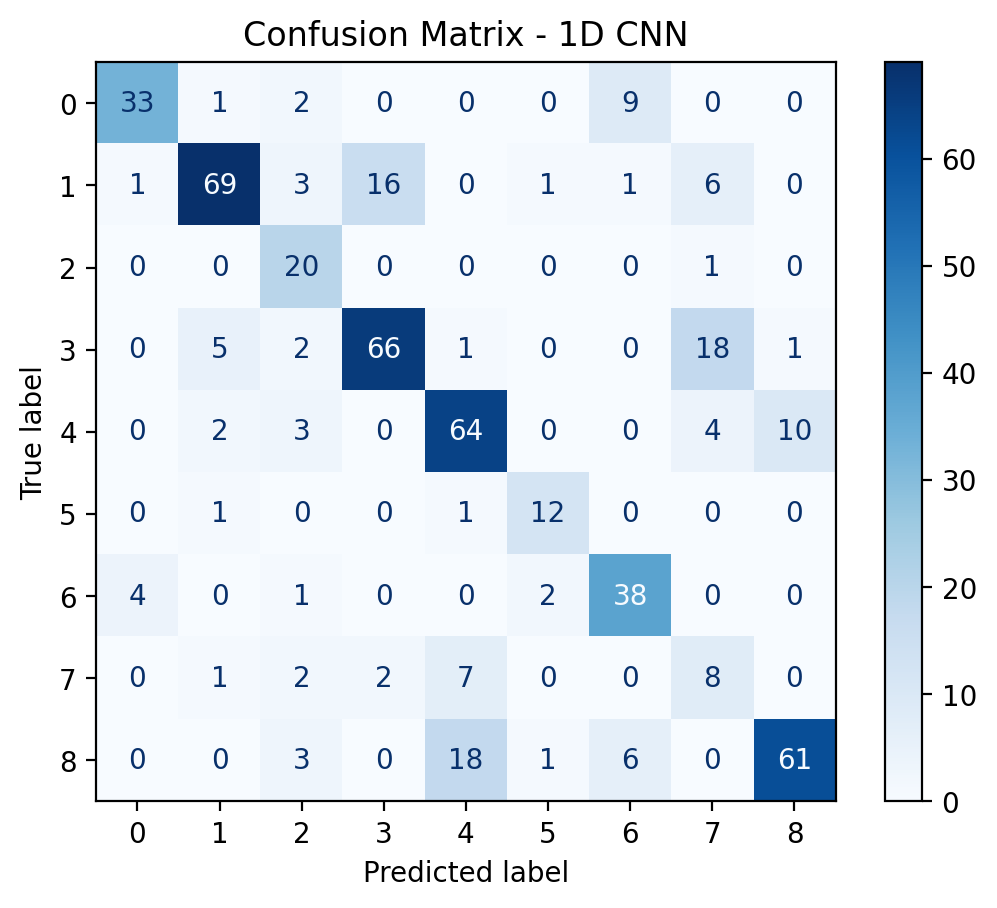}
  \caption{Confusion matrix (1D CNN, W)}
  \label{fig:cm_cnn_w}
\end{subfigure}
\caption{One-vs-all ROC and confusion matrix for 1D CNN with weighted cross-entropy.}
\label{fig:pair_cnn_w}
\end{figure}

\noindent\textbf{Figure~\ref{fig:pair_cnn_w} (1D CNN, W).}
ROC curves cluster near the top-left, indicating strong separability for most classes; macro AUC is high with only a few classes showing shallower curves. The confusion matrix is diagonally dominant, with residual errors dispersed across a handful of visually similar categories. Weighted loss improves recall for minority classes without noticeably degrading well-separated ones.

\begin{figure}[htbp]
\centering
\begin{subfigure}{0.48\linewidth}
  \centering
  \includegraphics[width=\linewidth]{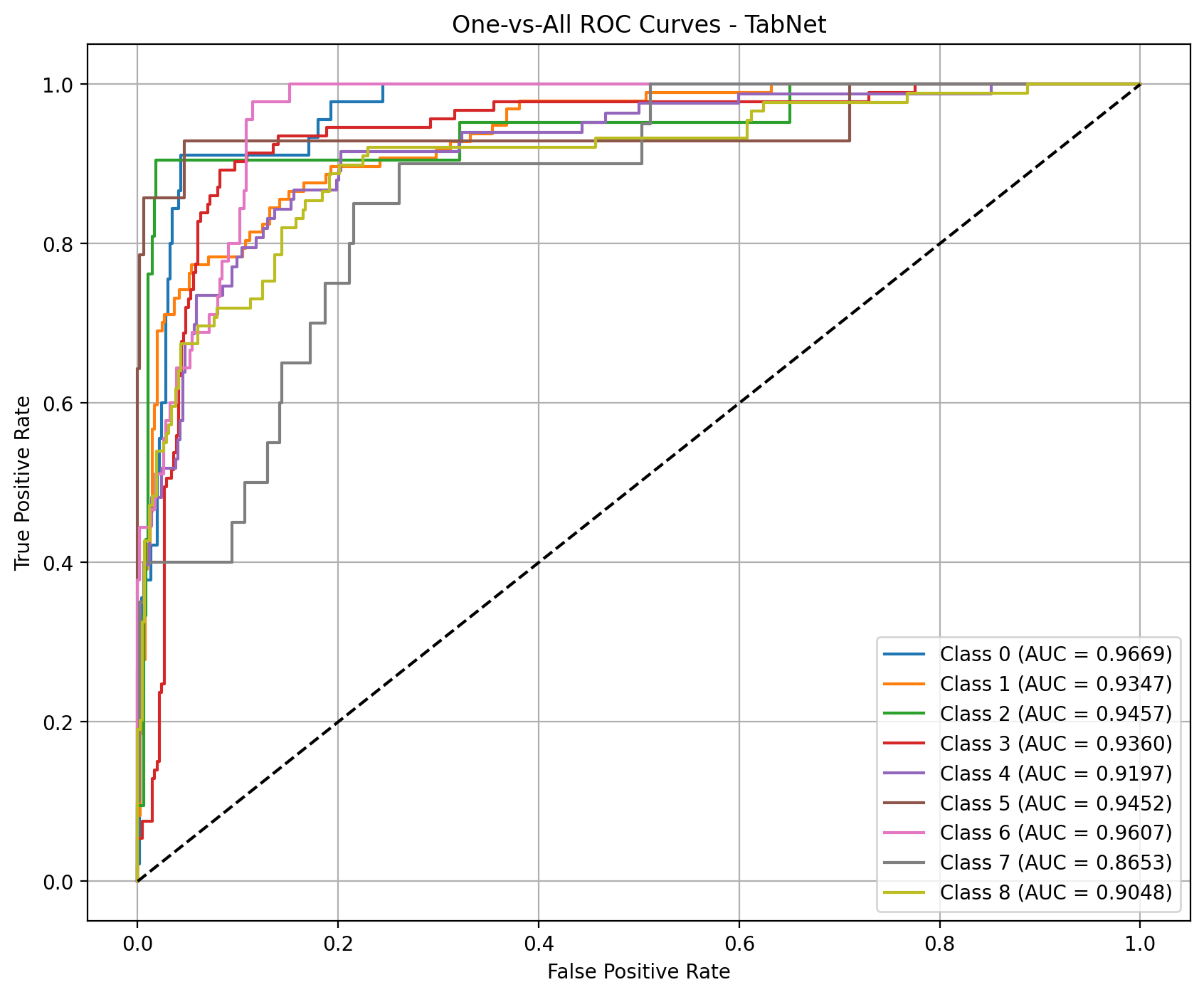}
  \caption{ROC (TabNet, W)}
  \label{fig:roc_tabnet_w}
\end{subfigure}\hfill
\begin{subfigure}{0.48\linewidth}
  \centering
  \includegraphics[width=\linewidth]{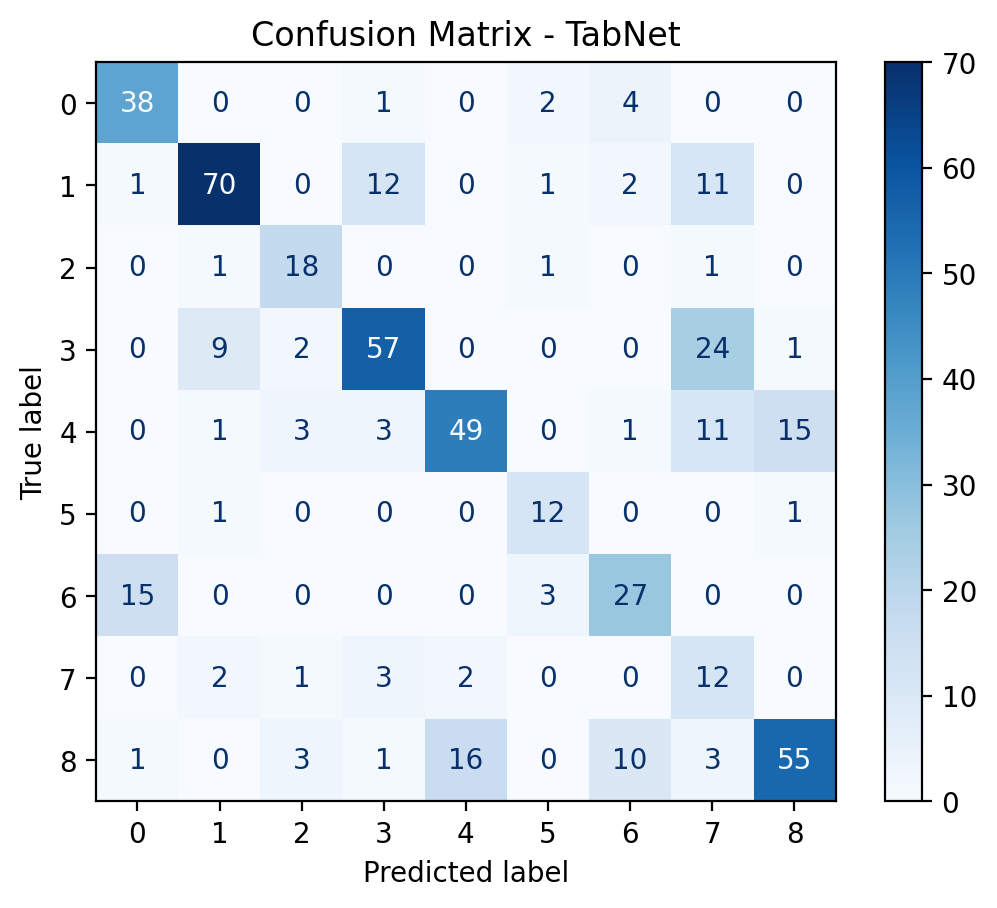}
  \caption{Confusion matrix (TabNet, W)}
  \label{fig:cm_tabnet_w}
\end{subfigure}
\caption{One-vs-all ROC and confusion matrix for TabNet with weighted cross-entropy.}
\label{fig:pair_tabnet_w}
\end{figure}

\medskip
\noindent\textbf{Figure~\ref{fig:pair_tabnet_w} (TabNet, W).}
ROC profiles are competitive overall but exhibit slightly flatter segments for some classes, consistent with greater sensitivity to class imbalance. The confusion matrix shows more off-diagonal mass than 1D~CNN, reflecting occasional swaps among related built-surface or vegetation classes. Attention-based feature selection helps, but errors concentrate in harder pairs.

\begin{figure}[htbp]
\centering
\begin{subfigure}{0.48\linewidth}
  \centering
  \includegraphics[width=\linewidth]{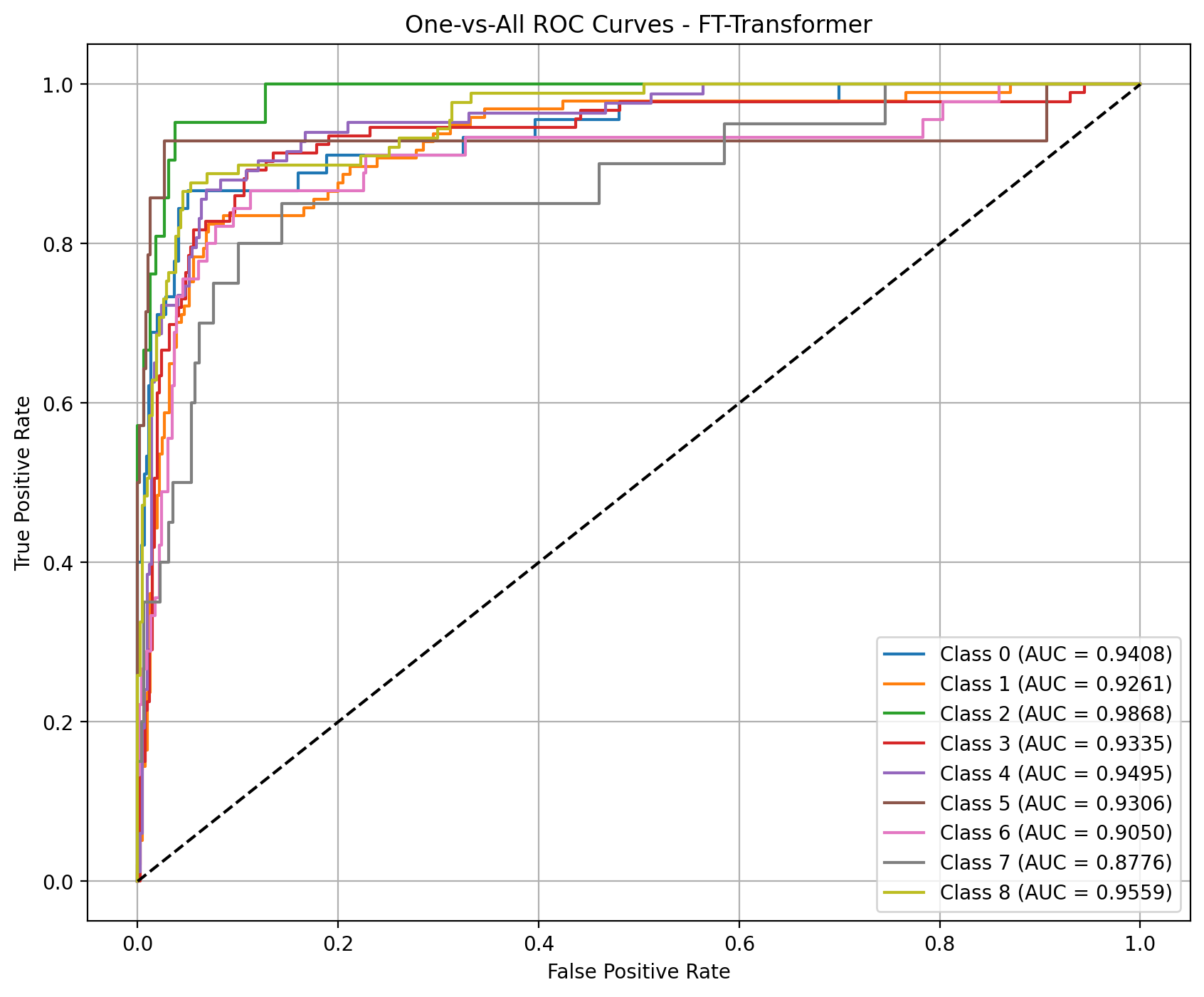}
  \caption{ROC (FT-Transformer, W)}
  \label{fig:roc_ft_w}
\end{subfigure}\hfill
\begin{subfigure}{0.48\linewidth}
  \centering
  \includegraphics[width=\linewidth]{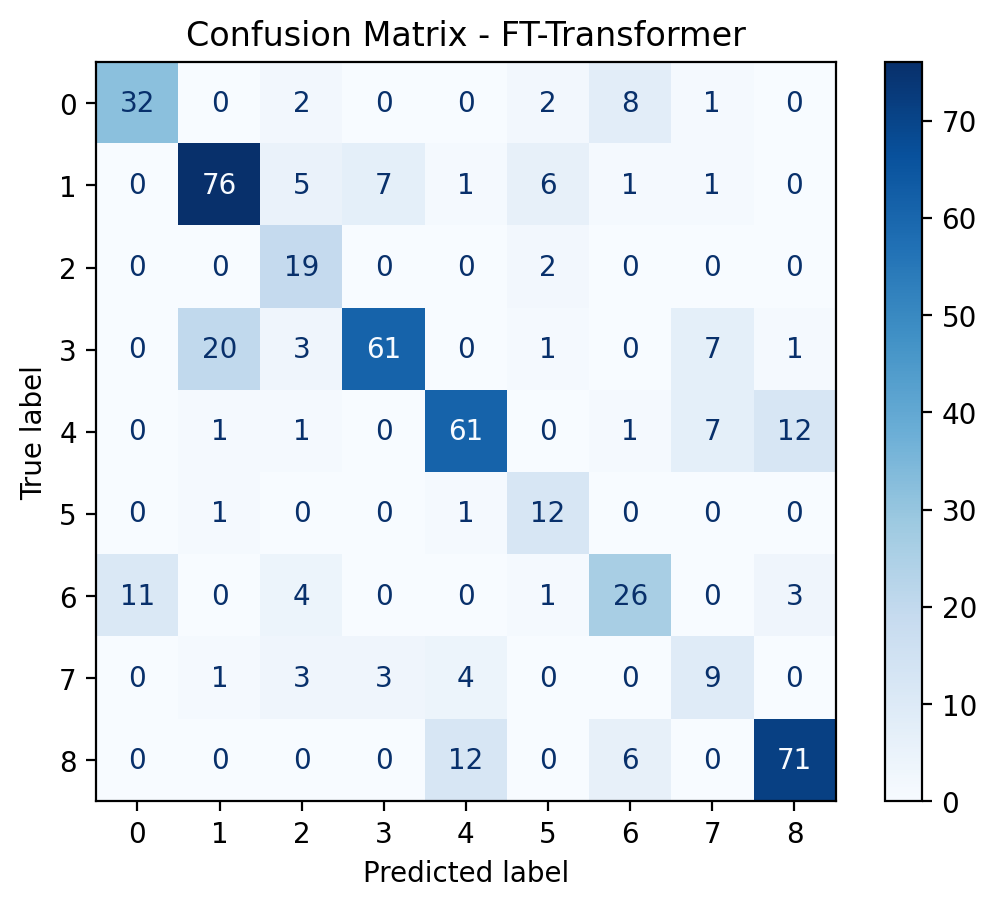}
  \caption{Confusion matrix (FT-Transformer, W)}
  \label{fig:cm_ft_w}
\end{subfigure}
\caption{One-vs-all ROC and confusion matrix for FT-Transformer with weighted cross-entropy.}
\label{fig:pair_ft_w}
\end{figure}

\medskip
\noindent\textbf{Figure~\ref{fig:pair_ft_w} (FT-Transformer, W).}
Transformer-based representations yield uniformly strong ROC curves with tight variance across classes. The confusion matrix remains largely diagonal, with misclassifications concentrated in a few class pairs suggesting remaining ambiguity in descriptors rather than systematic bias. Weighted CE chiefly raises recall on the rarer classes.

\begin{figure}[htbp]
\centering
\begin{subfigure}{0.48\linewidth}
  \centering
  \includegraphics[width=\linewidth]{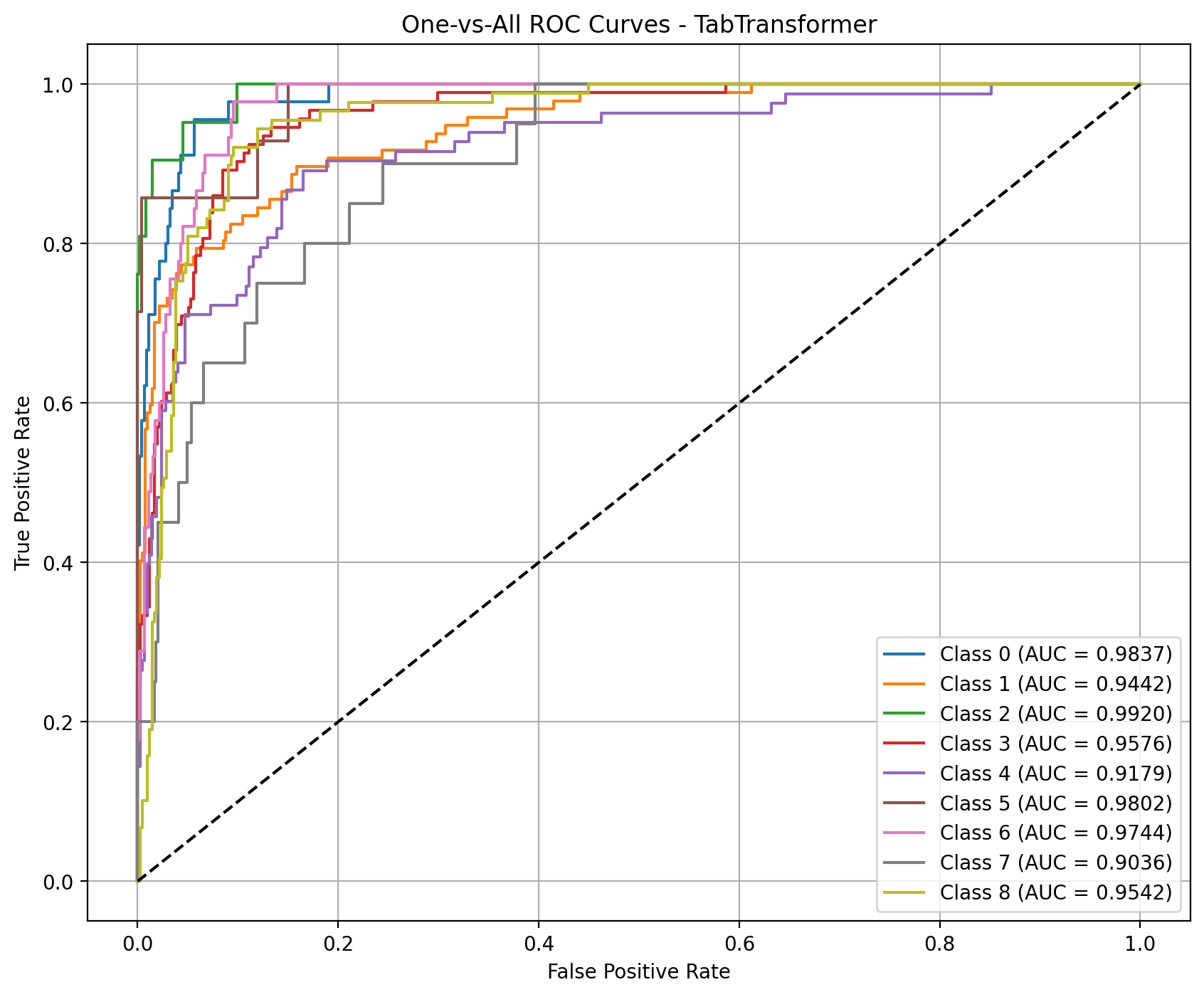}
  \caption{ROC (TabTransformer, W)}
  \label{fig:roc_tabt_w}
\end{subfigure}\hfill
\begin{subfigure}{0.48\linewidth}
  \centering
  \includegraphics[width=\linewidth]{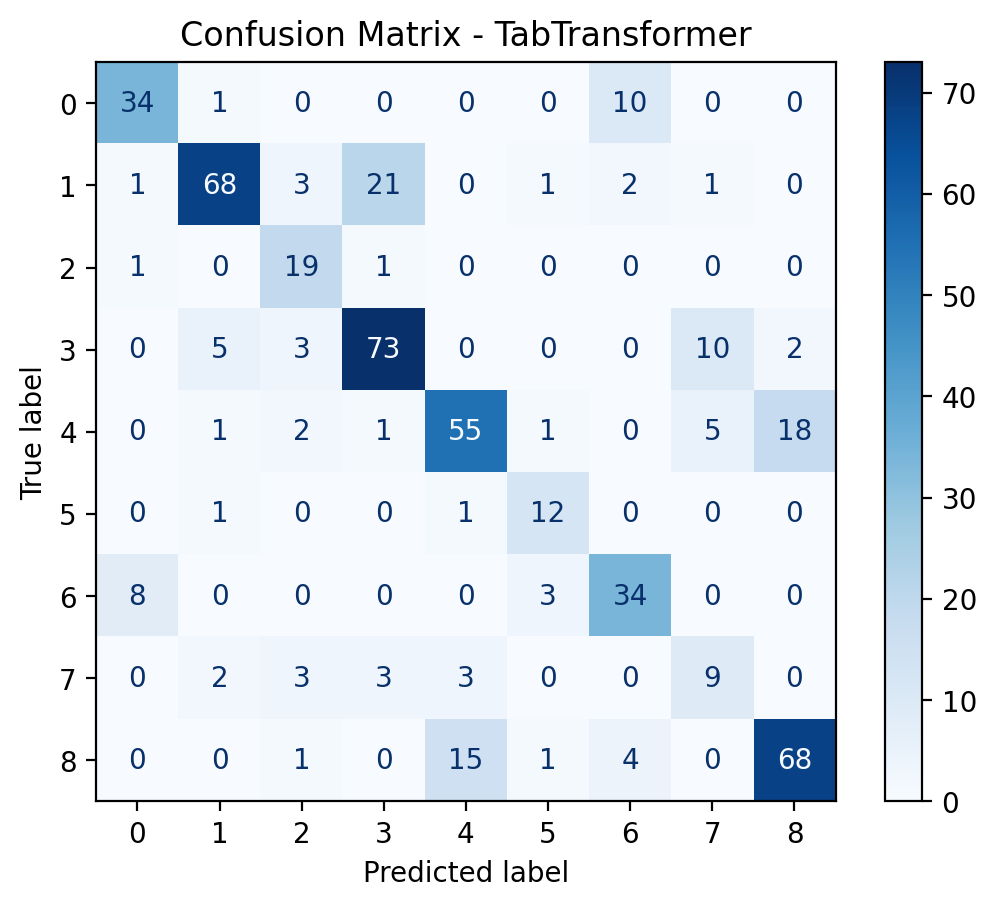}
  \caption{Confusion matrix (TabTransformer, W)}
  \label{fig:cm_tabt_w}
\end{subfigure}
\caption{One-vs-all ROC and confusion matrix for TabTransformer with weighted cross-entropy.}
\label{fig:pair_tabt_w}
\end{figure}

\medskip
\noindent\textbf{Figure~\ref{fig:pair_tabt_w} (TabTransformer, W).}
This model achieves some of the steepest ROC traces across classes, pointing to effective modeling of feature interactions. The confusion matrix shows high true-positive counts along the diagonal and fewer large off-diagonal cells, indicating balanced performance; remaining errors align with classes that are spectrally/structurally similar.

\begin{figure}[htbp]
\centering
\begin{subfigure}{0.48\linewidth}
  \centering
  \includegraphics[width=\linewidth]{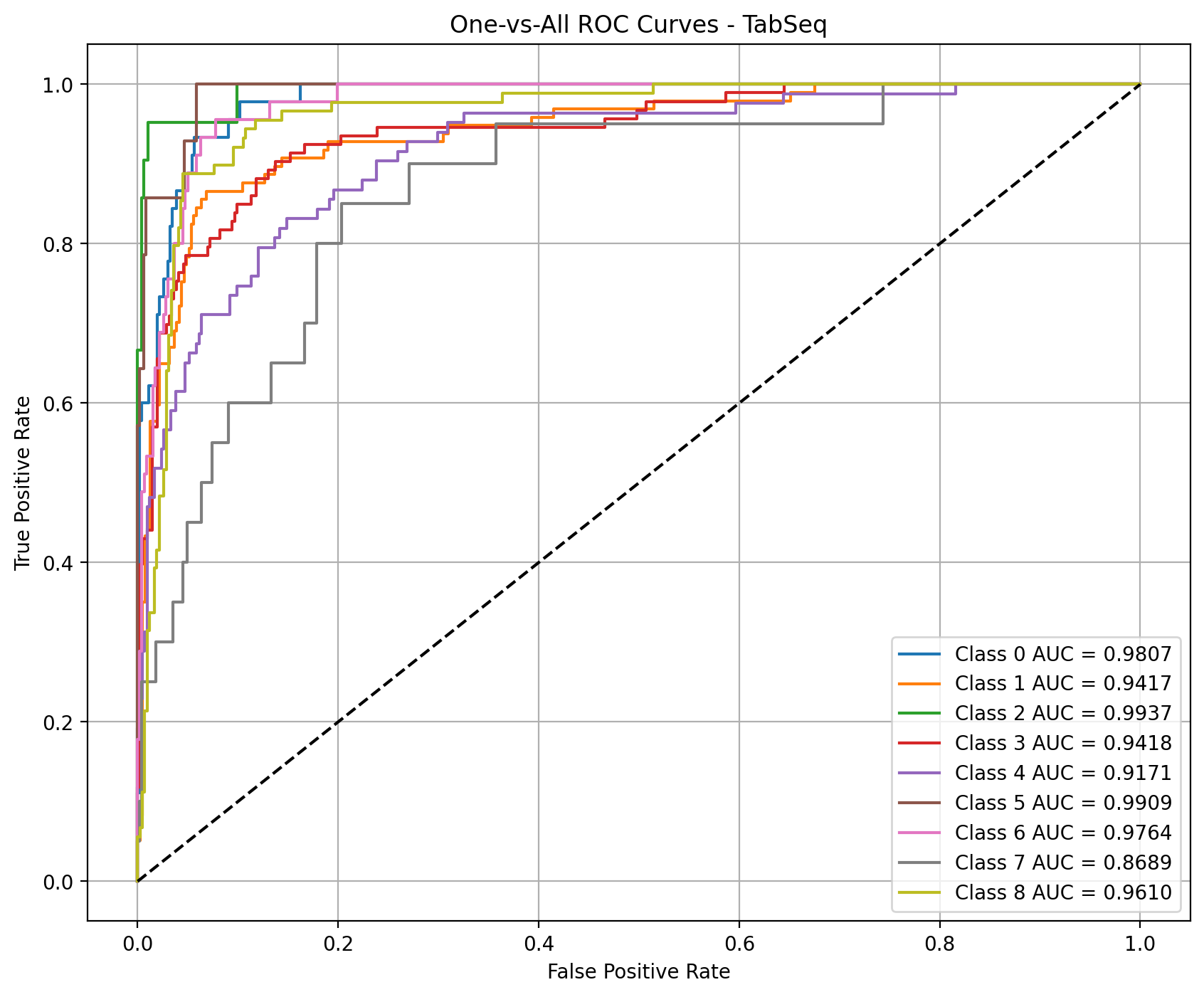}
  \caption{ROC (TabSeq, W)}
  \label{fig:roc_tabseq_w}
\end{subfigure}\hfill
\begin{subfigure}{0.48\linewidth}
  \centering
  \includegraphics[width=\linewidth]{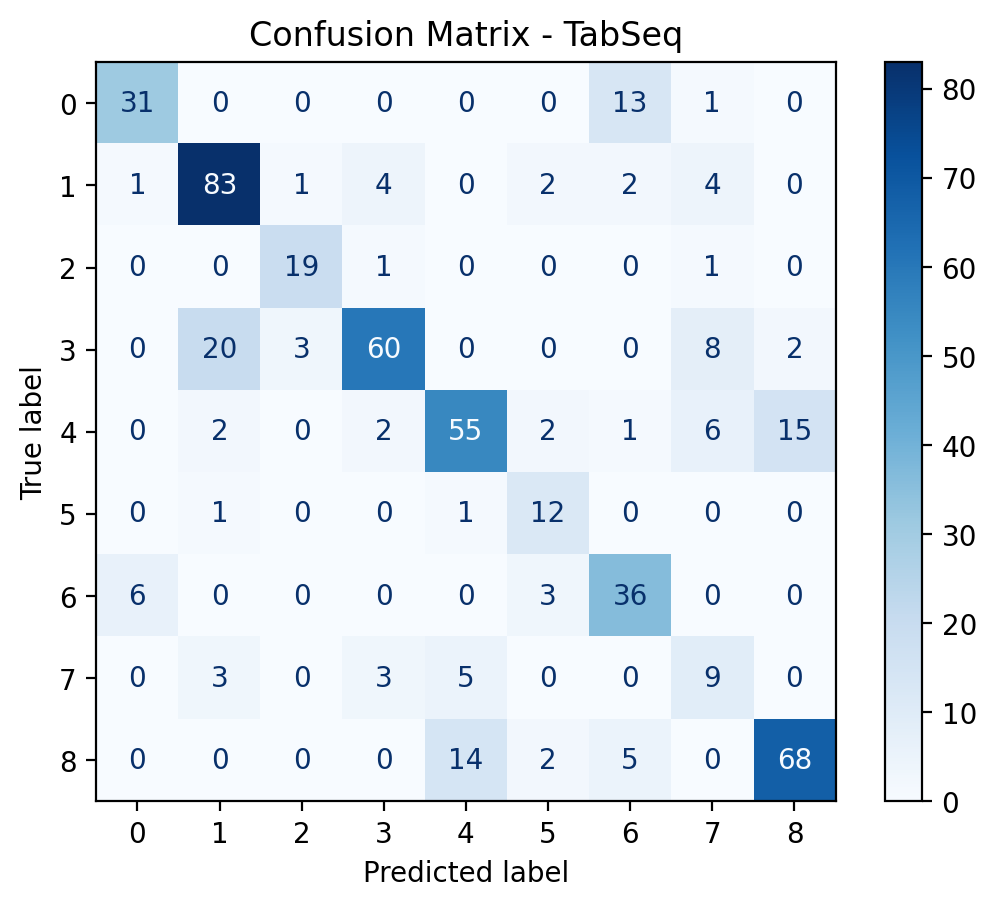}
  \caption{Confusion matrix (TabSeq, W)}
  \label{fig:cm_tabseq_w}
\end{subfigure}
\caption{One-vs-all ROC and confusion matrix for TabSeq with weighted cross-entropy.}
\label{fig:pair_tabseq_w}
\end{figure}

\medskip
\noindent\textbf{Figure~\ref{fig:pair_tabseq_w} (TabSeq, W).}
ROC curves are consistently high; ordering-and-denoising prior to classification appears to aid class separability. The confusion matrix is cleanly diagonal with a small number of localized confusions, suggesting that TabSeq’s learned feature ordering mitigates some overlap but challenging pairs persist.

\medskip
\noindent\textit{Reading guide.}
Across models, AUC values near~1.0 denote strong one-vs-rest separability; diagonally dominant confusion matrices reflect balanced accuracy. Off-diagonal clusters typically indicate confusable, semantically related classes or minority-class scarcity; weighted cross-entropy mainly improves recall in these regions.

% ===== 1) 1D CNN (W) =====

% ===== 2) TabNet (W) =====

% ===== 3) FT-Transformer (W) =====

% ===== 4) TabTransformer (W) =====

% ===== 5) TabSeq (W) =====

\subsection{ROC Curves and Confusion Matrices for All Models with Standard Cross Entropy}
\label{ad}

\begin{figure}[htbp]
\centering
\begin{subfigure}{0.48\linewidth}
  \centering
  \includegraphics[width=\linewidth]{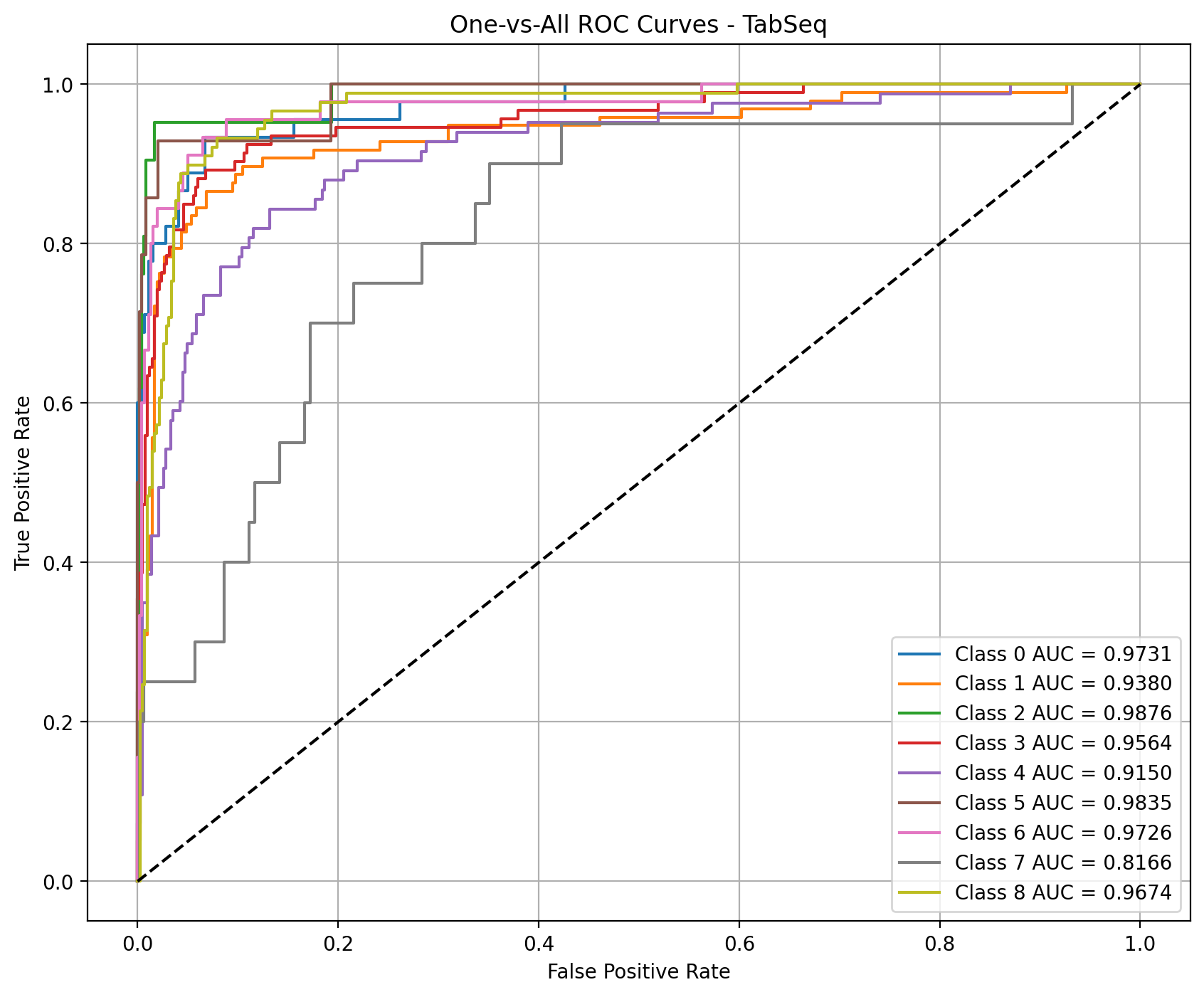}
  \caption{ROC (TabSeq)}
  \label{fig:roc_tabseq}
\end{subfigure}\hfill
\begin{subfigure}{0.48\linewidth}
  \centering
  \includegraphics[width=\linewidth]{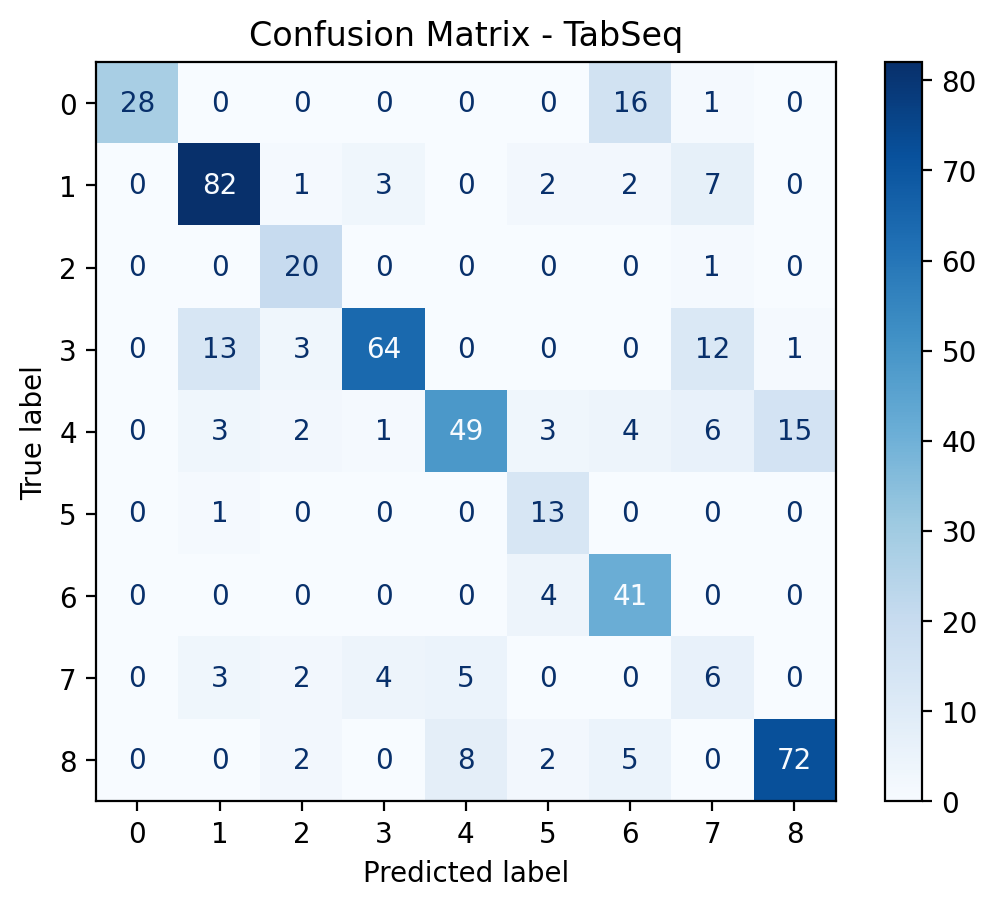}
  \caption{Confusion matrix (TabSeq)}
  \label{fig:cm_tabseq}
\end{subfigure}
\caption{One-vs-all ROC and confusion matrix for TabSeq with standard cross-entropy.}
\label{fig:pair_tabseq}
\end{figure}

\noindent\textbf{Fig.~\ref{fig:pair_tabseq} (TabSeq).}
High AUC across classes; the confusion matrix is largely diagonal with remaining errors concentrated in a few visually similar pairs consistent with TabSeq’s denoising + ordering benefit on correlated features.

\begin{figure}[htbp]
\centering
\begin{subfigure}{0.48\linewidth}
  \centering
  \includegraphics[width=\linewidth]{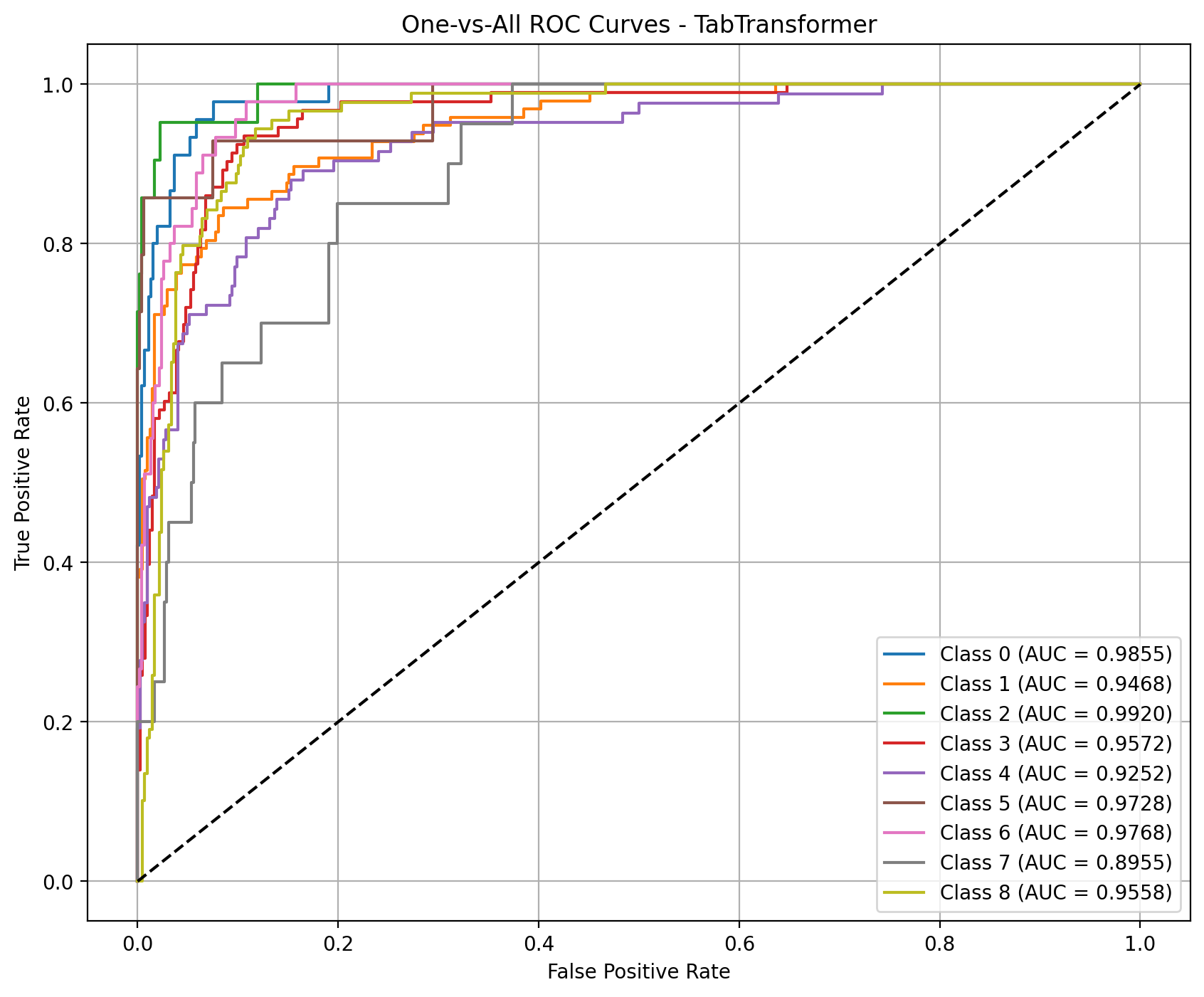}
  \caption{ROC (TabTransformer)}
  \label{fig:roc_tabt}
\end{subfigure}\hfill
\begin{subfigure}{0.48\linewidth}
  \centering
  \includegraphics[width=\linewidth]{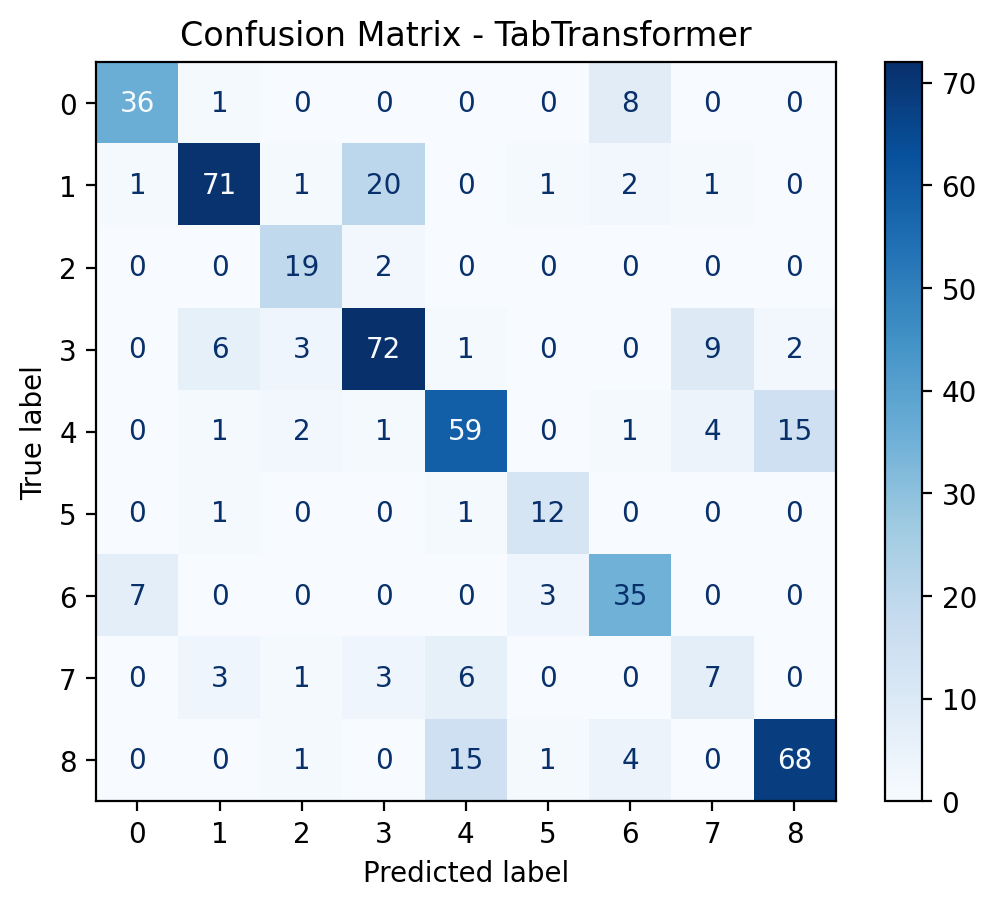}
  \caption{Confusion matrix (TabTransformer)}
  \label{fig:cm_tabt}
\end{subfigure}
\caption{One-vs-all ROC and confusion matrix for TabTransformer with standard cross-entropy.}
\label{fig:pair_tabt}
\end{figure}

\medskip
\noindent\textbf{Fig.~\ref{fig:pair_tabt} (TabTransformer).}
Steep ROC traces indicate strong one-vs-rest separability; off-diagonal mass is modest and localized, suggesting effective modeling of feature interactions without class weighting.

\begin{figure}[htbp]
\centering
\begin{subfigure}{0.48\linewidth}
  \centering
  \includegraphics[width=\linewidth]{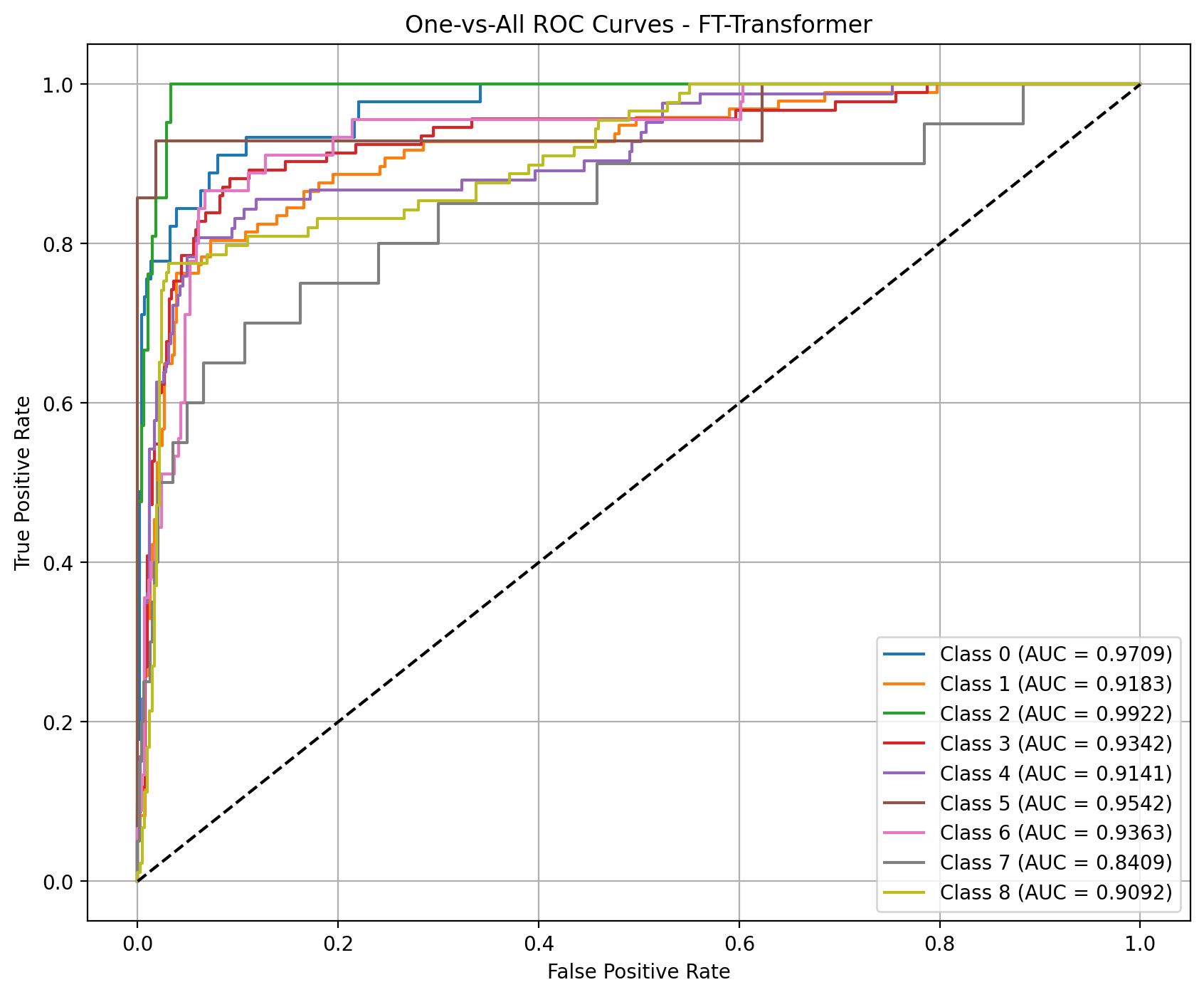}
  \caption{ROC (FT-Transformer)}
  \label{fig:roc_ft}
\end{subfigure}\hfill
\begin{subfigure}{0.48\linewidth}
  \centering
  \includegraphics[width=\linewidth]{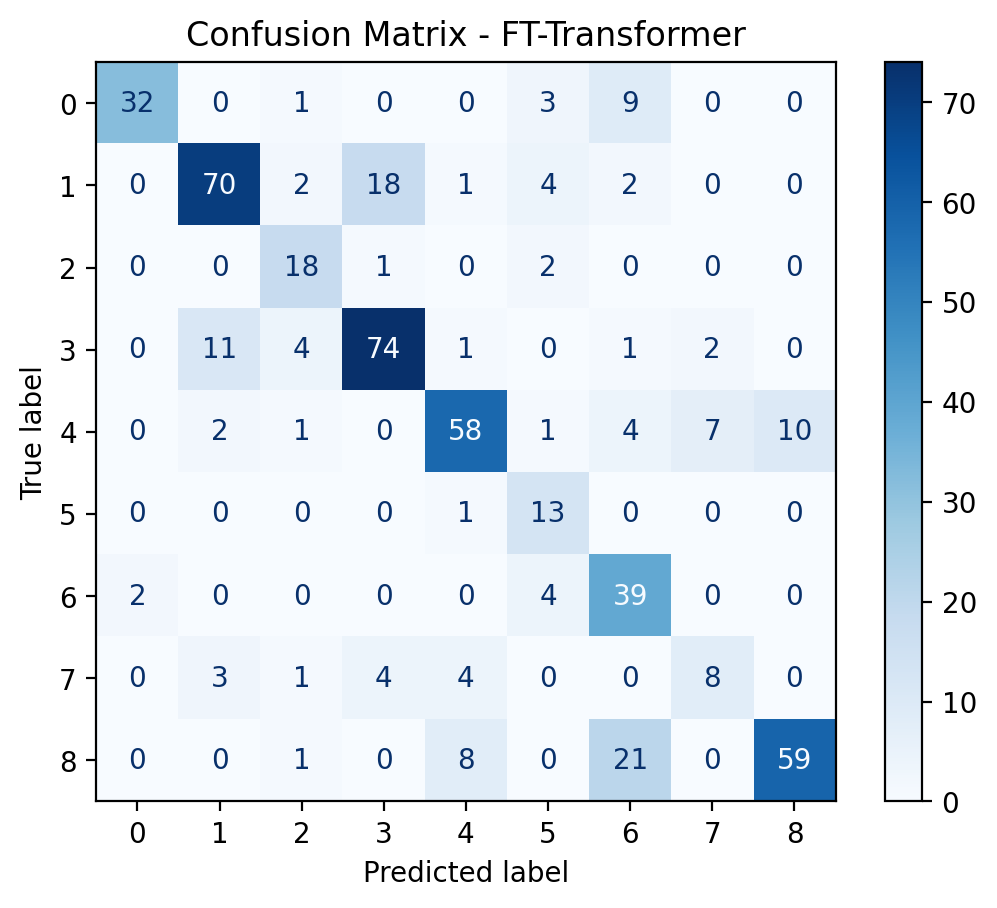}
  \caption{Confusion matrix (FT-Transformer)}
  \label{fig:cm_ft}
\end{subfigure}
\caption{One-vs-all ROC and confusion matrix for FT-Transformer with standard cross-entropy.}
\label{fig:pair_ft}
\end{figure}

\medskip
\noindent\textbf{Fig.~\ref{fig:pair_ft} (FT-Transformer).}
Uniformly high ROC curves with slightly larger variance for minority classes; confusion matrix shows a few systematic swaps, hinting at remaining imbalance sensitivity under standard CE.

\begin{figure}[htbp]
\centering
\begin{subfigure}{0.48\linewidth}
  \centering
  \includegraphics[width=\linewidth]{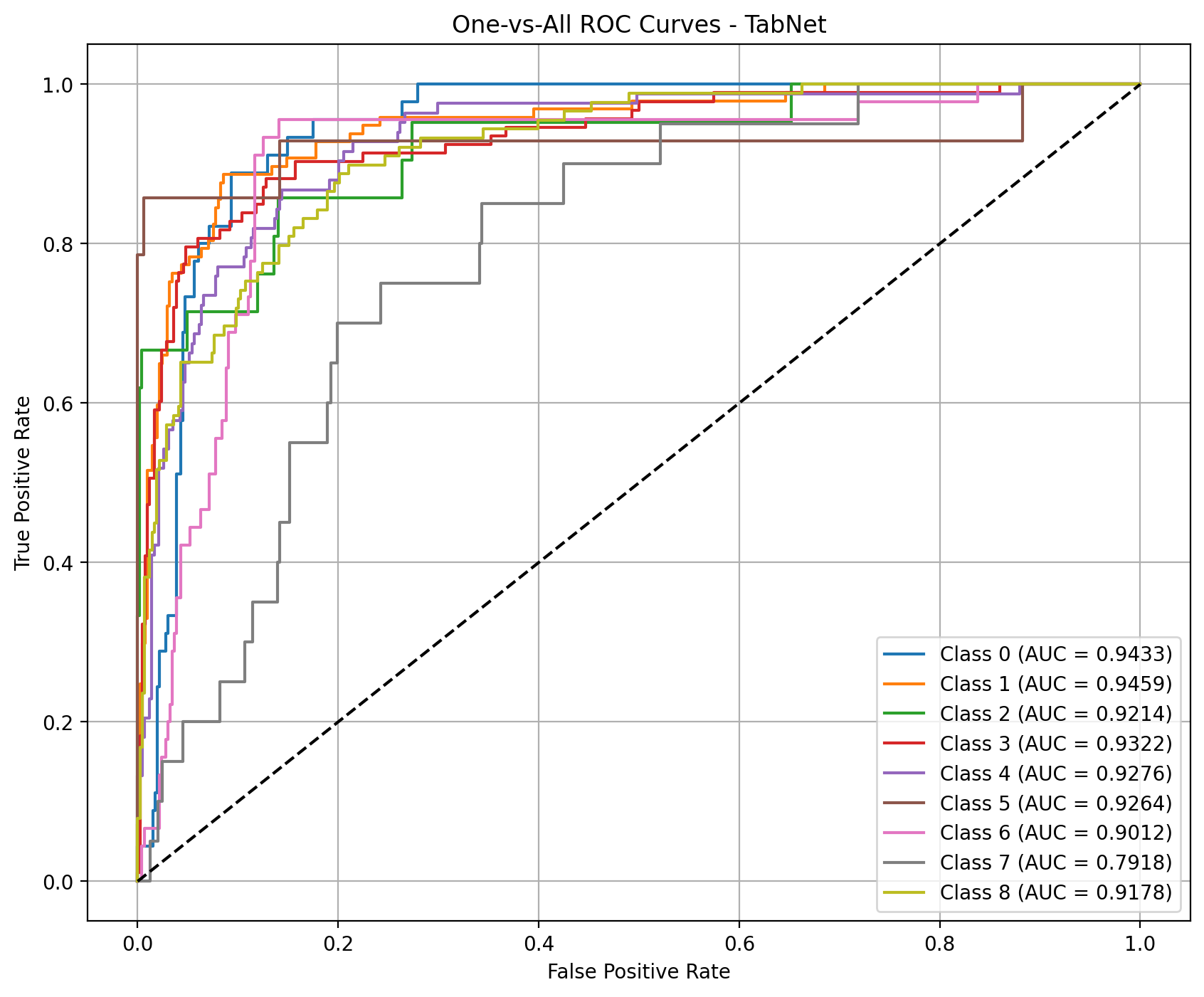}
  \caption{ROC (TabNet)}
  \label{fig:roc_tabnet}
\end{subfigure}\hfill
\begin{subfigure}{0.48\linewidth}
  \centering
  \includegraphics[width=\linewidth]{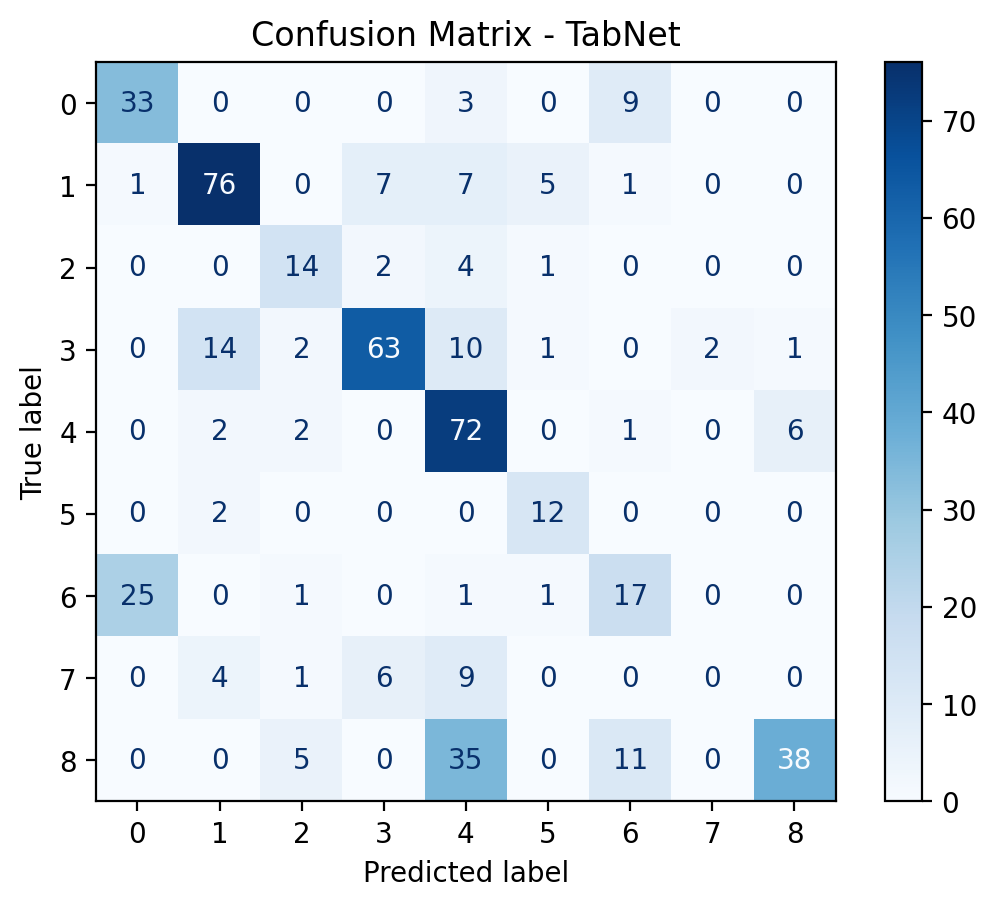}
  \caption{Confusion matrix (TabNet)}
  \label{fig:cm_tabnet}
\end{subfigure}
\caption{One-vs-all ROC and confusion matrix for TabNet with standard cross-entropy.}
\label{fig:pair_tabnet}
\end{figure}

\medskip
\noindent\textbf{Fig.~\ref{fig:pair_tabnet} (TabNet).}
Good but flatter ROC for some classes relative to transformer models; confusion patterns show occasional confusion among built-surface categories, aligning with attentive feature selection’s sensitivity to imbalance.

\begin{figure}[htbp]
\centering
\begin{subfigure}{0.48\linewidth}
  \centering
  \includegraphics[width=\linewidth]{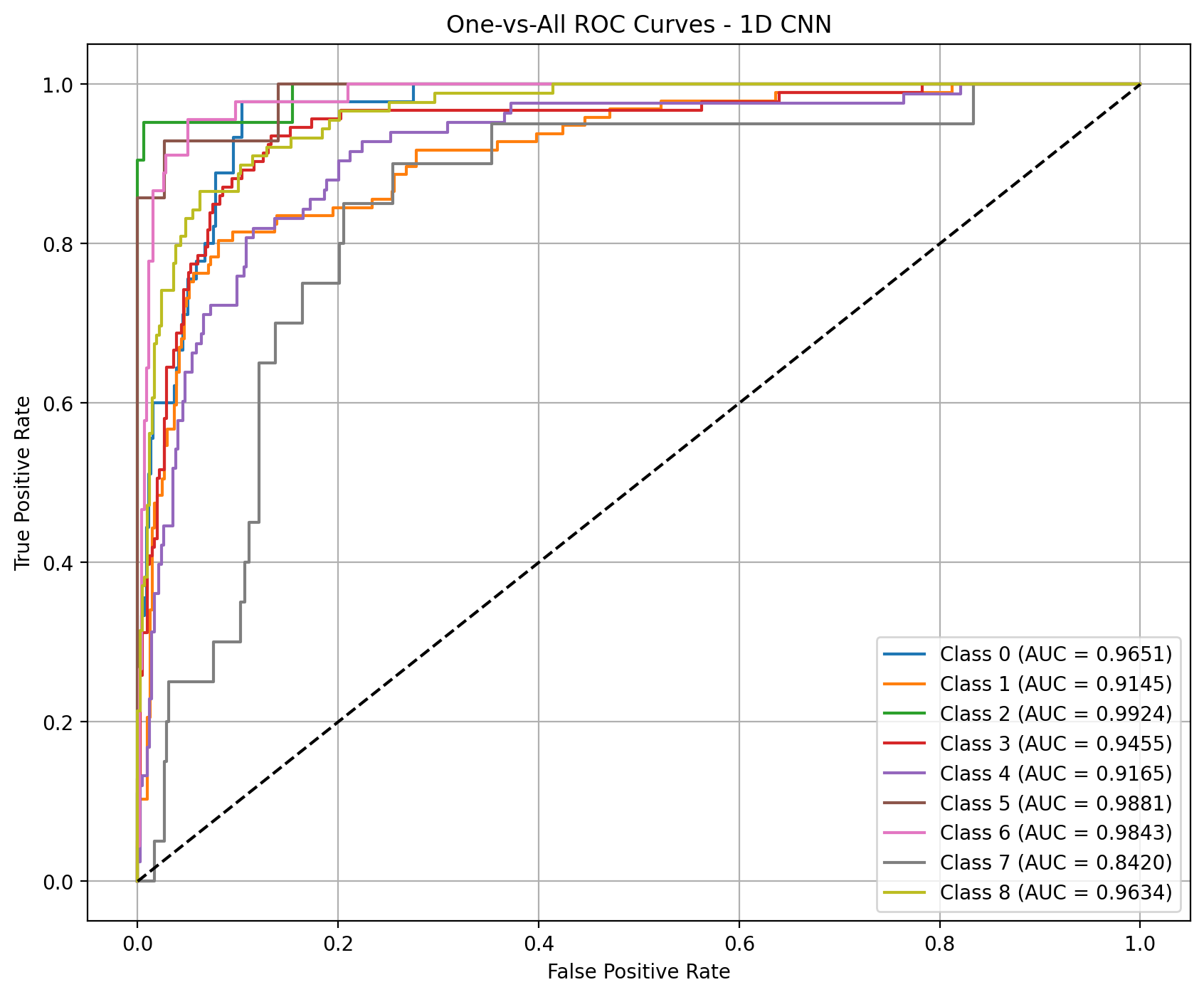}
  \caption{ROC (1D CNN)}
  \label{fig:roc_cnn}
\end{subfigure}\hfill
\begin{subfigure}{0.48\linewidth}
  \centering
  \includegraphics[width=\linewidth]{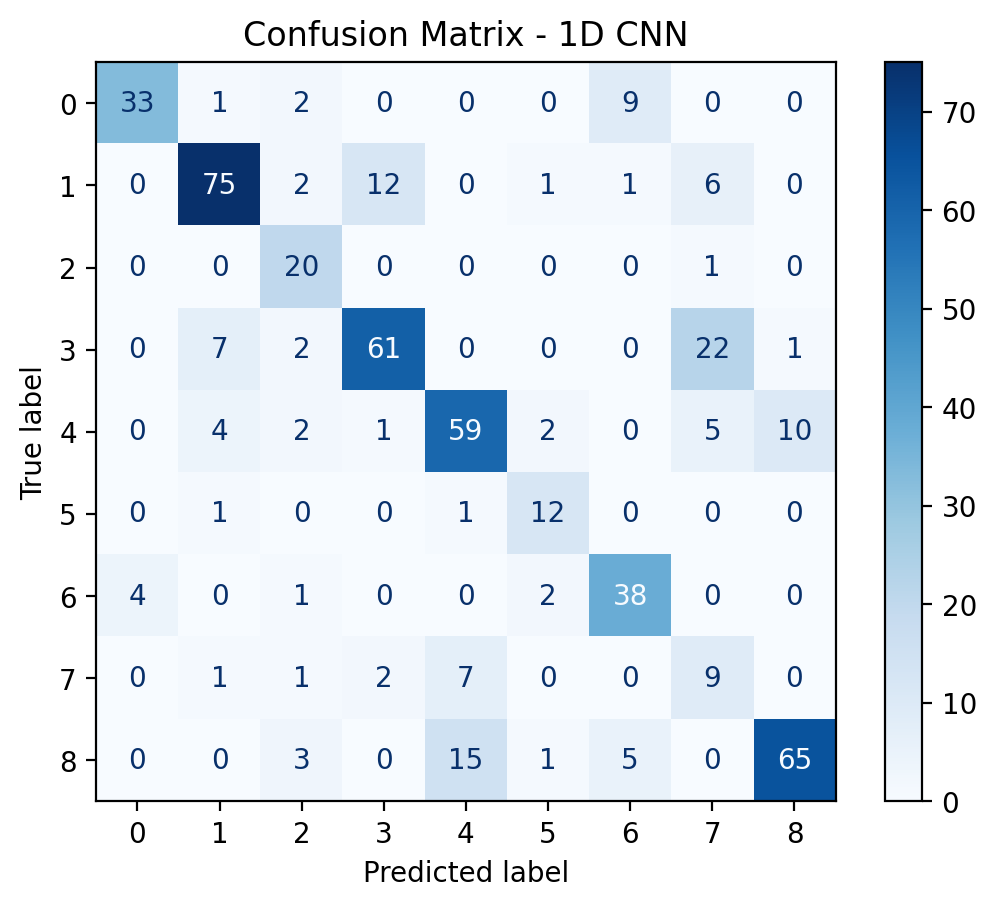}
  \caption{Confusion matrix (1D CNN)}
  \label{fig:cm_cnn}
\end{subfigure}
\caption{One-vs-all ROC and confusion matrix for 1D CNN with standard cross-entropy.}
\label{fig:pair_cnn}
\end{figure}

\medskip
\noindent\textbf{Fig.~\ref{fig:pair_cnn} (1D CNN).}
Strong ROC overall; the confusion matrix is diagonal-dominant with clustered mistakes where descriptors overlap. Sequential bias can help local feature groups but remains sensitive to feature permutations.

\begin{figure}[htbp]
\centering
\begin{subfigure}{0.48\linewidth}
  \centering
  \includegraphics[width=\linewidth]{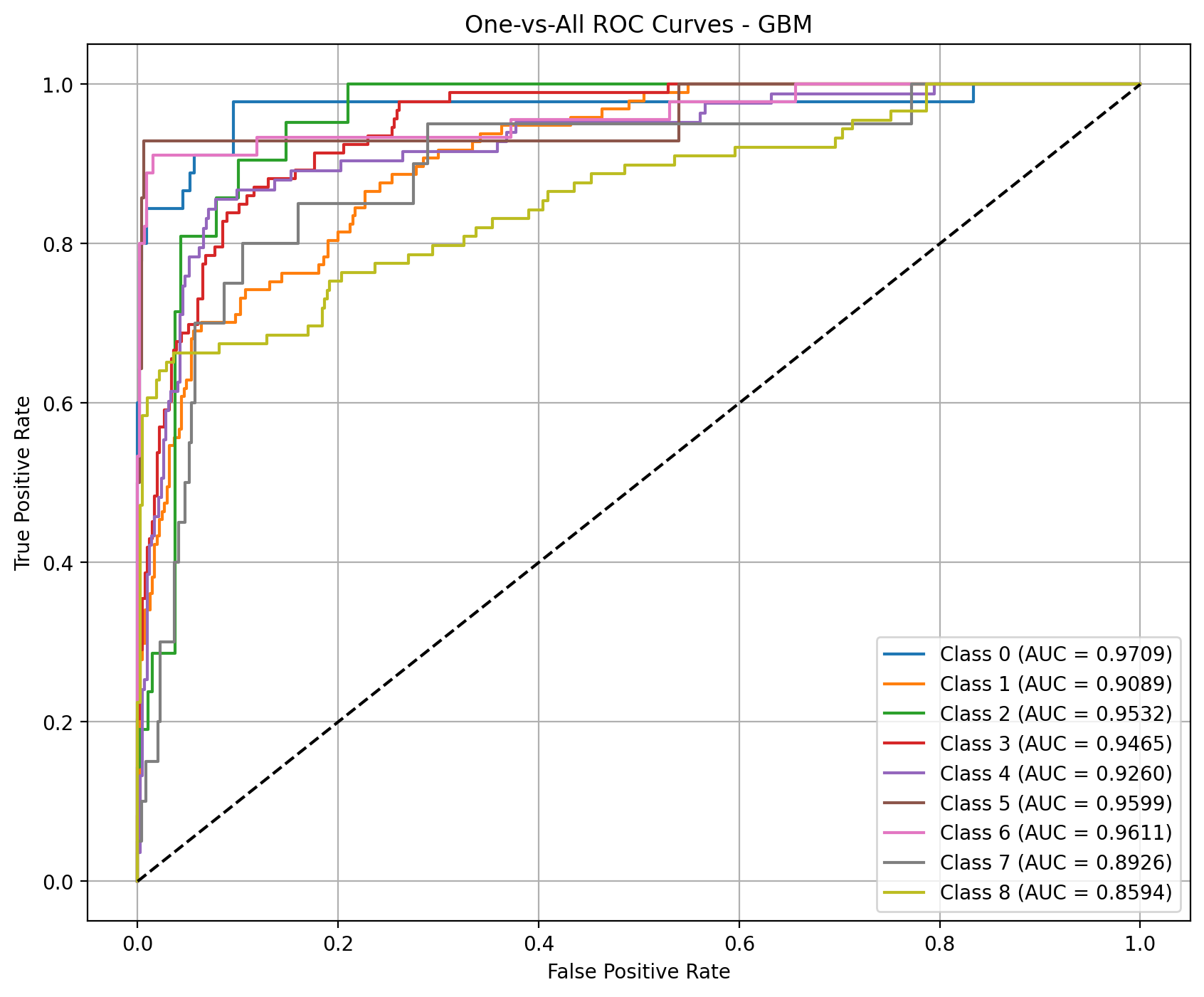}
  \caption{ROC (GBM)}
  \label{fig:roc_gbm}
\end{subfigure}\hfill
\begin{subfigure}{0.48\linewidth}
  \centering
  \includegraphics[width=\linewidth]{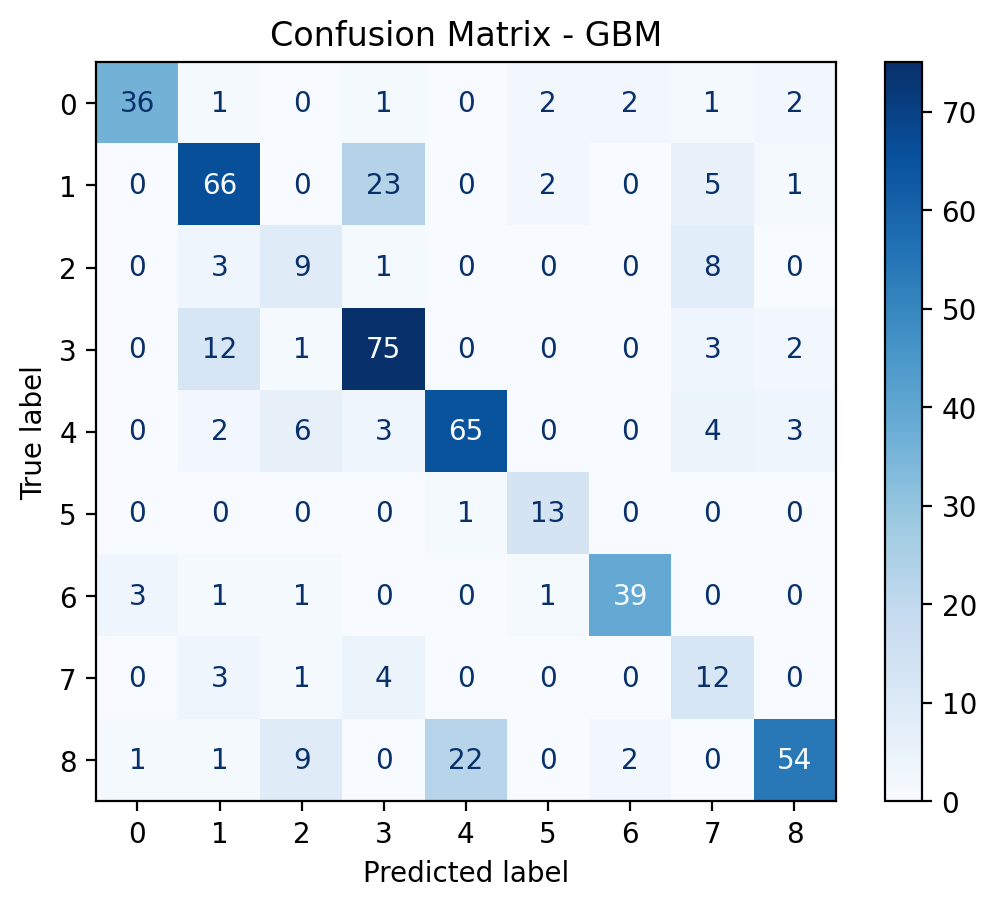}
  \caption{Confusion matrix (GBM)}
  \label{fig:cm_gbm}
\end{subfigure}
\caption{One-vs-all ROC and confusion matrix for Gradient Boosting (GBM).}
\label{fig:pair_gbm}
\end{figure}

\noindent\textbf{Fig.~\ref{fig:pair_gbm} (GBM).}
High AUC across most classes; confusion matrix is diagonally dominant with residual errors in a few built-surface vs.\ soil confusions.

\begin{figure}[htbp]
\centering
\begin{subfigure}{0.48\linewidth}
  \centering
  \includegraphics[width=\linewidth]{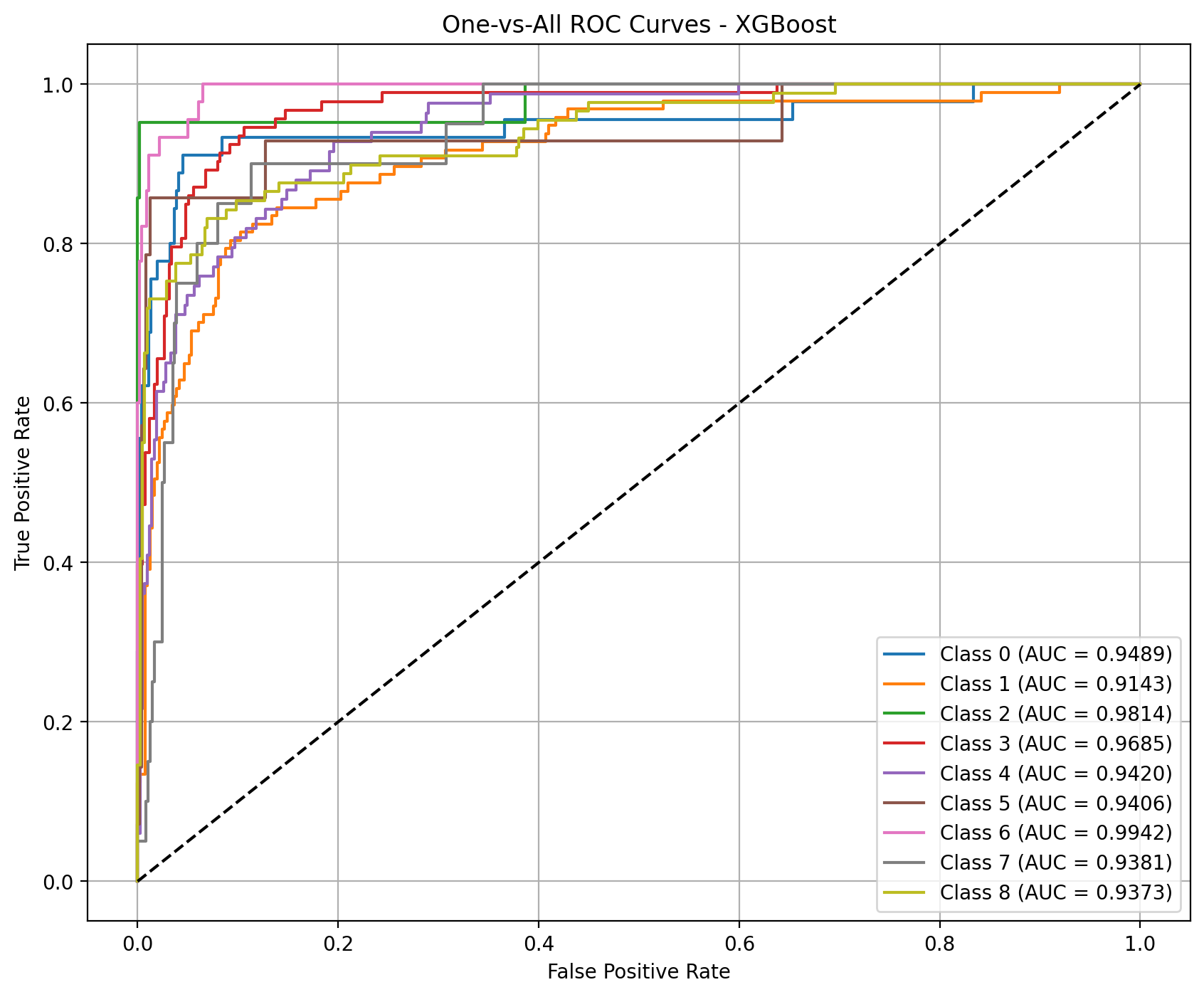}
  \caption{ROC (XGBoost)}
  \label{fig:roc_xgb}
\end{subfigure}\hfill
\begin{subfigure}{0.48\linewidth}
  \centering
  \includegraphics[width=\linewidth]{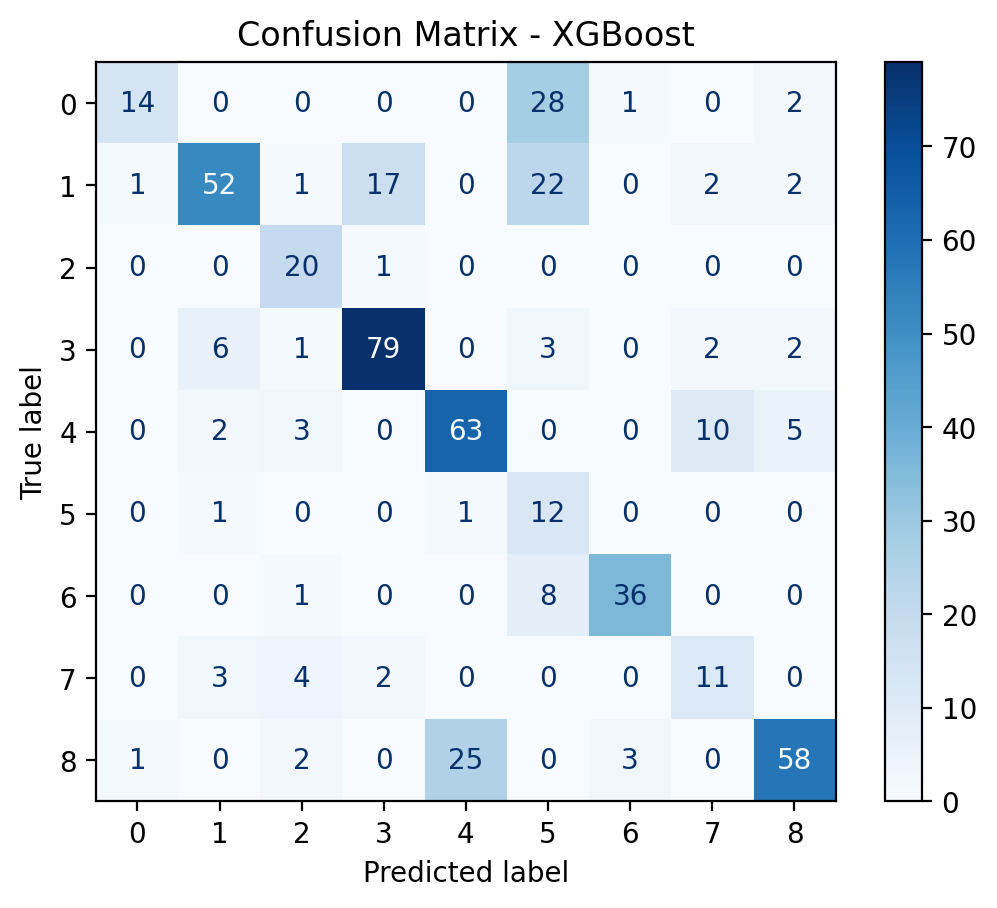}
  \caption{Confusion matrix (XGBoost)}
  \label{fig:cm_xgb}
\end{subfigure}
\caption{One-vs-all ROC and confusion matrix for XGBoost.}
\label{fig:pair_xgb}
\end{figure}

\medskip
\noindent\textbf{Fig.~\ref{fig:pair_xgb} (XGBoost).}
Consistently strong ROC traces and a clean confusion matrix; misclassifications are localized, reflecting robust handling of heterogeneous tabular features.

\begin{figure}[htbp]
\centering
\begin{subfigure}{0.48\linewidth}
  \centering
  \includegraphics[width=\linewidth]{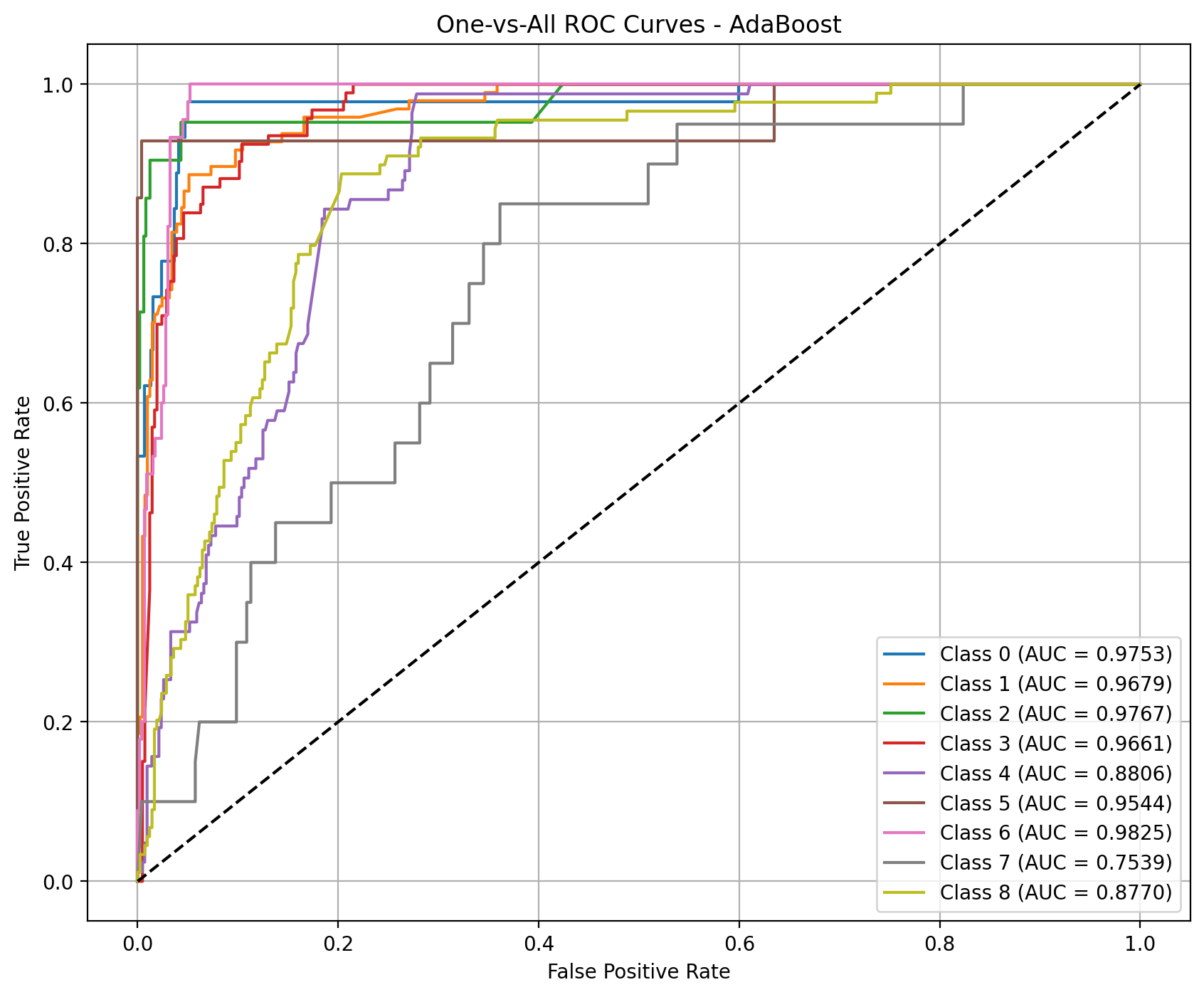}
  \caption{ROC (AdaBoost)}
  \label{fig:roc_ada}
\end{subfigure}\hfill
\begin{subfigure}{0.48\linewidth}
  \centering
  \includegraphics[width=\linewidth]{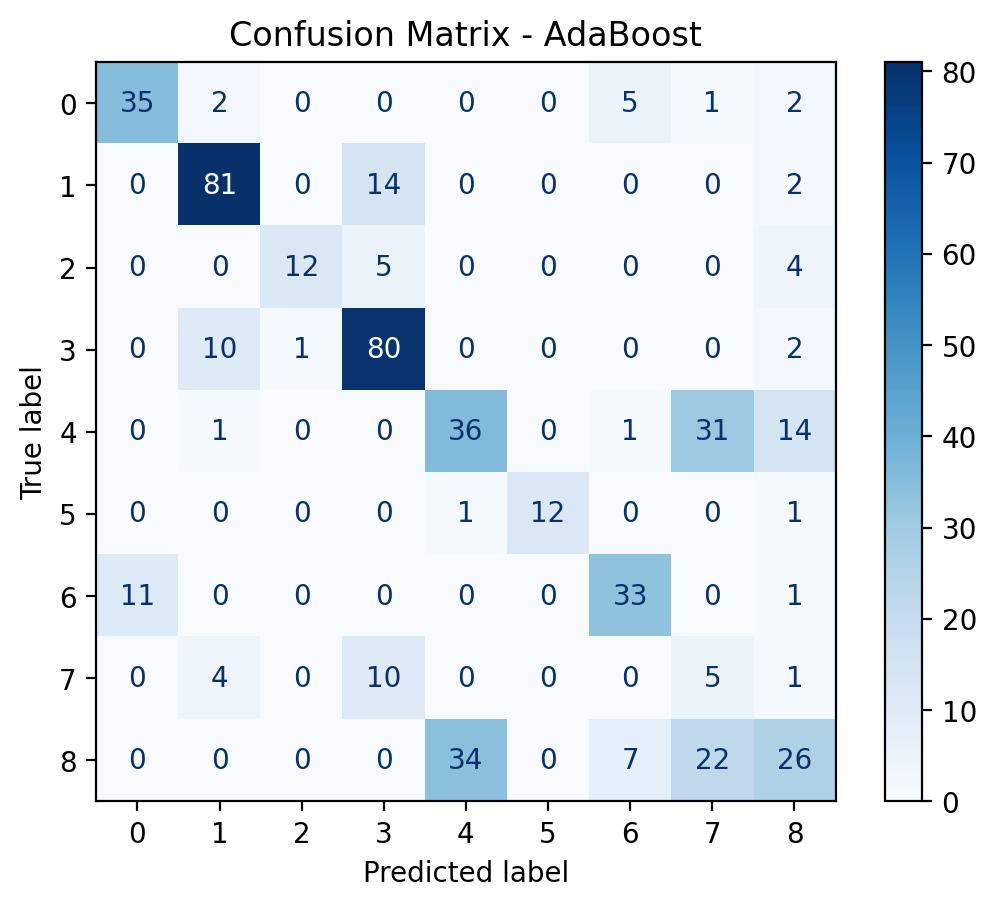}
  \caption{Confusion matrix (AdaBoost)}
  \label{fig:cm_ada}
\end{subfigure}
\caption{One-vs-all ROC and confusion matrix for AdaBoost.}
\label{fig:pair_ada}
\end{figure}

\medskip
\noindent\textbf{Fig.~\ref{fig:pair_ada} (AdaBoost).}
ROC performance is competitive but shows larger variance for minority classes; confusion matrix reveals a few systematic swaps, typical for boosting with limited data.

\begin{figure}[htbp]
\centering
\begin{subfigure}{0.48\linewidth}
  \centering
  \includegraphics[width=\linewidth]{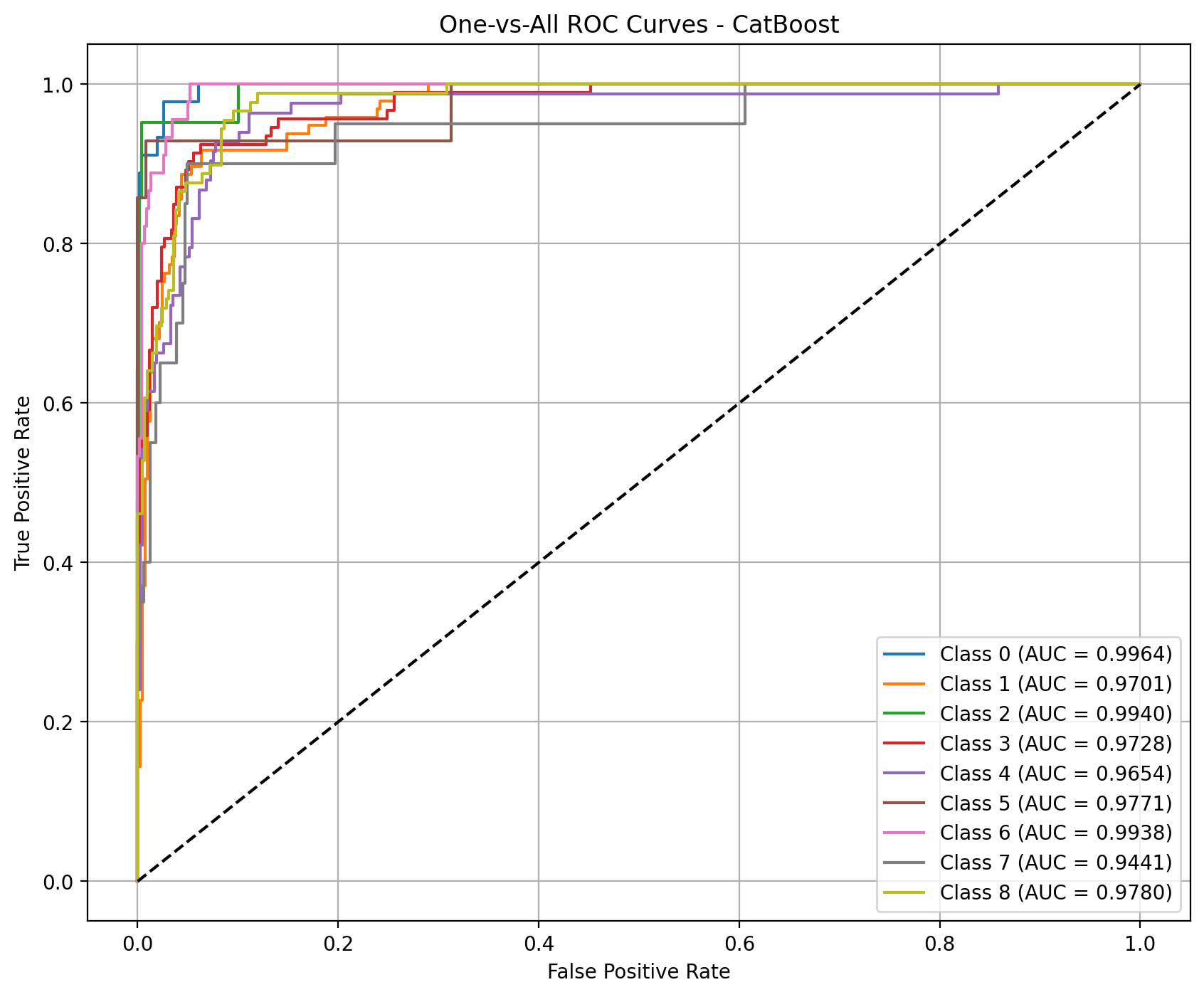}
  \caption{ROC (CatBoost)}
  \label{fig:roc_cat}
\end{subfigure}\hfill
\begin{subfigure}{0.48\linewidth}
  \centering
  \includegraphics[width=\linewidth]{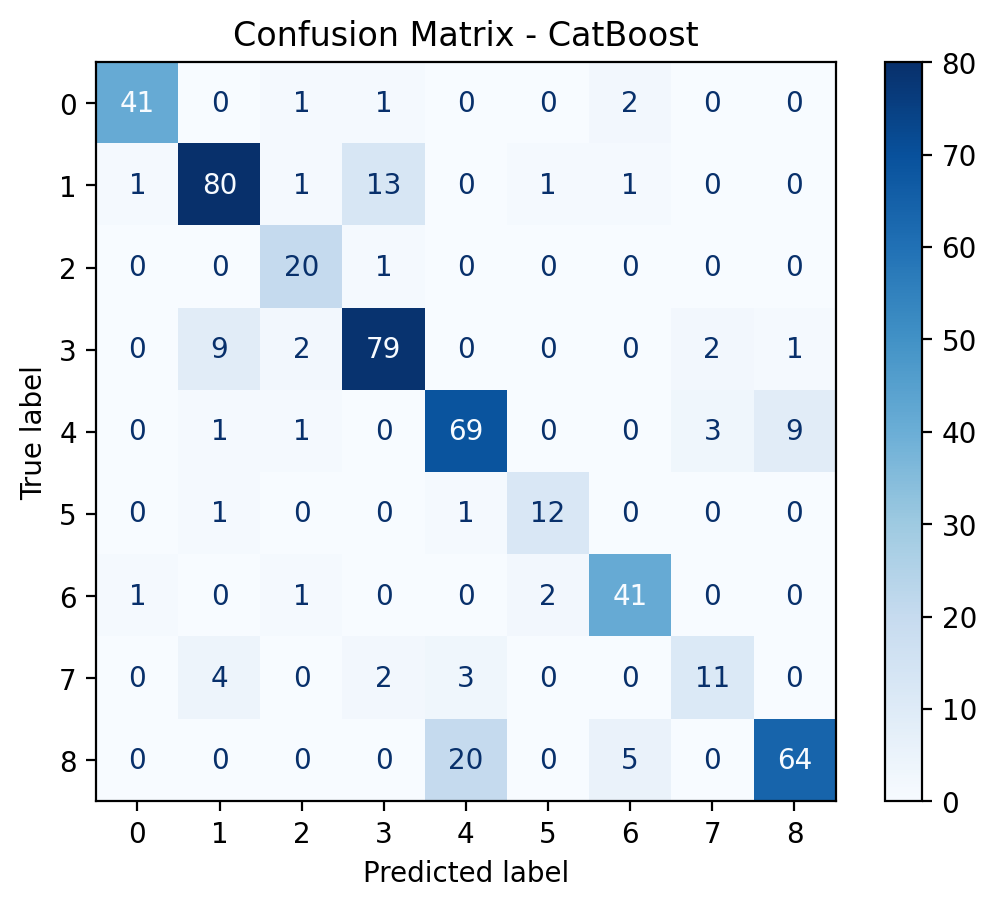}
  \caption{Confusion matrix (CatBoost)}
  \label{fig:cm_cat}
\end{subfigure}
\caption{One-vs-all ROC and confusion matrix for CatBoost.}
\label{fig:pair_cat}
\end{figure}

\medskip
\noindent\textbf{Fig.~\ref{fig:pair_cat} (CatBoost).}
Steep ROC curves with near-ceiling AUC on several classes; confusion matrix is the most diagonal among ensembles, aligning with CatBoost’s strong overall accuracy.

\begin{figure}[htbp]
\centering
\begin{subfigure}{0.48\linewidth}
  \centering
  \includegraphics[width=\linewidth]{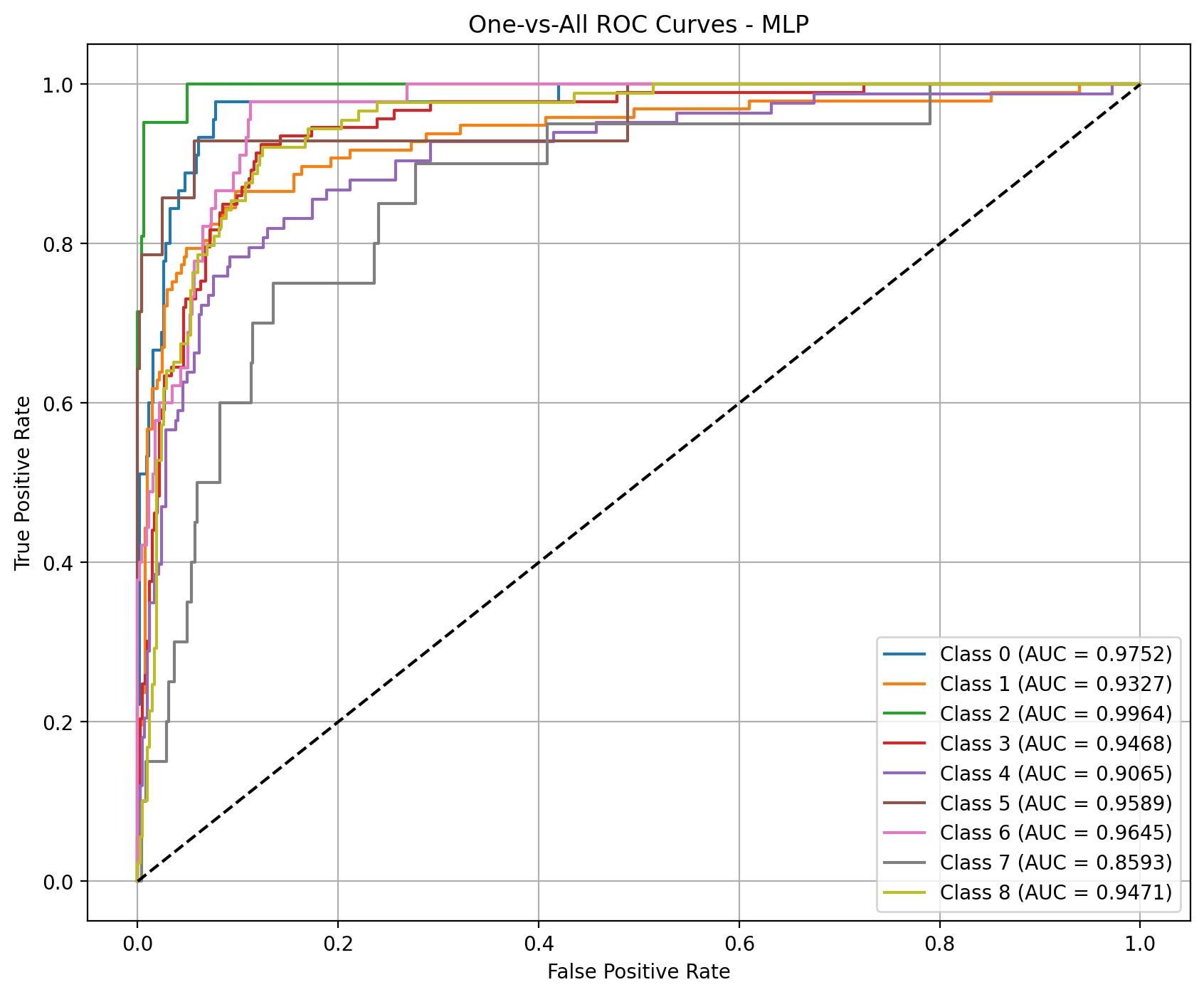}
  \caption{ROC (MLP)}
  \label{fig:roc_mlp}
\end{subfigure}\hfill
\begin{subfigure}{0.48\linewidth}
  \centering
  \includegraphics[width=\linewidth]{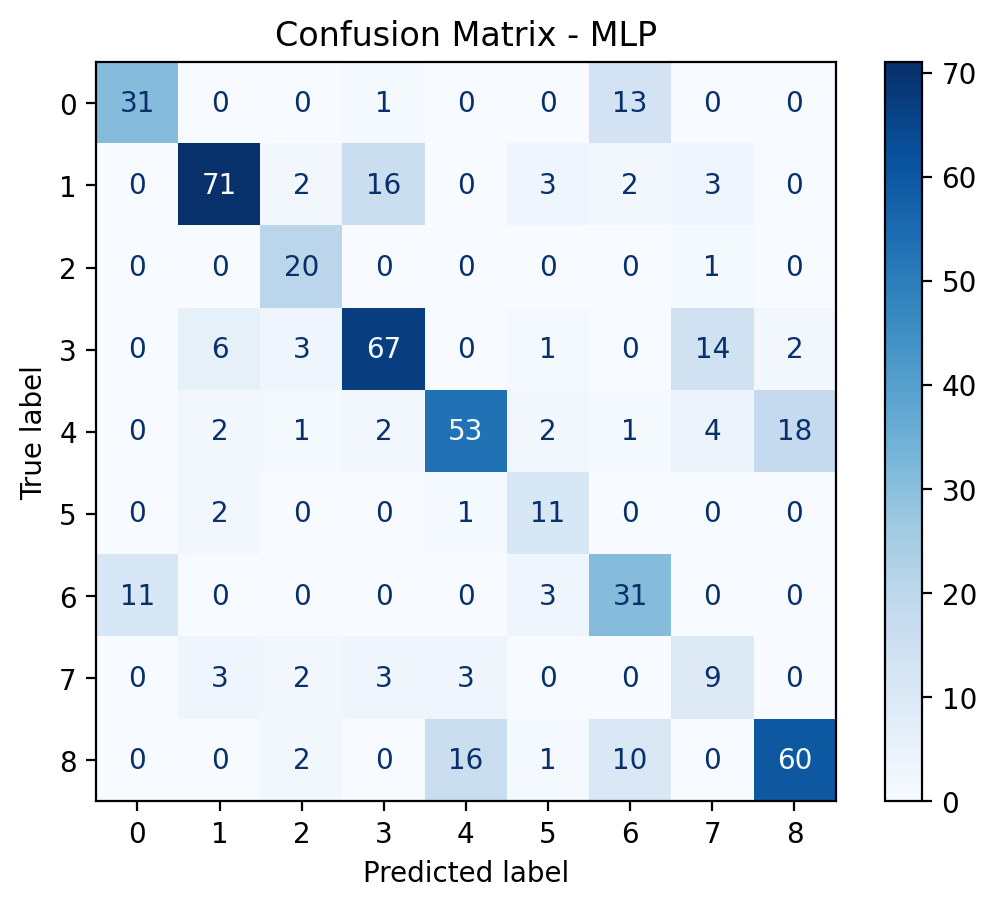}
  \caption{Confusion matrix (MLP)}
  \label{fig:cm_mlp}
\end{subfigure}
\caption{One-vs-all ROC and confusion matrix for a shallow MLP.}
\label{fig:pair_mlp}
\end{figure}

\medskip
\noindent\textbf{Fig.~\ref{fig:pair_mlp} (MLP).}
Neural baseline shows high AUC but slightly more off-diagonal mass, indicating sensitivity to class imbalance and feature scaling compared to tree ensembles.

\begin{figure*}[htbp]
\centering
\begin{subfigure}{0.49\linewidth}
  \centering
  \includegraphics[width=\linewidth]{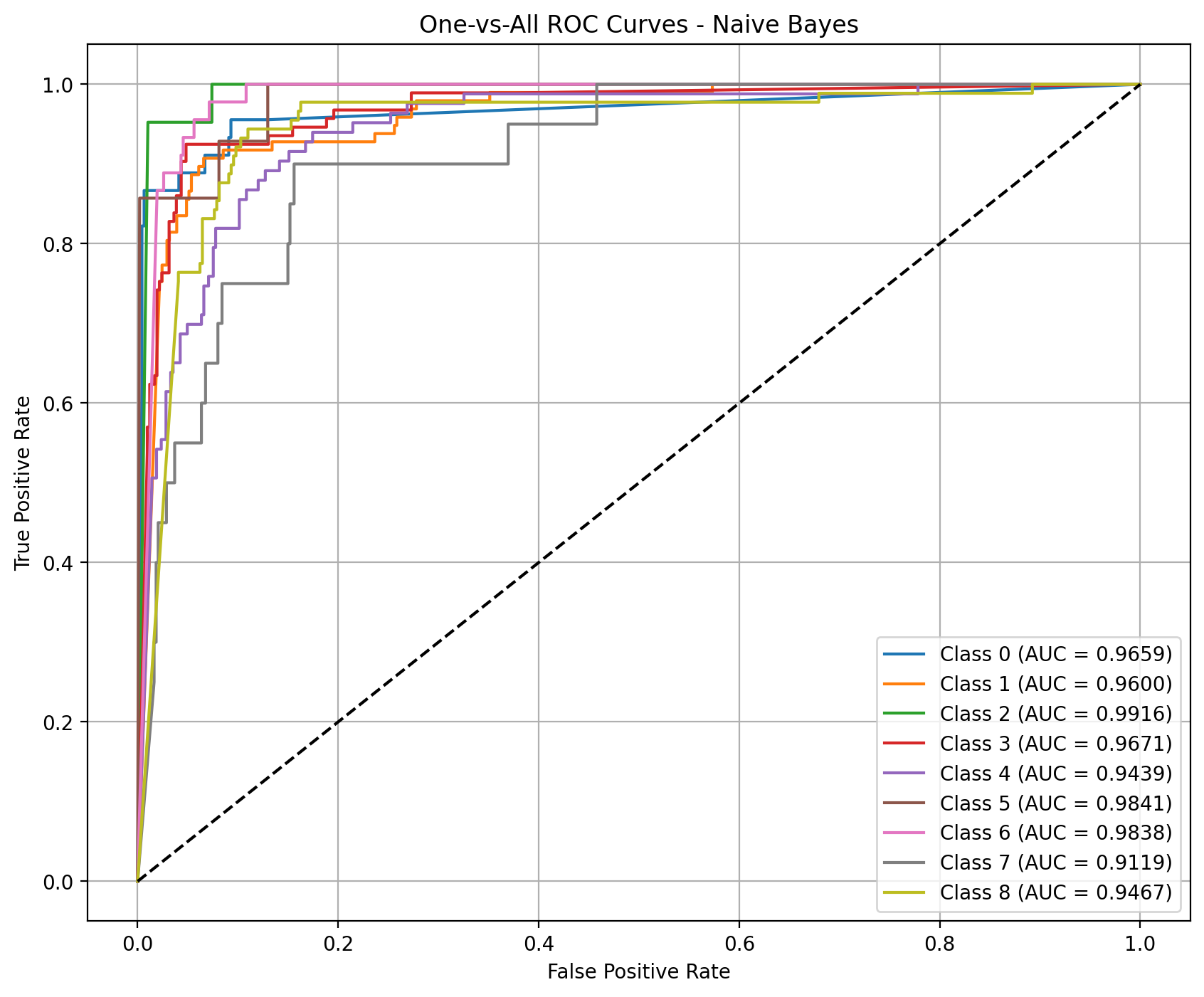}
  \caption{ROC (Naive Bayes)}
  \label{fig:nb_roc}
\end{subfigure}
\begin{subfigure}{0.49\linewidth}
  \centering
  \includegraphics[width=\linewidth]{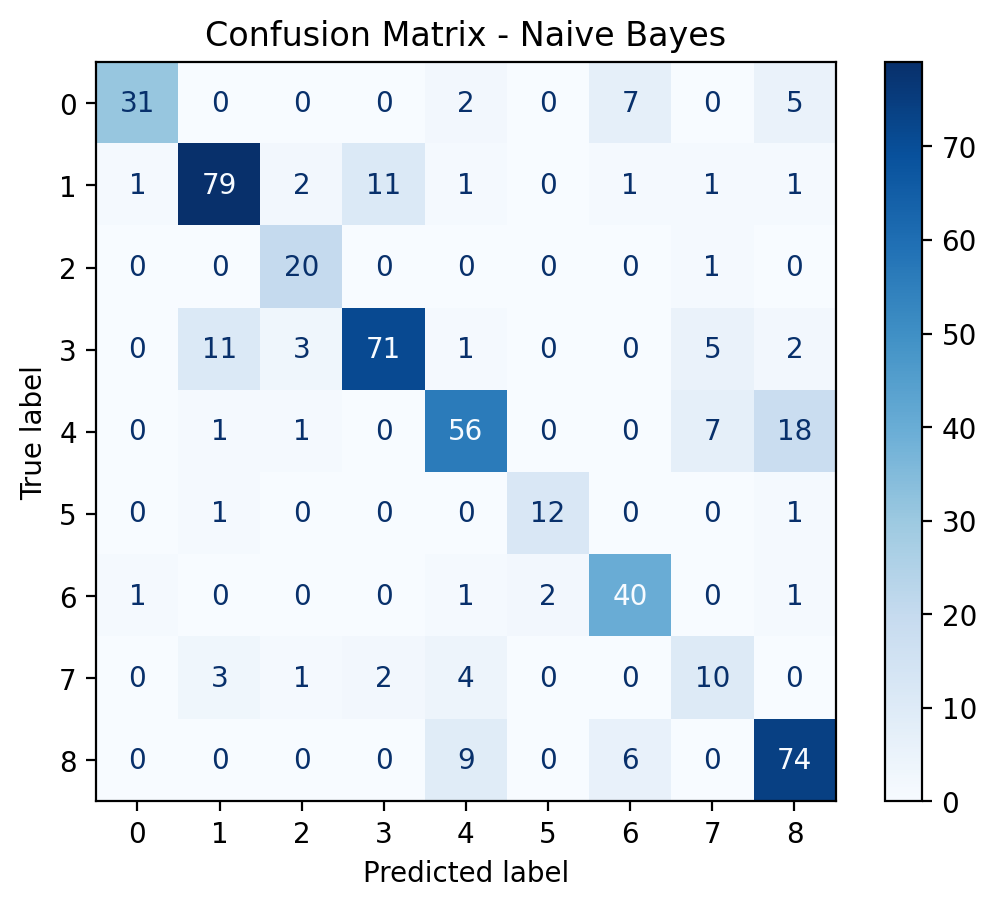}
  \caption{Confusion matrix (Naive Bayes).}
  \label{fig:nb_conf}
\end{subfigure}
\caption{One-vs-all ROC and confusion matrix for Naive Bayes.}
\label{fig:pair_nb}
\end{figure*}

\medskip
\noindent\textbf{Fig.~\ref{fig:pair_nb} (Naive Bayes).}
Consistently high AUCs and a largely diagonal matrix; remaining errors cluster among a few visually similar classes, reflecting conditional-independence limits.

\begin{figure*}[htbp]
\centering
\begin{subfigure}{0.49\linewidth}
  \centering
  \includegraphics[width=\linewidth]{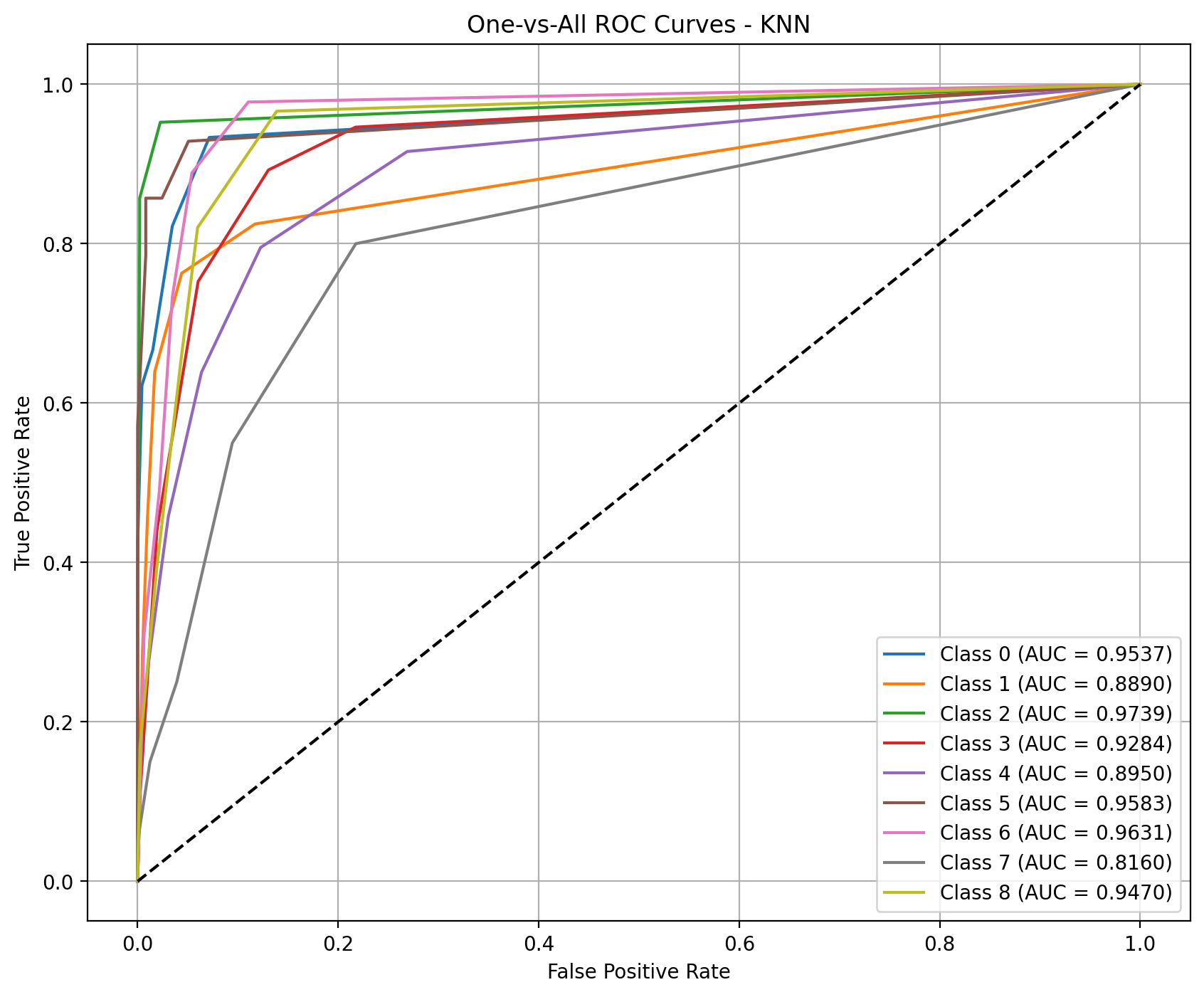}
  \caption{ROC (KNN)}
  \label{fig:knn_roc}
\end{subfigure}
\begin{subfigure}{0.49\linewidth}
  \centering
  \includegraphics[width=\linewidth]{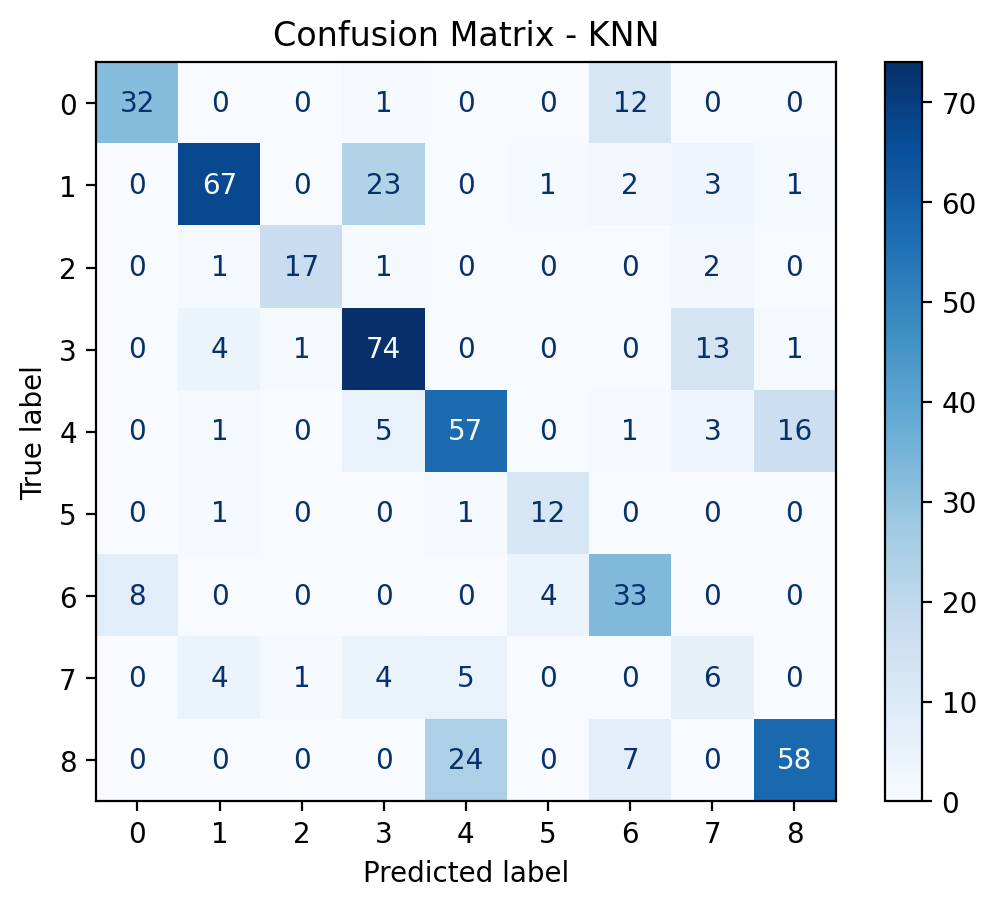}
  \caption{Confusion matrix (KNN).}
  \label{fig:knn_conf}
\end{subfigure}
\caption{One-vs-all ROC and confusion matrix for KNN.}
\label{fig:pair_knn}
\end{figure*}

\medskip
\noindent\textbf{Fig.~\ref{fig:pair_knn} (KNN).}
Competitive AUCs with thinner margins on several classes; confusion is diffuse rather than localized, typical of distance-based decision boundaries.

\begin{figure*}[htbp]
\centering
\begin{subfigure}{0.49\linewidth}
  \centering
  \includegraphics[width=\linewidth]{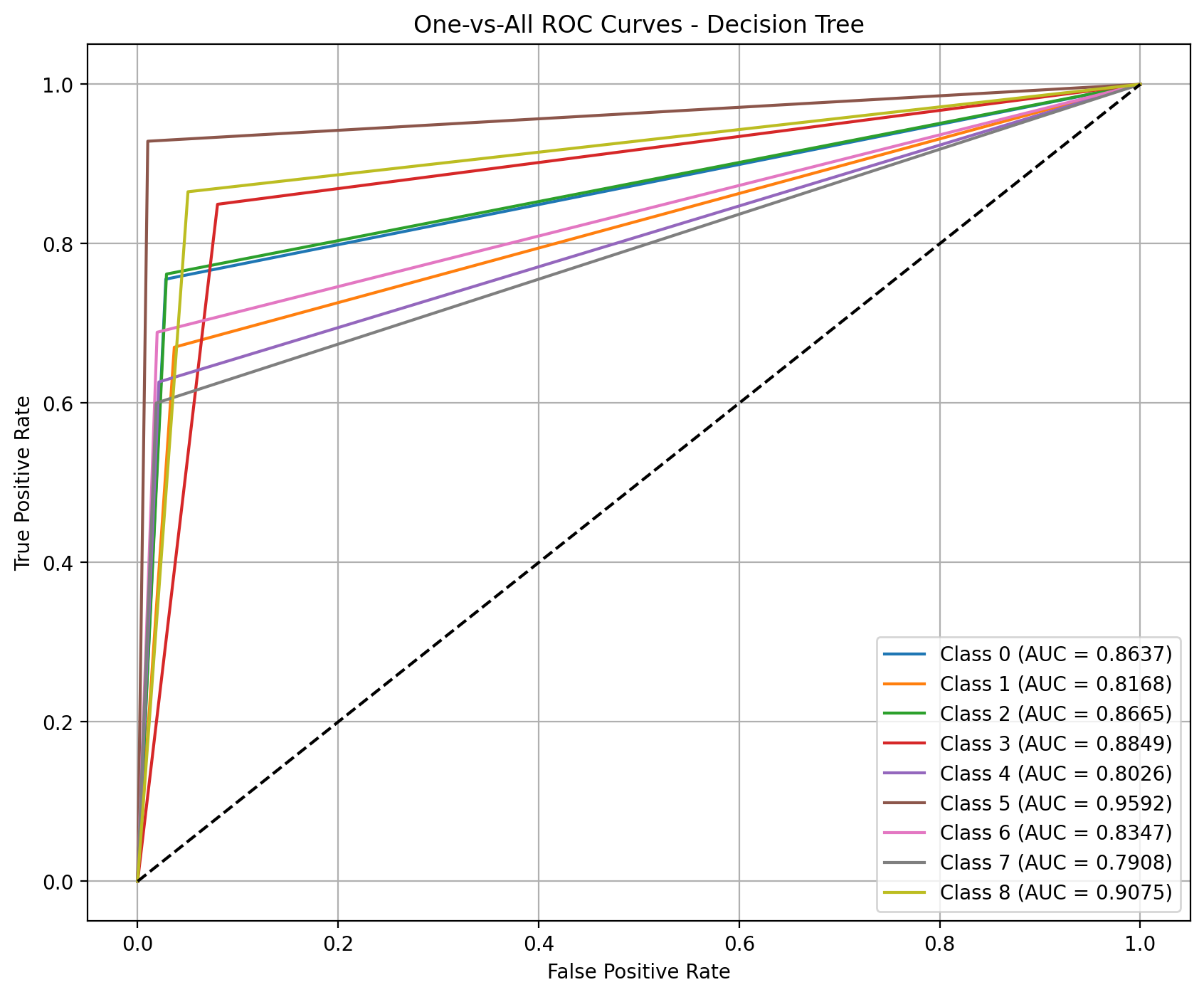}
  \caption{ROC (Decision Tree).}
  \label{fig:dt_roc}
\end{subfigure}
\begin{subfigure}{0.49\linewidth}
  \centering
  \includegraphics[width=\linewidth]{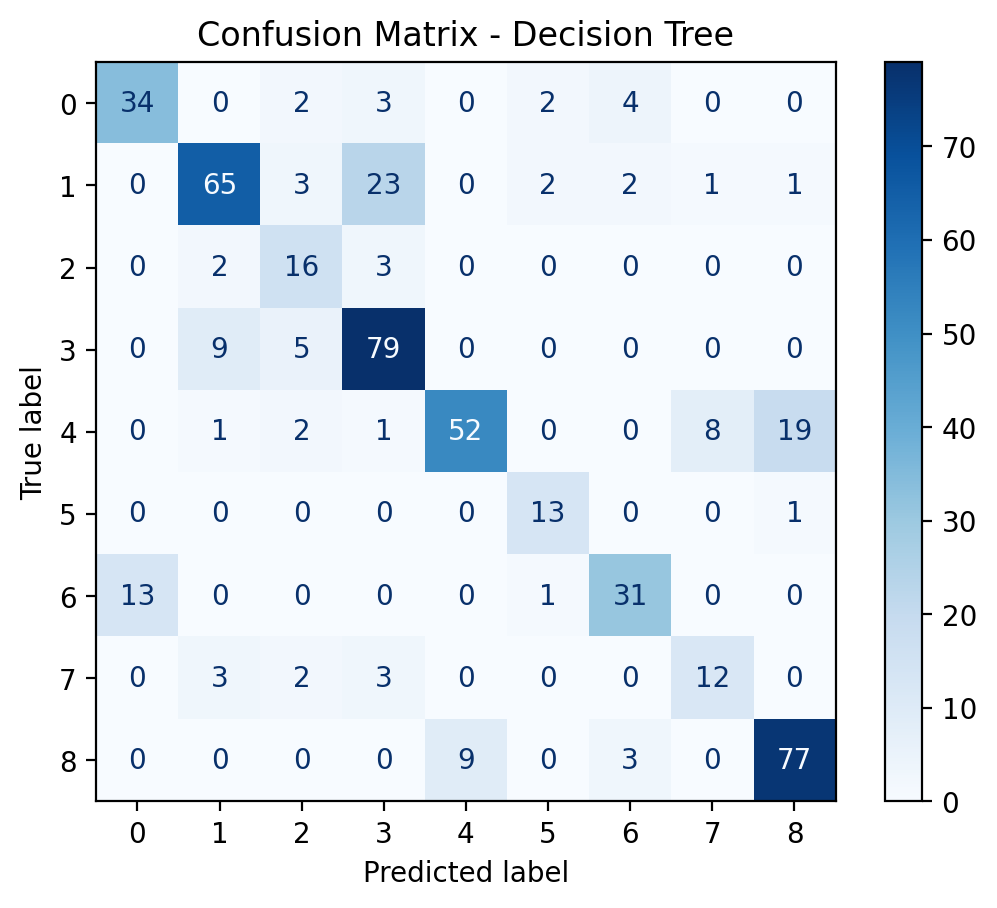}
  \caption{Confusion matrix (Decision Tree).}
  \label{fig:dt_conf}
\end{subfigure}
\caption{One-vs-all ROC and confusion matrix for Decision Tree.}
\label{fig:pair_dt}
\end{figure*}

\medskip
\noindent\textbf{Fig.~\ref{fig:pair_dt} (Decision Tree).}
Lower and more variable AUCs across classes; single-tree decision boundaries yield heavier off-diagonal mass, suggesting underfitting/instability relative to ensembles.

\begin{figure*}[htbp]
\centering
\begin{subfigure}{0.49\linewidth}
  \centering
  \includegraphics[width=\linewidth]{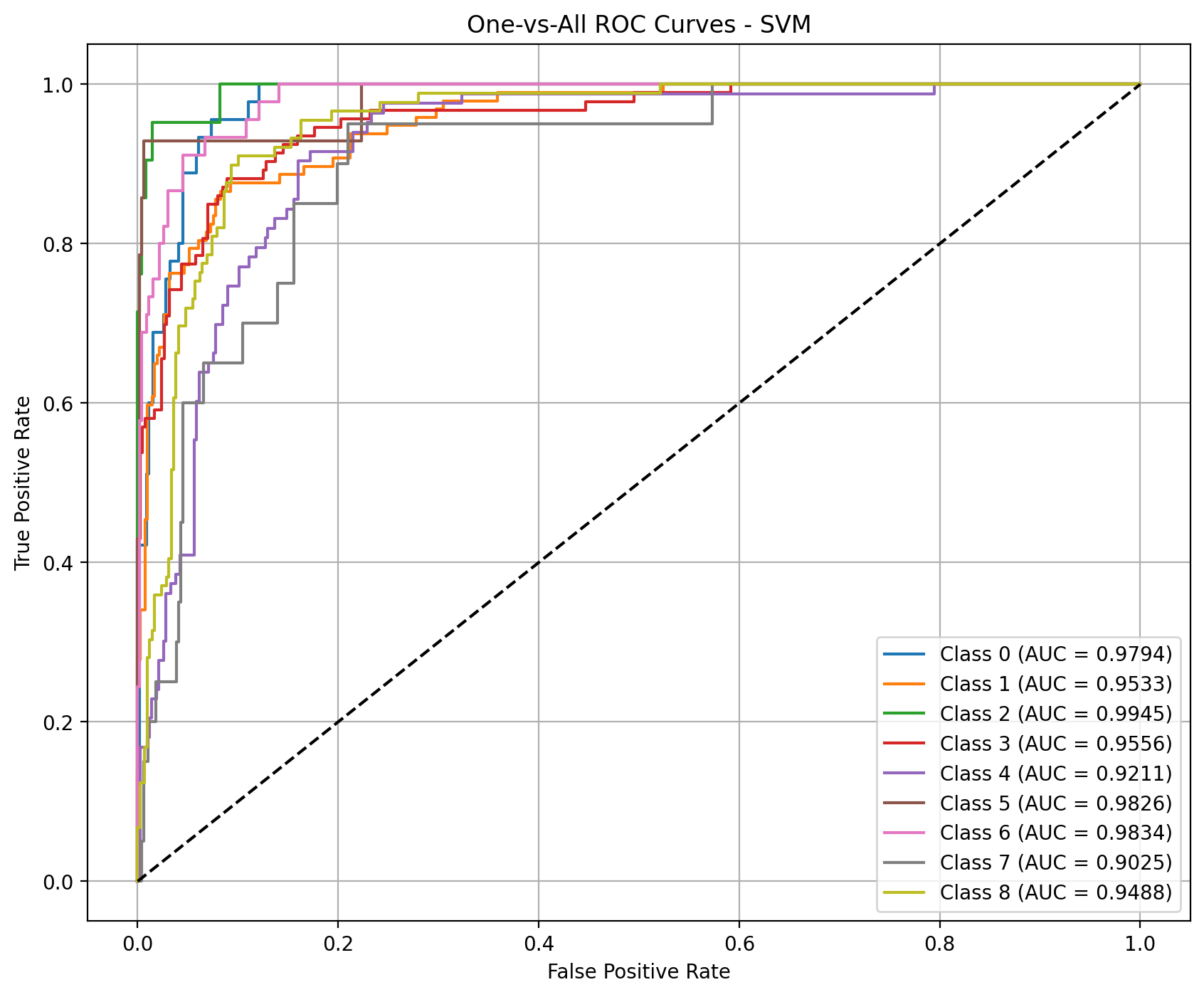}
  \caption{ROC (SVM).}
  \label{fig:svm_roc}
\end{subfigure}
\begin{subfigure}{0.49\linewidth}
  \centering
  \includegraphics[width=\linewidth]{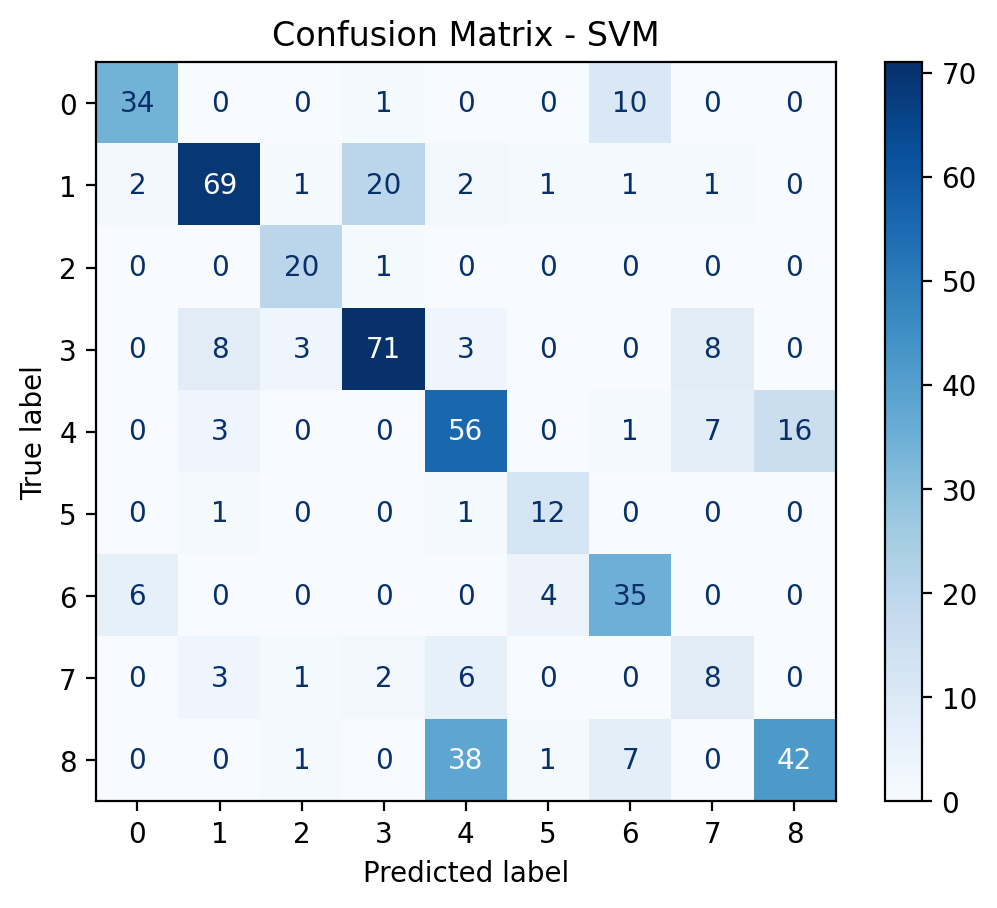}
  \caption{Confusion matrix (SVM).}
  \label{fig:svm_conf}
\end{subfigure}
\caption{One-vs-all ROC and confusion matrix for SVM.}
\label{fig:pair_svm}
\end{figure*}

\medskip
\noindent\textbf{Fig.~\ref{fig:pair_svm} (SVM).}
Strong one-vs-rest separability and clean diagonals for most classes; residual confusion is limited to a few neighboring categories, matching SVMs’ large-margin behavior.

\begin{figure}[htbp]
\centering
\captionsetup[sub]{font=small}

\begin{subfigure}{0.49\linewidth}
  \centering
  \includegraphics[width=\linewidth]{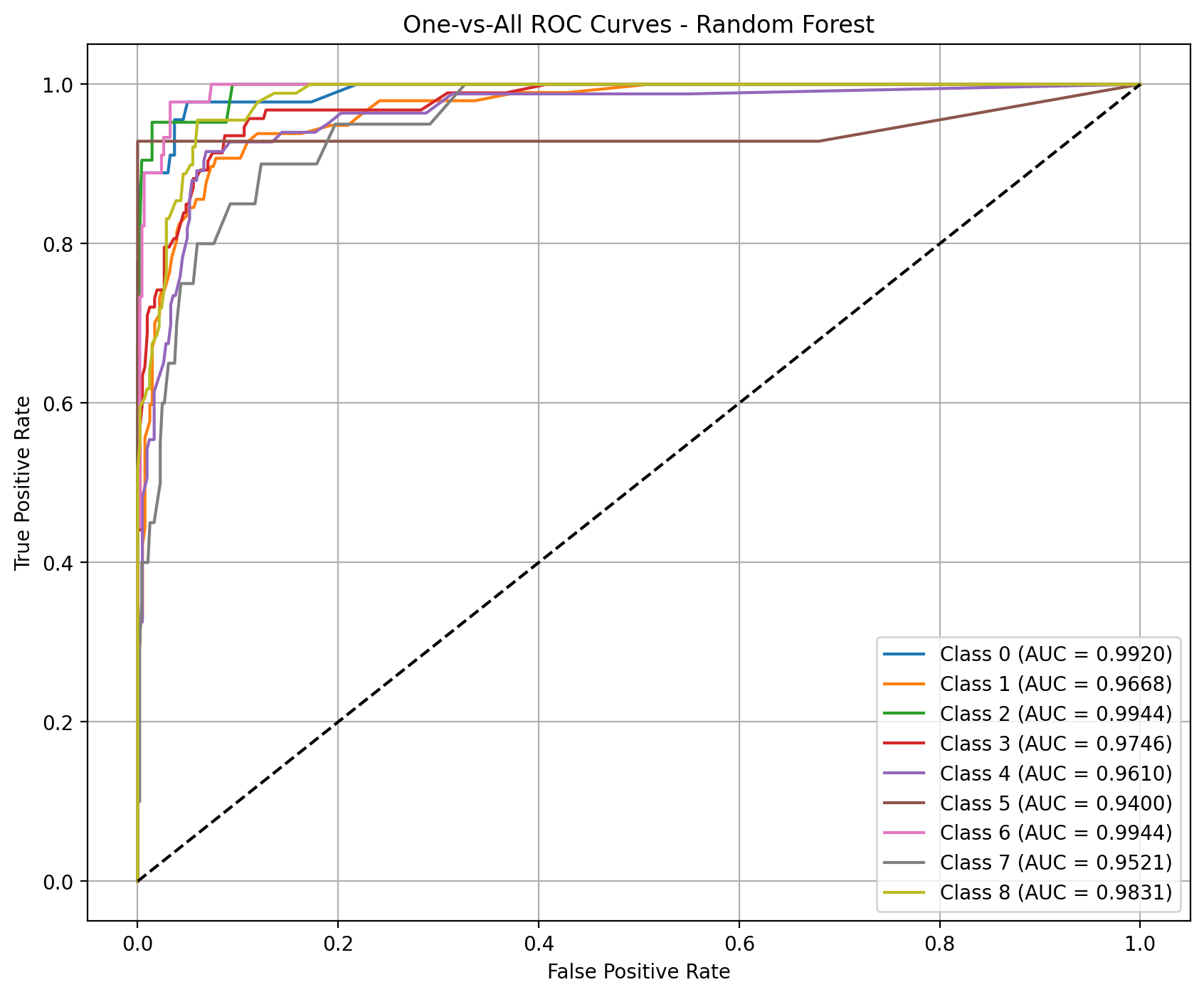}
  \caption{ROC (Random Forest).}
\end{subfigure}\hfill
\begin{subfigure}{0.49\linewidth}
  \centering
  \includegraphics[width=\linewidth]{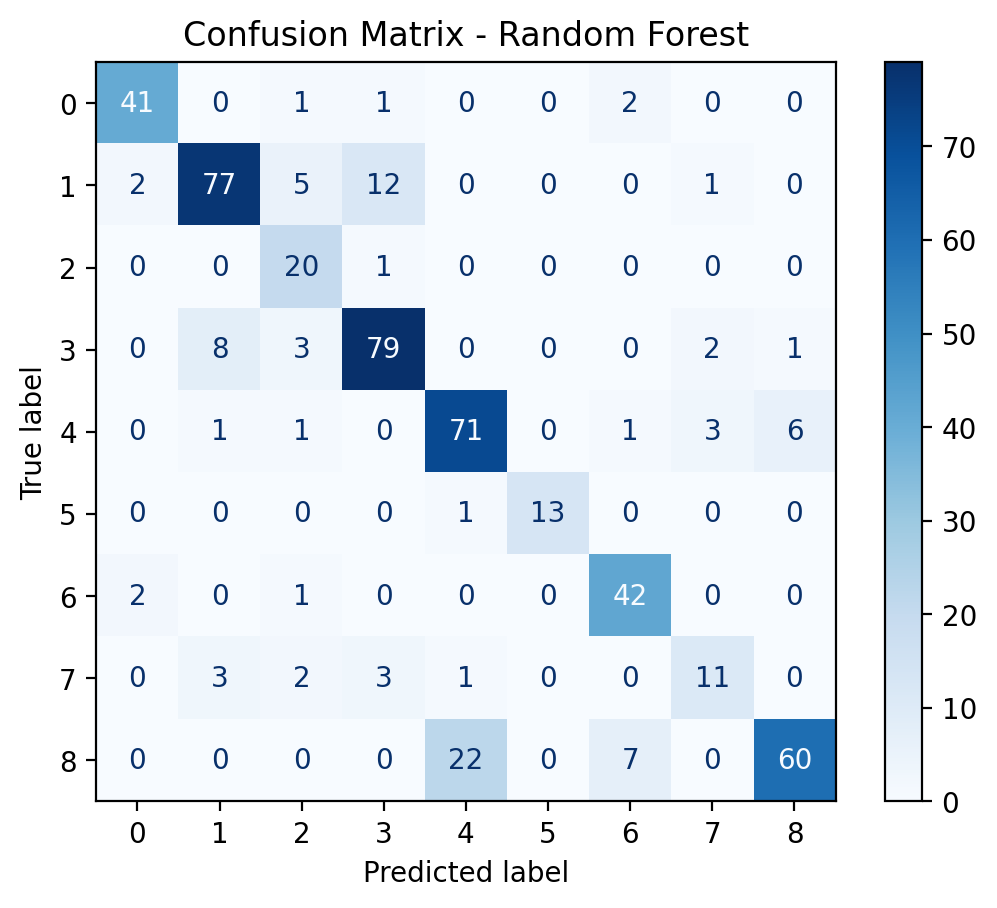}
  \caption{Confusion matrix (Random Forest).}
\end{subfigure}

\caption{One-vs-all ROC and confusion matrix for Random Forest.}
\label{fig:pair_rf}
\end{figure}

\medskip
\noindent\textbf{Fig.~\ref{fig:pair_rf} (Random Forest).}
Steep ROC curves with near-uniform high AUC across classes; the confusion matrix is sharply diagonal, matching RF’s strong accuracy and robustness on heterogeneous tabular features.

\begin{figure}[htbp]
\centering
\captionsetup[sub]{font=small}

\begin{subfigure}{0.49\linewidth}
  \centering
  \includegraphics[width=\linewidth]{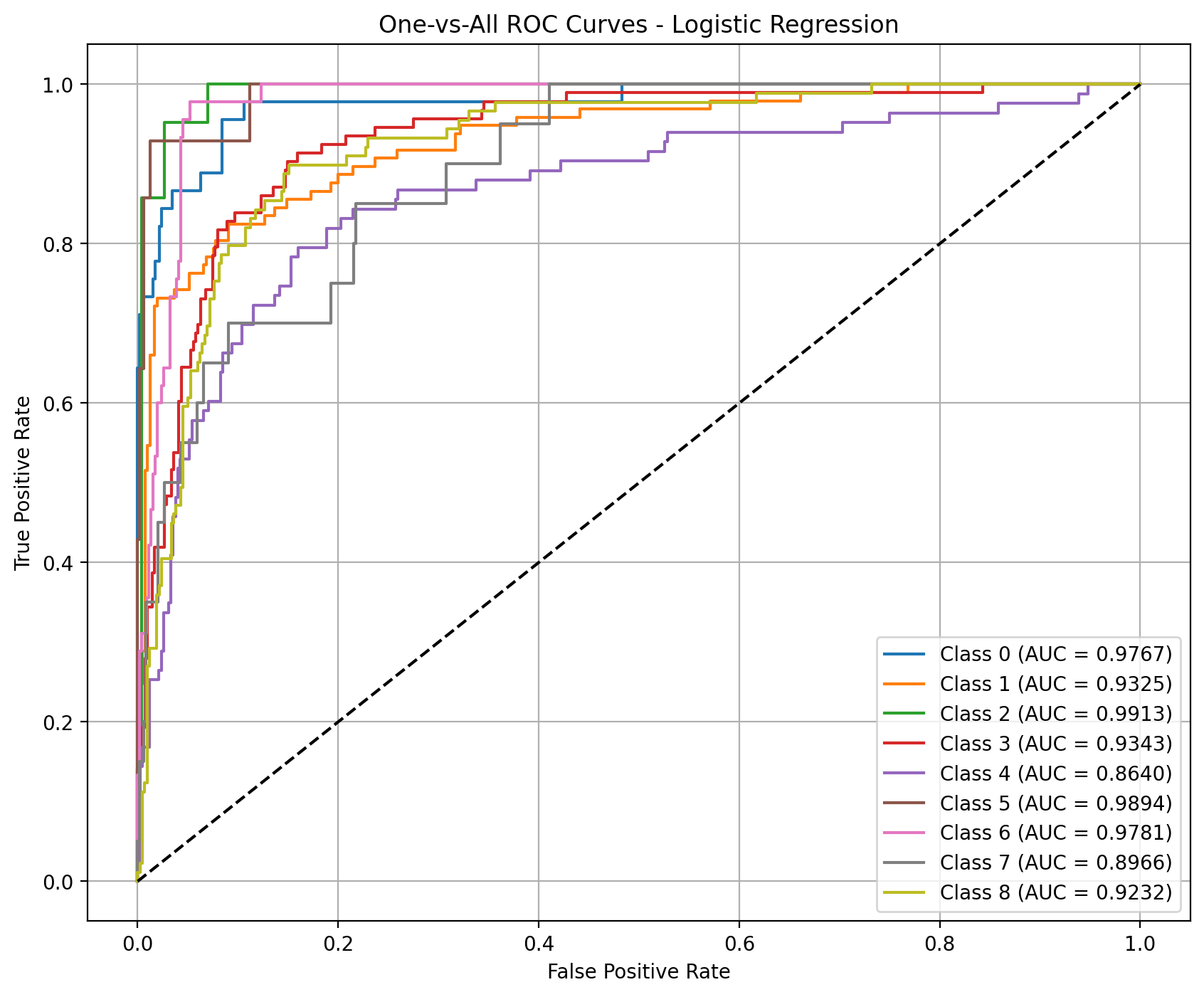}
  \caption{ROC (Logistic Regression).}
\end{subfigure}\hfill
\begin{subfigure}{0.49\linewidth}
  \centering
  \includegraphics[width=\linewidth]{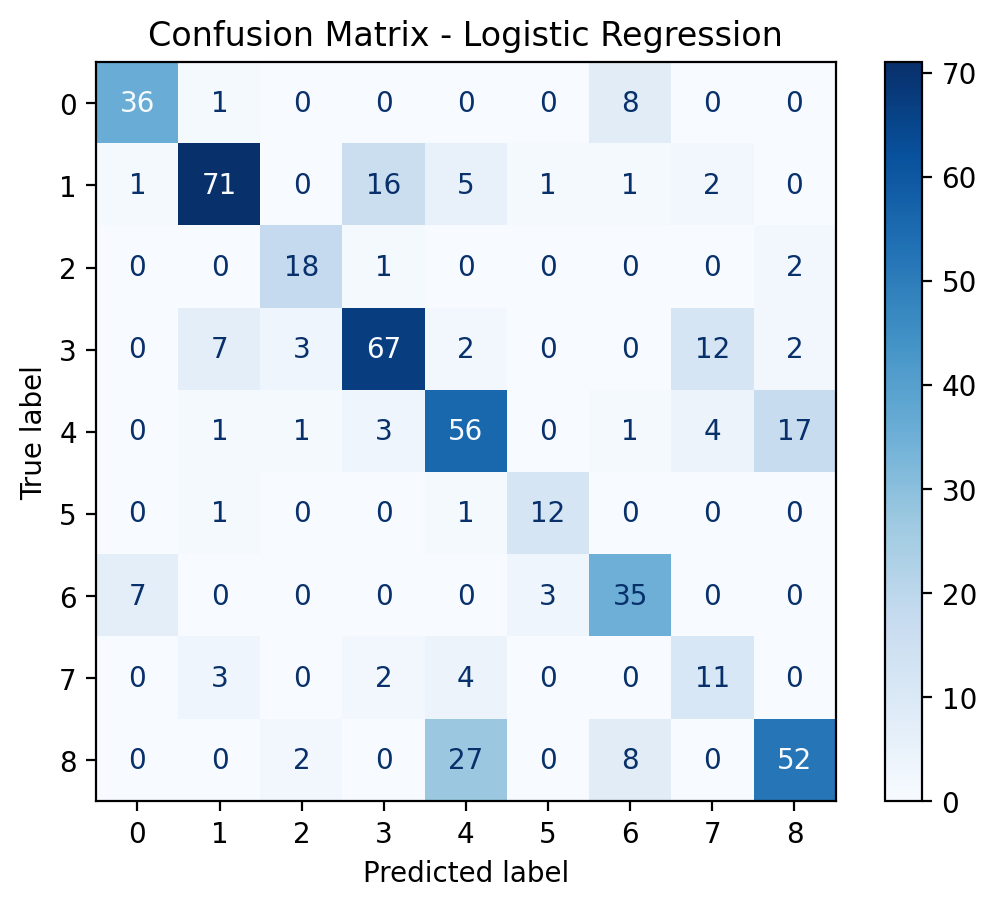}
  \caption{Confusion matrix (Logistic Regression).}
\end{subfigure}

\caption{One-vs-all ROC and confusion matrix for Logistic Regression.}
\label{fig:pair_lr}
\end{figure}

\medskip
\noindent\textbf{Fig.~\ref{fig:pair_lr} (Logistic Regression).}
High AUC on several classes, but the confusion matrix shows increased spillover among visually similar categories, consistent with linear boundaries under class overlap.

% ===== A) TabSeq (standard CE) =====

% ===== B) TabTransformer (standard CE) =====

% ===== C) FT-Transformer (standard CE) =====

% ===== D) TabNet (standard CE) =====

% ===== E) 1D CNN (standard CE) =====

% ===== 1) Gradient Boosting (GBM) =====

% ===== 2) XGBoost =====

% ===== 3) AdaBoost =====

% ===== 4) CatBoost =====

% ===== 5) MLP (non-tree baseline) =====

% ===================== Naive Bayes =====================

% ===================== KNN =====================

% ===================== Decision Tree =====================

% ===================== SVM =====================

% ------------ Random Forest: ROC + Confusion ------------

% ------------ Logistic Regression: ROC + Confusion ------------

\end{document}